\documentclass[11pt]{article}
\usepackage[final]{acl}
\usepackage{times}
\usepackage{latexsym}
\usepackage[T1]{fontenc}
\usepackage[utf8]{inputenc}
\usepackage{microtype}
\usepackage{inconsolata}
\usepackage{graphicx}
\usepackage{comment}
\usepackage{booktabs}  
\usepackage{colortbl} 
\usepackage{multirow} 
\usepackage{amssymb}
\usepackage{amsmath}
\usepackage{pdfpages}  
\usepackage{subcaption} 
\usepackage{graphicx}  
\usepackage{float}
\usepackage{hyperref}
\usepackage[whole]{bxcjkjatype}

\usepackage{subcaption} 
\usepackage{listings}

\title{Are Checklists Really Useful for Automatic Evaluation of Generative Tasks? }

\author{
Momoka Furuhashi${}^{1,2}$　
Kouta Nakayama${}^{2}$　
Takashi Kodama${}^{2}$　
Saku Sugawara${}^{3,2}$\\
${}^{1}$Tohoku University　${}^{2}$Research and Development Center for Large Language Models, \\ National Institute of Informatics　${}^{3}$National Institute of Informatics\\
\texttt{furuhashi.momoka.p4@dc.tohoku.ac.jp}
　\texttt{\{nakayama,tkodama,saku\}@nii.ac.jp}}
 
\begin{document}
\maketitle
\begin{abstract}
Automatic evaluation of generative tasks using large language models faces challenges due to ambiguous criteria. 
Although automatic checklist generation is a potentially promising approach, its usefulness remains underexplored.
We investigate whether checklists should be used for all questions or selectively, generate them using six methods, evaluate their effectiveness across eight model sizes, and identify checklist items that correlate with human evaluations.
Through experiments on pairwise comparison and direct scoring tasks, we find that selective checklist use tends to improve evaluation performance in pairwise settings, while its benefits are less consistent in direct scoring.
Our analysis also shows that even checklist items with low correlation to human scores often reflect human-written criteria, indicating potential inconsistencies in human evaluation. These findings highlight the need to more clearly define objective evaluation criteria to guide both human and automatic evaluations.
\footnote{Our code is available at~\url{https://github.com/momo0817/checklist-effectiveness-study}}

\end{abstract}
\section{Introduction}
Automatic evaluation using large language models (LLMs) has been widely adopted for generative tasks~\citep{chang2024survey, Ferraz2024-ba, gu2024survey, li2024generation}.
This approach provides an efficient and scalable alternative to costly and time-consuming human evaluation.
However, it faces two major challenges.
First, establishing clear and consistent evaluation criteria remains difficult, leading to potential ambiguity in scoring. Second, the correlation between LLM-based automatic evaluation and human evaluation is often unstable, limiting its reliability.
\begin{figure}[t]
    \centering
    \small
    \includegraphics[width=0.475\textwidth]{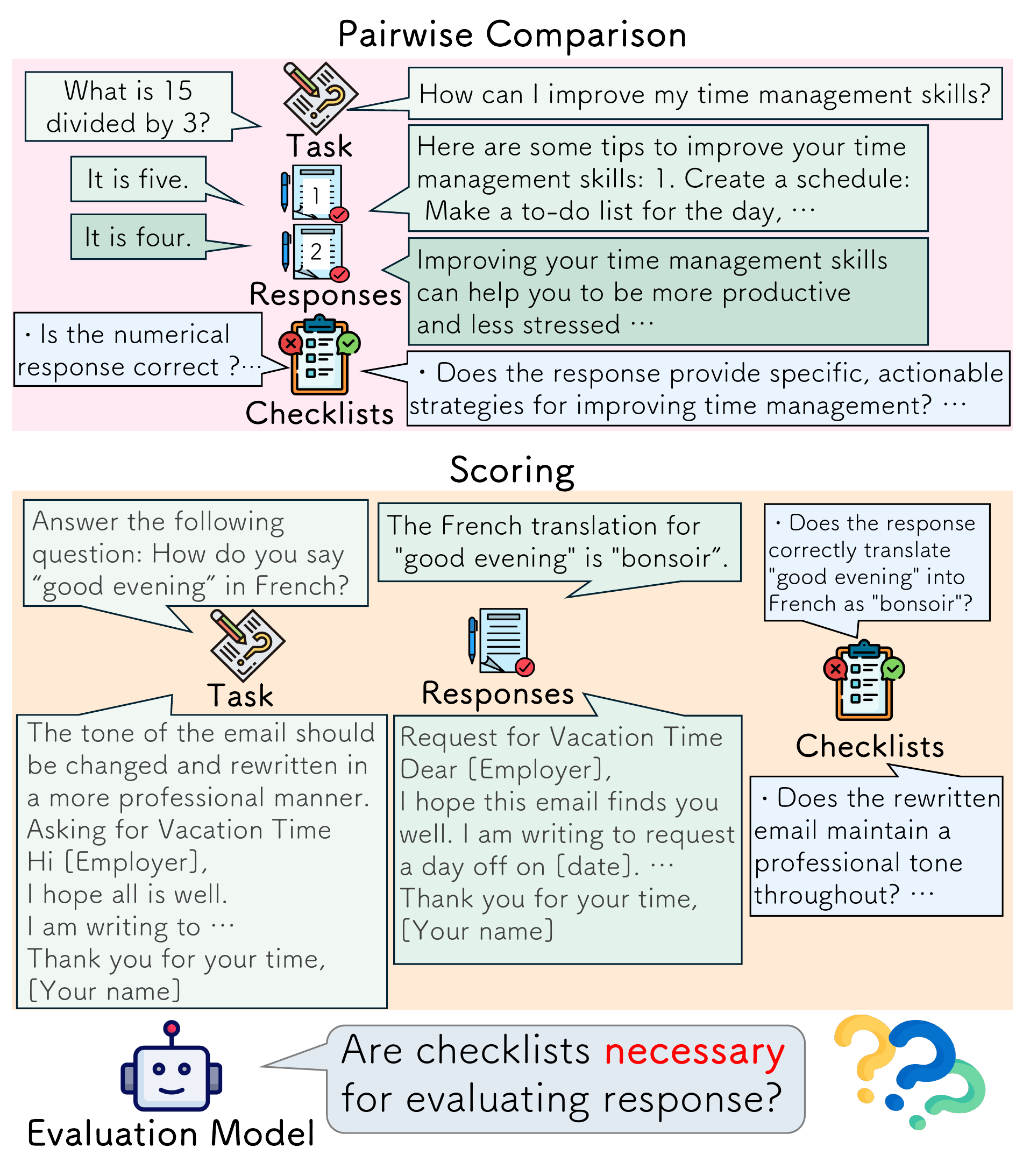} 
    \caption{
    Examples of using checklists in automatic evaluation by LLMs.
  Existing studies use checklists even in situations where fine-grained criteria may be unnecessary for evaluating the responses.}
    \label{fig:figure1}
\end{figure}
\begin{figure*}[t]
    \centering
    \small
    \includegraphics[width=1\textwidth]
    {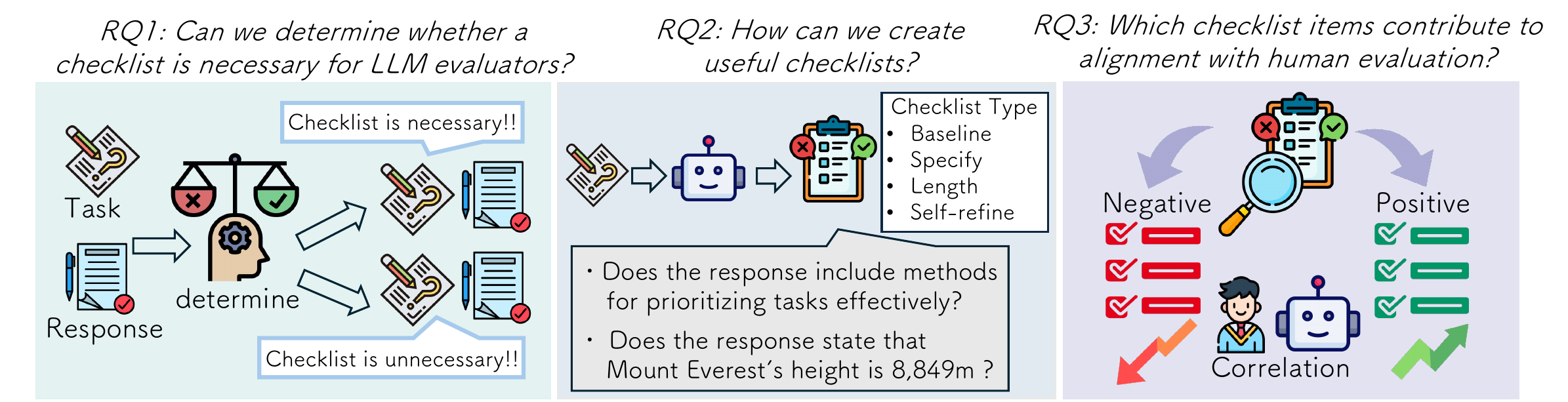} 
    \caption{Our research questions. First, we investigate whether we can identify which responses require checklist evaluation (RQ1). Next, we study how checklist generation affects alignment with human evaluations by evaluating eight models of different sizes (RQ2), comparing six different generation methods. Finally, we analyze which checklist items contribute most to alignment with human evaluation (RQ3).}
    \label{fig:overview}
\end{figure*}

To address these challenges, previous studies have introduced a \textit{checklist} approach that decomposes evaluation criteria into specific, fine-grained items \citep{lee2024checkeval, qin2024InFoBench, yuchen2024wildbench}.
As shown in Figure~\ref{fig:figure1}, when an LLM evaluates responses to mathematical problems, evaluator models can refer to a detailed checklist with items such as ``Does the calculated value match the correct answer?'' and ``Does the response contain unnecessary decimal points?''
Although checklists are easy to use and understand, previous studies have not fully investigated three key aspects: when checklists are necessary, how we can create them, and how checklist items relate to alignment with human evaluation.
We examine the usefulness of checklists in automatic evaluation by answering the following three questions shown in Figure~\ref{fig:overview}.
RQ1: \textit{Can we determine whether a checklist is necessary for LLM evaluators?}
RQ2: \textit{How can we create useful checklists?}
RQ3: \textit{Which checklist items contribute to alignment with human evaluation?}

To investigate RQ1 and RQ2, we conduct three controlled experiments.
First, focusing on the consistency of multiple automatic evaluations, we set a threshold to decide whether a checklist is necessary for each evaluation case.
Second, we investigate which checklist features improve correlation.
In this process, we create three methods (control checklist length, more specified generation, and self-refine) in addition to the baseline and existing method.
Third, we evaluate a total of eight models, including gpt-4o-2024-08-06~\cite{Achiam2023GPT4TR} and Qwen2.5-7B-Instruct~\cite{Yang2024Qwen25TR}, to investigate which checklist items contribute to alignment with human evaluation.
For each of these three aspects, we test their effect on alignment using two types of human evaluation data: pairwise comparison and direct scoring.
To investigate RQ3, we also conduct a more detailed checklist-based analysis on the smallest model.
We analyze checklist items both quantitatively and qualitatively, focusing on their overlap with human-written ones.

Our experiments yield three key findings:
First, we observe that the effectiveness of selective checklist application varies by task; in some cases, it achieves comparable or better correlations than full application, while in others it does not.
This suggests that checklist use is not universally beneficial.
Second, our analysis reveals that the most useful checklist creation method varies across different evaluation models and tasks, suggesting that no single approach works best in all settings.
Third, our analysis shows that many checklist items, although useless for improving human correlation, still overlap substantially with human-written items.
This suggests that inconsistencies may stem from the subjective nature of human evaluations and underscores the need to rethink the objective criteria we expect from responses.
%\footnote{We will make our scripts and data publicly available.}

Our contributions are summarized as follows:
\begin{itemize}
\item We find that selective checklist use sometimes improves evaluation outcomes, suggesting that omitting checklists can be justified in specific settings.
\item We show that no universally optimal checklist generation method exists, as usefulness varies significantly depending on the evaluation model and use case.
\item We find that even checklist items with low correlation to human evaluations often overlap with human-written ones, indicating they may still capture valid criteria. This highlights the subjective nature of human evaluations and calls for more objective evaluation design.
\end{itemize}

\section{Related Work}
Recent studies have investigated the use of LLMs as evaluators for generative tasks~\citep{chang2024survey, gu2024survey, li2024generation}.
Automatic evaluation methods using LLMs fall into two methods: pairwise comparison~\citep{wei2023chainofthoughtprompting, wang-etal-2024-large-language-models-fair,   zeng2024LLMBar, RewardBench,judgebench2024} and direct scoring~\citep{Ye2023FLASKFL, kim2024prometheus, liu-etal-2024-benchmarking}.
These approaches rely on human evaluation as gold labels and assess performance via correlation and agreement rates.
However, both methods have inherent limitations: pairwise comparison suffers from ambiguity in evaluation criteria, while direct scoring faces difficulties in metric definition.
\paragraph{Fine-grained Evaluation Criteria}
Previous studies have explored breaking down evaluation criteria into smaller components to improve correlation.
\citet{min-etal-2023-factscore} evaluate factual accuracy by splitting responses into individual statements, each containing a single piece of information.
\citet{kim2024prometheus} manually create 50 scoring rubrics focusing on critical aspects of response evaluation and then expand these rubrics using GPT-4. \citet{Ye2023FLASKFL} enhances evaluation reliability by decomposing their evaluation into skill-level scoring sets for each instruction.

\paragraph{Checklist-based Evaluation}
 Previous studies have proposed a checklist-based approach that breaks down complex evaluation criteria into smaller, more specific points of evaluation~\citep{lee2024checkeval, qin2024InFoBench, yuchen2024wildbench, Cook2024TICKingAT}．
CheckEval~\cite{lee2024checkeval} decomposes evaluation criteria such as fluency into manually created checklists for summarization tasks, where each item requires a binary yes/no response, with the final score derived from the ratio of \textit{yes} responses.
Additionally, \citet{qin2024InFoBench} manually create 2,500 checklists based on 500 distinct instructions and conduct comprehensive evaluations using six evaluation models.
Furthermore, WildBench~\cite{yuchen2024wildbench} establishes a benchmark for evaluating LLMs on real-world-inspired tasks, generating five to ten checklist items for each question task by using GPT-4-Turbo and Claude-3-Opus.
While \citet{Cook2024TICKingAT} shows that LLM-generated checklists improve correlations, previous studies have not investigated when checklists are actually needed or how useful they are.

\section{Dissecting Checklist-based Evaluation}
To investigate RQ1 and RQ2, we conduct three controlled experiments:
First, we investigate whether it is possible to identify instances where checklists are unnecessary for automated evaluation (Session~\ref{Identify_checklist});
second, we evaluate six different checklist generation methods to determine which types of checklists are most useful (Session~\ref{variations_checklist});
third, we examine the usefulness of checklists using eight different models, ranging from small to large size, to assess their practicality (Session~\ref{choice_model}).

\subsection{Identifying When Responses Need Checklist Evaluation}
\label{Identify_checklist}
To address RQ1: \textit { Can we determine whether a checklist is necessary for LLM evaluators?}, we compare how well model evaluations correlate with human evaluation both with and without checklists. 
We hypothesise that checklists are necessary when LLM evaluations lack consistency.
Therefore, we conduct multiple evaluations without checklists and apply checklists only to responses that receive inconsistent labels above a threshold.
We then compare this selective approach against two baselines: using no checklists at all (\textit{None}) and using checklists for every response (\textit{All}).
Through this comparison, we determine if targeting the checklist use to low-reliability cases improves overall correlation.
We conduct this analysis across different checklist variations described in Section~\ref{variations_checklist}.

\subsection{Checklist Generation Policy}
\label{variations_checklist}
To address RQ2: \textit{How can we create useful checklists?}, we vary the level of detail and number of items. To better control these factors, we generate checklists for evaluating generative tasks using three methods.
We analyze how each method correlates to identify the most useful checklist types.
Below, we describe each checklist generation method.

\paragraph{Baseline}
\noindent
In this study, we examine how limiting the number of items and adjusting the level of detail affect checklist generation.
We incorporate the following three elements:
(1) Each item must allow a simple \textit{yes} or \textit{no} answer, where \textit{yes} confirms success.
(2) Criteria must directly relate to essential task requirements.
(3) Questions must use specific wording and reference input phrasing directly, concrete wording that directly relates to the task, avoiding vague or ambiguous language.

\paragraph{Specify}
\noindent
Previous studies distinguish between two types of checklist items: surface-level evaluation (e.g., response correctness) and content-specific evaluation (e.g., Does the response state that  Mount Everest's height is 8,849m?).
Therefore, we add to the baseline that checklist questions should be designed considering possible answers to the input.
 
\paragraph{Checklist Length}
\noindent
While previous studies~\citep{yuchen2024wildbench, Cook2024TICKingAT} use a fixed number of items in their checklists, we hypothesize that the optimal number of items depends on the task and should be adjusted accordingly.
Therefore, we evaluate how performance changes when we generate checklists containing 0.5 and 1.5 times the number of items for given Baseline checklists.

\paragraph{Self-refine}
\noindent
~\citet{Cook2024TICKingAT} use LLMs to generate both responses and checklists for tasks, then evaluate responses using these checklists and perform multiple rounds of self-refine on the responses.
However, they do not apply self-refine to the checklist generation process itself.
In this study, we extend their approach by implementing self-refine for the checklists to improve their quality.
Specifically, our checklist generation model generates a Likert scale evaluation and accompanying feedback based on the baseline prompt and uses this feedback to regenerate improved checklists.

\paragraph{Ticking}
\label{ticking}
\noindent
As a representative of existing methods, we use ~\citet{Cook2024TICKingAT}'s original prompts.
This prompt includes several examples and a limit on the number of checklist items, ranging from two to eight.
However, since the original paper does not specify the examples they use, we remove them from our implementation.

\subsection{Evaluator Models of Different Sizes}
\label{choice_model}
Previous studies have used a limited variety of evaluation models. While the InFoBench~\cite{qin2024InFoBench} uses LLMs such as GPT-4 for evaluation, their smallest model is vicuna-13b-v1.5~\cite{vicuna2023}, limiting practical applications. 
Moreover, their analysis includes only a single smaller model without comparing different sizes of the same model or exploring how checklist usage affects correlation across varying model sizes. To address these limitations, we evaluate the usefulness of checklists across eight models ranging from 7B to 32B parameters, including multiple sizes of the same model family, as detailed in Section~\ref{evaluation_models}.

\begin{comment}
\begin{figure}[t]
    \centering
    \includegraphics[width=0.45
    \textwidth, height=0.3\textheight, keepaspectratio]
    {figure/the LLMBar0.5_rate.pdf} 
    \caption{Coverage level (\%) of baseline checklist implementation in the LLMBar.}
    \label{fig:the LLMBar0.5_rate}
\end{figure}
\end{comment}

\section{Experiments}
To investigate the usefulness of checklist-based automatic evaluation, this study conducts experiments on two tasks: (1) a pairwise comparison task, in which pairs of LLM's responses are judged for relative quality, and (2) a direct scoring task, in which LLM's responses are rated using a Likert scale. 

\subsection{Dataset}
We use datasets with human-annotated LLM responses, which include reliable evaluation labels and cover diverse real-world tasks with multiple subsets. Such datasets are rare, as it is uncommon to find ones that combine both high-quality human evaluation and broad task diversity. The two datasets we employ sufficiently meet these criteria.

\paragraph{Pairwise Comparison}
For the pairwise comparison task, we use the LLMBar~\cite{zeng2024LLMBar} dataset, which comprises eight English subsets, including three major categories: \textit{Adversarial}, \textit{Natural}, and \textit{Processed}. The \textit{Adversarial} subset includes inputs specifically designed to mislead LLMs when used as evaluators, while the \textit{Natural} subset contains inputs collected and modified from existing human preference datasets. the LLMBar dataset exhibits an inter-annotator agreement exceeding 90\%, demonstrating its reliability.
The \textit{Processed} subset consists of processed versions of three existing datasets  (FairEval~\cite{wang-etal-2024-large-language-models-fair},  LLMEval-2~\cite{zhang2023wLLMEval}, and MT-Bench ~\cite{zheng2023judgingllmasajudgemtbenchchatbot}). These datasets have been refined by~\citet{zheng2023judgingllmasajudgemtbenchchatbot} to improve data fairness.
Finally, we obtained manual annotations for 885 response pairs.

\paragraph{Direct Scoring}
For the direct scoring task, we use the InFoBench~\cite{qin2024InFoBench} dataset.
the InFoBench dataset uses a Likert scale~(1 to 5) as its metric and consists of two English subsets: a simpler section (\textit{Easy} subset) and a more challenging section (\textit{Hard} subset).
For each input prompt, responses are collected from five distinct language models (GPT-3.5-turbo~\cite{Ouyang2022TrainingLM}, GPT-4, Claude-v1~\cite{Bai2022TrainingAH}, Alpaca-7B~\cite{alpaca}, and Vicuna-13B~\cite{vicuna2023}).
The responses are then manually annotated by three expert evaluators, who are natural language processing specialists according to prior research.
The correlation coefficient is reported as 0.353 for the \textit{Easy} set and 0.519 for the \textit{Hard} set.
This dataset consists of the manual evaluation results for five LLMs' responses across 50 tasks, resulting in a total of 250 annotated samples. 
In this study, we determine the gold label for each sample by rounding the mean of the three manually annotated labels.

\begin{comment}
\begin{figure}[t]
    \centering
    \includegraphics[width=0.45
    \textwidth, height=0.3\textheight, keepaspectratio]
    {figure/the InFoBench_rate.pdf} 
    \caption{Coverage level (\%) of baseline checklist implementation in the InFoBench.}
    
    \label{fig:the InFoBench_rate}
\end{figure}
\end{comment}

\subsection{Checklist Generation}
Regardless of the variation, each checklist item is formatted to allow for either a \textit{yes} or \textit{no} response. We employ six different checklist generation policies (detailed in Section~\ref{variations_checklist}) using gpt-4o-2024-08-06~\cite{Achiam2023GPT4TR} as the generation model.

\begin{figure*}[tb]
  \centering

  \begin{subfigure}{\linewidth}
  \centering
    \caption{Pairwise comparison (the LLMBar)}   
    \label{fig:the LLMBar_variation_comparison}
    \includegraphics[width=0.99\linewidth]{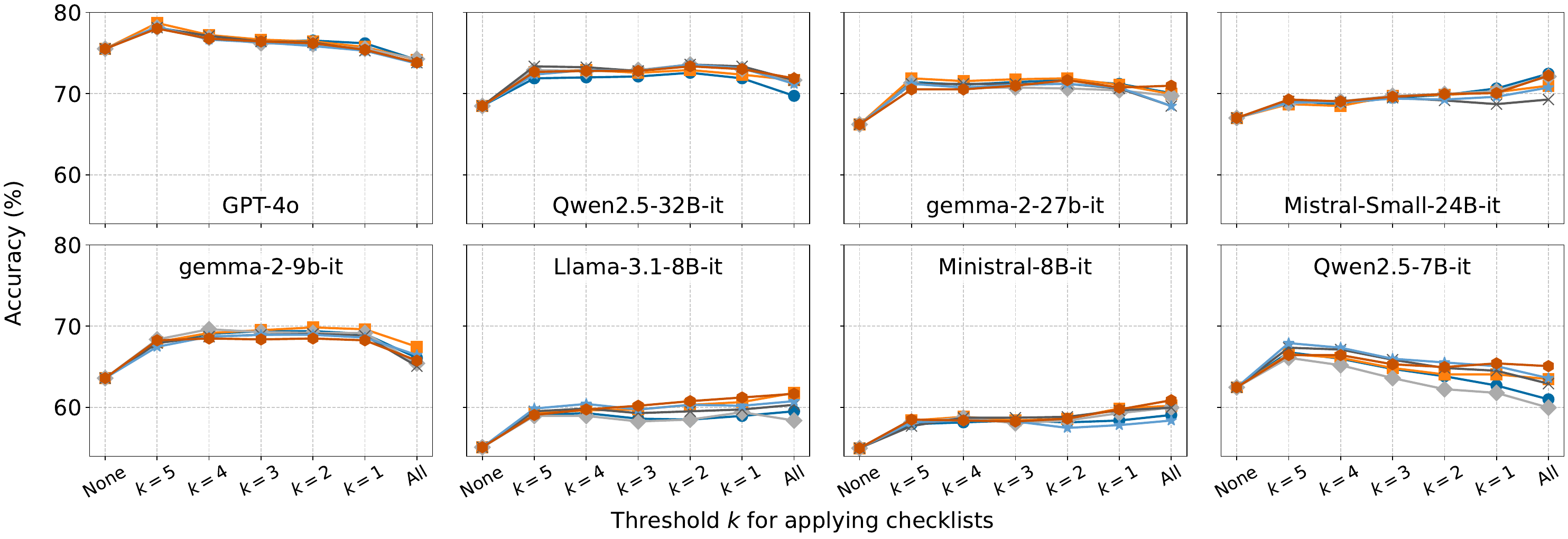}
    \label{fig:hogehoge}
  \end{subfigure}

  \vspace{-0.5em}                  
  
  \begin{subfigure}{\linewidth}
  \centering
    \caption{Direct scoring (the InFoBench)}
    \label{fig:the InFoBench_variation_comparison}
    \includegraphics[width=\linewidth]{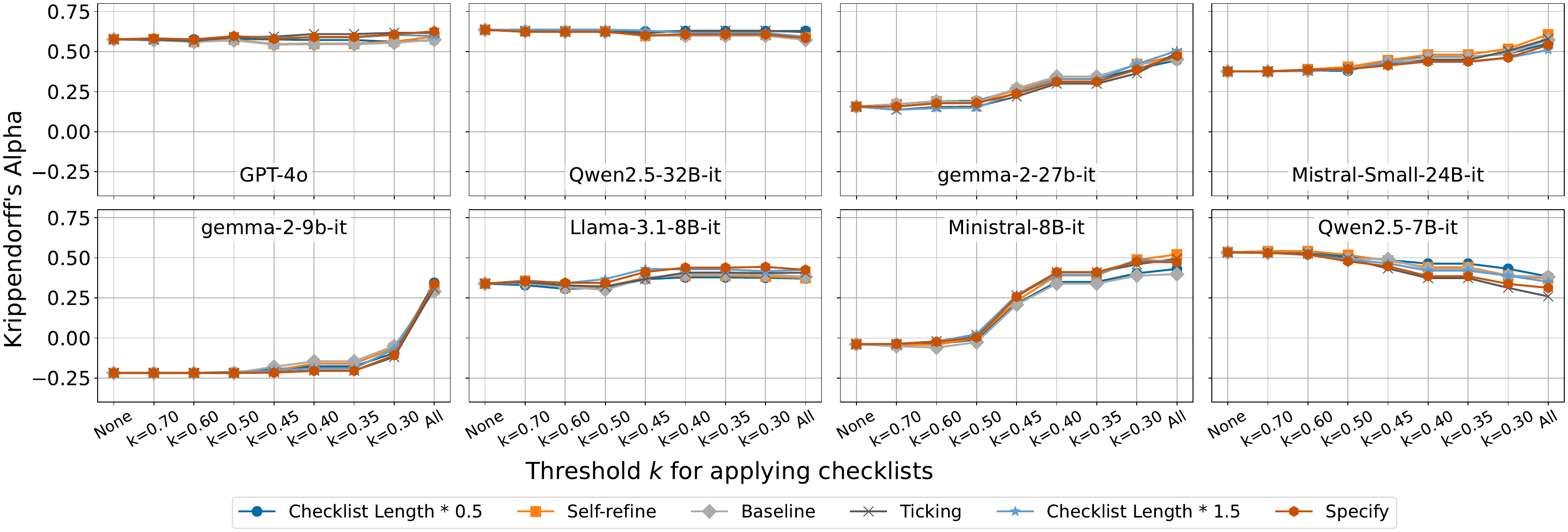}
    \label{fig:hugehuge}
  \end{subfigure}

  \caption{Comparison of accuracy for checklist application method. \textit{None} indicates that the checklist is not used during evaluation, while \textit{All} signifies that the checklist is applied to every evaluation. The parameter $k$ represents the threshold for applying the checklist; the smaller the value of $k$, the more frequently the checklist is employed.}
  \label{fig:stacked}
\end{figure*}

\subsection{Automatic Evaluation}
\label{evaluation_models}

To investigate the usefulness of checklists across different model sizes, we evaluate eight models: gpt-4o-2024-08-06 (GPT-4o), Qwen2.5-32B-Instruct (Qwen2.5-32B-it), Qwen2.5-7B-Instruct (Qwen2.5-7B-it)~\cite{Yang2024Qwen25TR}, Mistral-Small-24B-Instruct-2501 (Mistral-Small-24B-it), Ministral-8B-Instruct-2410 (Ministral-8B-it)~\footnote{\url{https://huggingface.co/mistralai/Mistral-Small-24B-Instruct-2501}, \url{https://huggingface.co/mistralai/Ministral-8B-Instruct-2410}}, 
Gemma-2-27b-it, gemma-2-9b-it~\cite{gemma_2024}, Llama-3.1-8B-Instruct (Llama-3.1-8B-it)~\cite{dubey2024llama}.
These models represent different parameter sizes and capabilities.

For evaluations without checklists, we use Chain-of-Thought prompting~\cite{wei2023chainofthoughtprompting}.  
We first prompt the model to output the reason for its evaluation, and then obtain the final result. 
For evaluations with checklists, we first ask the evaluation model to choose the evaluation result for each checklist item from \textit{yes}, \textit{no}, or \textit{n/a}, and then obtain the final evaluation result. \textit{n/a} indicates the item is skipped as it does not apply to the response.

For both tasks, we evaluate each response ten times. To mitigate position bias in pairwise comparison, we use each order five times.
The checklist generation prompts are provided in Appendix~\ref{checklist_generation_prompt}.

\subsection{Evaluation Metrics}
For the pairwise comparison dataset, we use accuracy as the evaluation metric for evaluating the performance of our automatic evaluation.
The final evaluation result is determined by a majority vote across multiple evaluations; if the votes are evenly split, the outcome is considered a tie.
However, because the LLMBar~\cite{zeng2024LLMBar} provides only binary labels (win or lose) and does not include a tie label, we assign a score of 0.5 to a tie when calculating accuracy.
This allows us to treat accuracy as an expected value under realistic deployment scenarios.
This adjustment reflects our goal of evaluating the potential practical benefits of checklist-based evaluation.

For the direct scoring dataset, we use Krippendorff's alpha~\cite{Hayes01042007} to measure the agreement between automatic and human labels.
The final evaluation result of automatic evaluation is obtained by taking the mean of ten evaluations and rounding the value.

\section{Results}
In total, we generate 22,985 checklist items, specifically 21,475 items in the LLMBar and 1,510 from the InFoBench.
For detailed statistical analysis of variations and thresholds, see Appendix~\ref{checklist stats}.
\subsection{Identifying When Responses Need Checklist Evaluation}
\label{subsec:identifying_when}
We define a threshold $k$ to determine when to apply checklists based on evaluation inconsistency. Due to the different nature of our evaluation tasks, we use different inconsistency metrics $x_{\mathrm{pairwise}}$ and $x_{\mathrm{direct}}$ for each setting.

\begin{table*}[h]
\scriptsize 
\centering
\tabcolsep 3pt 
\renewcommand{\arraystretch}{1.5} 
\begin{tabular}{lcccccccc}
\toprule
Model & GPT-4o & Qwen2.5-32B-it & gemma-2-27b-it & Ministral-24B-it& gemma-2-9b-it & Ministral-8B-it & Llama-3.1-8B-it & Qwen2.5-7B-it\\ 
\midrule
\rowcolor{gray!20}Best Policy & Specify &  Length * (0.5, 1.5) & Specify & Ticking & Self-refine & Ticking & Ticking & Self-refine \\ 
Worst Policy & Self-refine &  None & None & None & None & None & None & Length * 0.5 \\ 
\bottomrule
\end{tabular}

\caption{Best and worst settings of checklist use for each evaluation model in the pairwise comparison.}
\label{tb:the LLMBar_top_variation}
\end{table*}
\begin{comment}
\begin{table*}[t]
\scriptsize % フォントサイズを小さく設定
\centering
\tabcolsep 3pt % カラム間隔をさらに小さく設定
\renewcommand{\arraystretch}{1.5} % 表の高さ調整
\begin{tabular}{lcccccccc}
\toprule
Model & GPT-4o & Qwen2.5-32B-it & gemma-2-27b-it & Ministral-24B-it& gemma-2-9b-it & Ministral-8B-it & Llama-3.1-8B-it & Qwen2.5-7B-it\\ 
\midrule
\rowcolor{gray!20} Policy & Specify & Specify, Length * 0.5& Length * 1.5 & Self-refine& Length * 0.5 & Self-refine & Specify & Self-refine \\ 
Threshold  & All & None & All & All  &All& All & 0.4 & 0.7 \\ 
\rowcolor{gray!20}Krippendorff's $\alpha$   & 0.635 & 0.637 &  0.493& 0.568&0.341& 0.523& 0.481  & 0.557 \\ 
$\Delta$ from Baseline  & 0.05 & 0 & 0.368 & 0.21& 0.583 & 0.60 & 0.188 & 0.019 \\ 
\bottomrule
\end{tabular}
\caption{Best settings of checklist use for each evaluation model in the direct scoring.}
\label{tb:infobench_top_variation}
\end{table*}
\end{comment}

\begin{table*}[t]
\scriptsize 
\centering
\tabcolsep 3pt 
\renewcommand{\arraystretch}{1.5}
\begin{tabular}{lcccccccc}
\toprule
Model & GPT-4o & Qwen2.5-32B-it & gemma-2-27b-it & Mistral-Small-24B-it & gemma-2-9b-it & Llama-3.1-8B-it & Ministral-8B-it & Qwen2.5-7B-it \\
\midrule
\rowcolor{gray!20}Best Policy & Specify & Length * 1.5 & Length * 1.5 & Self-refine &  Length * 0.5 & Specify & Self-refine & Self-refine \\
Worst Policy & Baseline &  Baseline & Length * 1.5 & Length * 1.5 & Specify & Baseline & Baseline & Ticking \\

\bottomrule
\end{tabular}
\caption{Best and worst settings of checklist use for each evaluation model in the direct scoring.}
\label{tb:infobench_top_variation}
\end{table*}

\paragraph{Pairwise Comparison Setting}
We define the inconsistency value $x_{\mathrm{pairwise}}$ as the number of votes the less-preferred response receives. For example, if the evaluations for Response 1 and Response 2 are $[1,1,1,1,1,1,1,2,2,2]$, $x_{\mathrm{pairwise}}=3$; for $[1,1,1,1,1,2,2,2,2,2]$, $x_{\mathrm{pairwise}}=5$. 
We apply checklists only when $x_{\mathrm{pairwise}} \geq k$, where $k$ is selected from $\{1, 2, 3, 4, 5\}$.

\paragraph{Direct Scoring Setting}
 We define the inconsistency value $x_{\mathrm{direct}}$ as the standard deviation of these evaluations.
 For example, if the evaluation labels are $[3,3,3,3,4]$, $x_{\mathrm{direct}}=0.4$; for $[2,3,3,3,4]$, $x_{\mathrm{direct}}=0.63$.
We apply checklists only when $x_{\mathrm{direct}} \geq k$.
Based on our observations, most $x_{\mathrm{direct}}$ values fall in the range of 0.3 to 0.8, with a notable concentration between 0.3 and 0.5. Therefore, we select $k$ from $\{0.3, 0.35, 0.4, 0.45, 0.5, 0.6, 0.7\}$.

\paragraph{} Figures~\ref{fig:the LLMBar_variation_comparison} and~\ref{fig:the InFoBench_variation_comparison} show the experimental results.
\textit{None} indicates that no checklist is used during evaluation, while \textit{All} denotes that all available checklists are applied.
Detailed checklist application rates are provided in Appendix~\ref{Checklists Application Rate}.

Our results demonstrate that the impact of selective checklist application varies across datasets.
In the pairwise comparison, we observe that selective checklist application often improves evaluation performance over both the \textit{None} and \textit{All}, for several models, including GPT-4o, Qwen2.5-32B-it, Gemma-2-27B-it, Gemma-2-9B-it, and Qwen2.5-7B-it.
In the direct scoring, we observe no improvements from selective checklist usage in direct scoring, where its performance often matches or falls below that of the \textit{None} and \textit{All}.

\paragraph{Bootstrap Sampling} 
We also conduct a bootstrap test to evaluate whether the selective application of checklists leads to improvements.
For pairwise comparison, we observe statistically significant differences in 20 out of 48 cases, suggesting that selectively applying checklist items can be beneficial under certain conditions. 
 On the other hand, for direct scoring, we observe no statistically significant differences across any of the six checklist-generation policies evaluated with eight evaluation models, indicating that this approach does not yield measurable improvements under the tested conditions.  
For detailed settings and results, see Appendix~\ref{Bootstrap Sampling}.

\begin{comment}
\begin{figure}[t]
    \centering
    \includegraphics[width=0.45
    \textwidth, height=0.3\textheight, keepaspectratio]
    {figure/the InFoBench_baseline.pdf} 
    \caption{Evaluation results of the InFoBench using Krippendorff's Alpha across different language models with baseline checklist. The thresholds were varied from 0.1 to 1.0 in increments of 0.1, as no examples showed standard deviations above 1.1 .}
    
    \label{fig:the InFoBench_baseline}
\end{figure}
\end{comment}

\subsection{Checklist Generation Policy}
Next, we present the results of the checklist generation policy in Figures~\ref{fig:the LLMBar_variation_comparison},~\ref{fig:the InFoBench_variation_comparison}, and Appendix~\ref{Checklists Application Rate}.
We do not find any specific variation that consistently outperforms others. The best checklist variation depends on the evaluation tasks and evaluator models.
For pairwise comparison, Specify works well with GPT-4o and Gemma-2-27b-it, while Ticking suits Ministral-24B-it, Ministral-8B-it, and Llama-3.1-8B-it.
In direct scoring, Specify is effective for GPT-4o and Llama-3.1-8B-it, whereas Self-refine performs best with Ministral-24B-it, Ministral-8B-it, and Qwen2.5-7B-it.
These findings suggest that checklist methods should be adapted to specific evaluator models and tasks.

\paragraph{Useful and Not Useful Checklist Settings}
We do not find any checklist generation policies that are consistently superior or inferior across all settings. 
However, we conduct an in-depth analysis of which policies are \textit{useful} or \textit{useless} for datasets.

Tables~\ref{tb:the LLMBar_top_variation} and~\ref{tb:infobench_top_variation} summarize the best and worst performing checklist generation methods, including \textit{None} for each dataset.
Both Self-refine and Specify tend to perform well across the two datasets.
Specify is useful because it produces checklists containing more detailed information, which helps clarify the evaluation criteria.
Self-refine, on the other hand, involves having the LLMs revise the baseline checklist, often resulting in more refined and input-relevant items. This iterative refinement may improve evaluation quality.

Conversely, in the LLMBar, the worst-performing approach is \textit{None}.
For 6 out of 8 evaluation models, \textit{None} results in the lowest performance.
In the InFoBench, Baseline shows the lowest correlation for half of the evaluation models.
These findings suggest that using any checklist benefits pairwise comparison evaluation, whereas the choice of generation method requires more careful consideration for direct scoring evaluation.

\subsection{Model Sizes}
Finally, we analyze how much correlation with human evaluation improves when small evaluator models use checklists.
In the direct scoring dataset, we observe that increasing checklist usage (i.e., lowering the threshold $k$) contributes to higher correlation with human ratings for some models, such as Gemma-2-27b-it and Mistral-8B-it.
In contrast, for other models—including both larger and smaller ones—checklist application does not substantially affect correlation, suggesting a limited contribution to alignment with human evaluation.
In the pairwise comparison dataset, checklists only slightly improve the accuracy of small models, indicating limited usefulness since evaluators may already implicitly consider checklist elements.

\section{Analysis}
To investigate RQ3: \textit{Which checklist items contribute to alignment with human evaluation?}, We conduct ablation and qualitative analyses to identify factors affecting evaluation performance.

\subsection{Ablation on Checklist Effectiveness}
\paragraph{Experimental Setup}
We define two types of checklist items.
A \textit{positive} item is one whose removal from the checklist leads to a \textit{decrease} in correlation with human evaluation, while a \textit{negative} item is one whose removal leads to an \textit{increase} in correlation. 
These definitions indicate whether the presence of a checklist item contributes to or hinders alignment with human evaluation.
Since ablating each individual checklist item is computationally expensive, we adopt a two-step approach.
In the first step, we classify each checklist--not individual items--based on whether its use improves alignment with human evaluation.
To measure alignment, we define a score $\Delta \bar{s}_{\text{all}}$ as:
\begin{multline}
\Delta \bar{s}_{\text{all}} = 
|\bar{s}_{\text{gold}} - \bar{s}_{\text{none}}|
- |\bar{s}_{\text{gold}} - \bar{s}_{\text{all}}|
\end{multline}
\noindent
where $ \bar{s}_{\text{gold}} $ represents the mean score given by human annotators, while $ \bar{s}_{\text{all}} $ and $ \bar{s}_{\text{none}} $ are mean scores from the model  (Qwen2.5-7B-it) with and without checklists, respectively. 
Based on $\Delta \bar{s}_{\text{all}}$ and predefined thresholds, we classify each checklist as \textit{positive} (it improves alignment with human evaluations) or \textit{negative} (it reduces alignment). 

In the second step,  we analyze individual checklist items within each group.
To quantify the contribution of each item, we define another score:
\begin{multline}
\Delta \bar{s}_{\text{abl}} =
|\bar{s}_{\text{gold}} - \bar{s}_{\text{all}}|
- |\bar{s}_{\text{gold}} - \bar{s}_{\text{abl}}|
\end{multline}
\noindent
where $ \bar{s}_{\text{abl}} $ is the mean evaluation score after removing a specific checklist item.
 If the removal of an item leads to lower alignment, it is classified as a \textit{positive} checklist item; if it leads to higher alignment, it is classified as a \textit{negative} one.
For details on the classification and analysis of checklist items, see Appendix~\ref{Checklist Classification} and~\ref{Ablation Checklist Items}.

\begin{figure}[t]
    \centering
    \includegraphics[width=\linewidth]{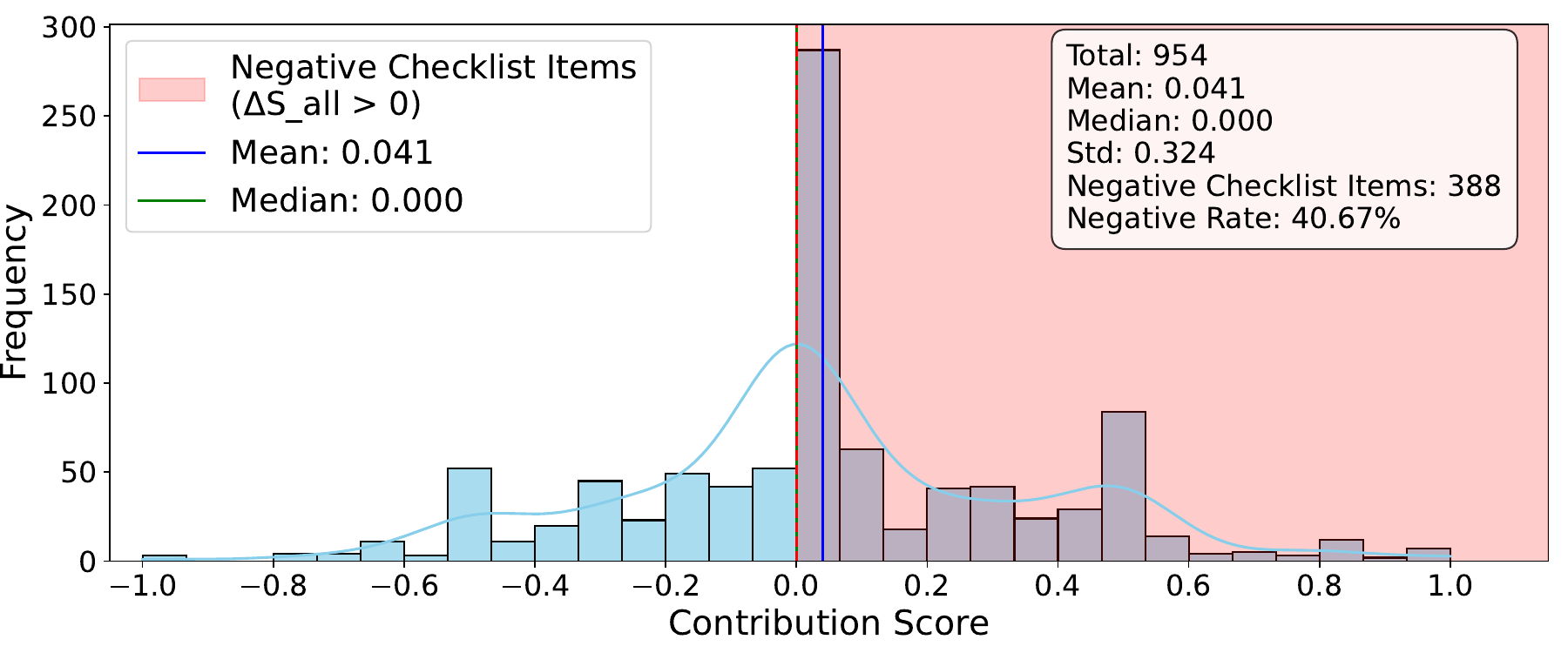}

    \caption{\textit{Negative} checklists ablation results on LLMBar dataset. Each plot shows how removing checklist items impacts correlation with human evaluation. \textit{Negative checklist items} ($\Delta \bar{s}_{\text{abl}}$) are highlighted in red. About 40\% of checklist items fall in the \textit{negative} region.}
    \label{LLMBar_ablation_negative_baseline}
\end{figure}
\paragraph{Quantitative Results}
We first report the classification result of the checklist items contained in generated checklists for the LLMBar and the InFoBench.
In the LLMBar, 53.5\% of the checklists used in \textit{positive checklists} are classified as \textit{positive checklist items} (1,079 out of 2,018), while 42.0\% of the items in \textit{negative checklists} are classified as \textit{negative checklist items} (3,847 out of 9,157).
In the InFoBench, all items in \textit{positive checklists} are classified as \textit{positive checklist items} (56 out of 56), while 40.7\% of the items in \textit{negative checklists} are classified as \textit{negative} (599 out of 1,472).
We then examine the impact of the \textit{negative checklist items} on alignment with human evaluations, as shown in Figure~\ref{LLMBar_ablation_negative_baseline},
using results generated by the Baseline.
These figures show how removing such items affects evaluation scores compared to using all checklist items.
For a comprehensive view of the effects across different checklist types and generation methods, including \textit{positive} and \textit{negative} items, refer to Appendix~\ref{Results of Checklists After Ablation}.

These results indicate that a substantial portion of the generated checklist items in \textit{negative checklists} contribute to reduced alignment with human evaluations.
However, the impact of such negative items-measured by the change in $\Delta \bar{s}_{\text{abl}}$ is generally small, suggesting that they do not significantly degrade evaluation quality even when present.

\begin{figure*}[t]
    \centering
    \small
    \includegraphics[width=1\textwidth]
    {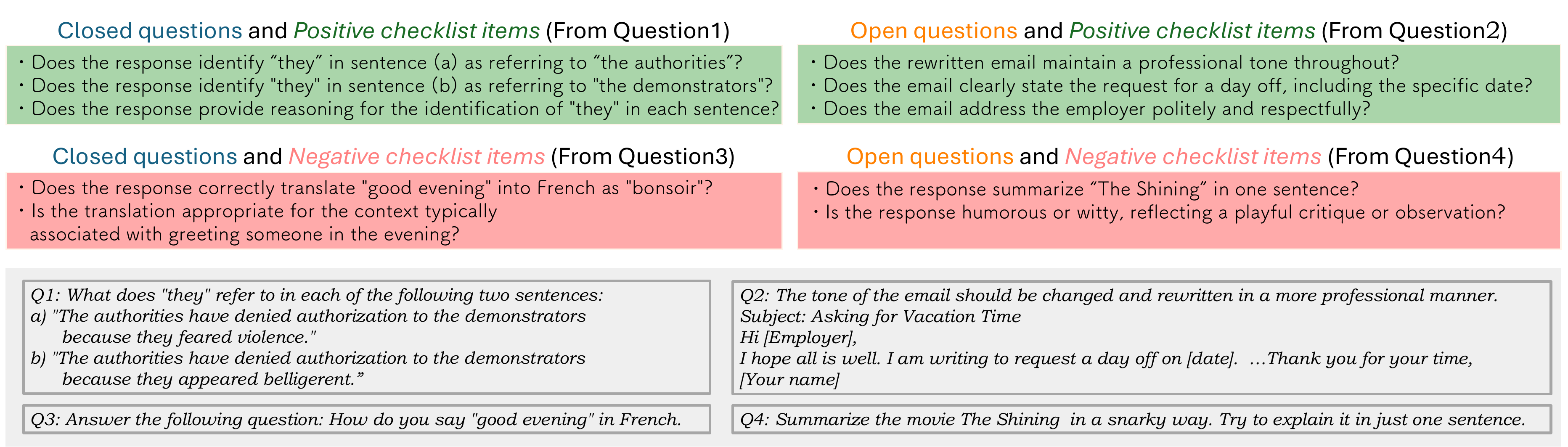} 
    \caption{Examples of \textit{positive} and \textit{negative} checklist items by question type (\textit{open} vs. \textit{closed}).}
    \label{fig:checklist_items}
\end{figure*}
\begin{table}[t]
\centering \small
\begin{tabular}{lcc}
\toprule
Dataset & Open & Closed\\ 
\midrule
LLMBar    & 75 & 10 \\ 
InFoBench & 43 & 7 \\ 
\bottomrule
\end{tabular}

\caption{Open vs. closed classification of questions. 
the LLMBar questions are sampled as 10\% from each of the 8 subsets (85 in total), 
while all 50 questions from the InFoBench are classified. }
\label{tb:classificate_question}
\end{table}
\begin{table}[t]
\centering \small
\begin{tabular}{ccccccc}
\toprule
\multirow{2}{*}{\# Label}& \multicolumn{3}{c}{Positive} & \multicolumn{3}{c}{Negative} \\
\cmidrule{2-4} \cmidrule{5-7} 
 & B & H & G & B & H & G \\
\midrule
H & 17 & 3 & - & 50 & 46 & - \\
% \midrule
G & 29 & - & 27 & 75 & - & 27 \\
\bottomrule
\end{tabular}

\caption{Checklist quality comparison in the InFoBench. We analyze 274 items (116 human-written, 158 generated).
B, H, and G denote items appearing in both, only human, and only generated checklists, respectively.
More than half of the items appear in both sets, indicating a notable overlap between human and generated checklist items.}
\label{tb:assessing_consistency_with_human_checklist}
\end{table}

\subsection{Open vs. Closed Question Classification}
\label{Open vs. Closed Question Classification }
We hypothesize that the effectiveness of checklists may depend on the type of question: closed questions may yield more consistent evaluations, whereas open-ended ones can introduce greater variability.
To explore this, we manually classify questions in each dataset based on whether their responses tend to converge (\textit{closed}) or diverge (\textit{open}).
For the LLMBar, we sample 10\% of questions from each of its eight subsets, while for the InFoBench, we analyze all 50 questions.
Table~\ref{tb:classificate_question} shows the classification results.
In both datasets, \textit{open} questions outnumber \textit{closed} ones, suggesting that subjective or ambiguous questions are more prevalent.
Such questions are more likely to lead to unstable evaluation outcomes and lower agreement with human evaluations, even when using checklists.

\subsection{Overlap Analysis of Human and Generated Checklists in the InFoBench}
\label{Overlap Analysis of Human and AI-Generated Checklists in the InFoBench}
We manually check the InFoBench checklist items to analyze the extent of overlap between human-written and generated items (274 in total: 116 human-written, 158 generated).
Over half of the items appear in both sets, indicating substantial overlap.
Here, \textit{both} means checklist items that are semantically equivalent and correspond to the same question.
For instance, the generated checklist item \textit{``Does the letter have approximately 250 words?''} closely corresponds to the human-written item \textit{``Is the generated recommendation letter around 250 words? (Output Attribute)''}.
Table~\ref{tb:assessing_consistency_with_human_checklist} summarizes the distribution of items by checklist type.
Items unique to the generated set—those without overlap with human-written items—often reflect additional perspectives or considerations that, while not explicitly stated in the question, are important for evaluating the response.
In contrast, checklist items exclusive to the human-written set tend to focus more on verifying the output format.

\subsection{Human Annotation of Checklist Items}
\label{Classification of Checklist Items}
To identify checklist item characteristics affecting alignment with human evaluations, we qualitatively analyze 293 items, including 89 \textit{positive} and 102 \textit{negative} items (see Appendix~\ref{Threshold Determination for Qualitative Analysis of Checklist Effectiveness}). We define six functional labels for \textit{positive} items and four for \textit{negative} ones.
Representative examples of generated checklist items are shown in Figure~\ref{fig:checklist_items}.
Our annotation shows that 60\% of \textit{positive} items explicitly reflect key question elements, aligning with essential response components, while about 30\% capture important evaluative aspects not explicitly mentioned. For \textit{negative} items, around 10\% fail to adequately address response content, suggesting room for improvement; however, over 85\% are consistent and deemed usable upon manual review. Furthermore, 77\% of these \textit{non-negative} items overlap with criteria created by human-written checklists (see Appendix~\ref{In-Depth Qualitative Analysis of Generated Checklist Items} for details).

\subsection{Analysis by Checklist Generation Policy}
We also analyze checklist generation policies to examine their characteristics.
For example, when handling mathematical problems like ``Compute the derivative of $2x^2 + 5x$'', the Baseline method generates checklists that break down elements into individual items, such as ``Does the response correctly apply the power rule to compute the derivative of $2x^2$?''．
For this problem, we observe little difference in the generated checklists among the Checklist Length * 0.5, Length * 1.5,  and Self-refine methods.
In contrast, the Specify method can generate more specific check items that include correct responses while maintaining itemization similar to the Baseline, such as ``Did the response simplify the derivative correctly to $4x + 5$?''
We also find that different generation methods generate similar checklists.
For example, when asking to explain machine learning and its types—supervised, unsupervised, and reinforcement learning—with real-world examples, all methods generate similar items checking basic elements, such as ``Does the response elaborate on the differences between supervised, unsupervised, and reinforcement learning?''.
In contrast, for the task asks how to increase productivity while working from home,
the Baseline generates abstract items, such as ``Are the suggestions in the response actionable and clear?'', while the Checklist Length * 1.5  includes more specific requirements, such as ``Is there guidance on setting goals or prioritizing tasks while working from home?''.
% We also find that different generation methods generate similar checklists.
% When asking to explain machine learning and its types—supervised, unsupervised, and reinforcement learning—with real-world examples, all methods generate similar items checking basic elements, such as ``Does the response elaborate on the differences between supervised, unsupervised, and reinforcement learning?''.
% In contrast, different generation methods produce similar checklists for various tasks.  
% For instance, the task asks how to increase productivity while working from home.
% The Baseline generates abstract items, such as ``Are the suggestions in the response actionable and clear?'', while the Length * 1.5  includes more specific requirements, such as ``Is there guidance on setting goals or prioritizing tasks while working from home?''.
For detailed checklist examples, see Appendix~\ref{sec:generate_checklist_example}.

\paragraph{Discussion}
These findings suggest two key directions for future work.
First, human evaluations sometimes rely on checklist items with ambiguous criteria or unclear scoring methods that fail to accurately capture response quality. This highlights the need to improve the design of human evaluation protocols. Notably, even \textit{negative} checklist items often overlap with human-written ones, underscoring the difficulty of establishing clear evaluation standards. This observation aligns with \citet{hosking2024humanfeedbackgoldstandard}, who highlight inconsistencies and biases in human evaluations.
Second, the overlap between generated and human-written checklist items suggests that LLMs can produce reliable and interpretable ones. Combining such generated checklists with human evaluation could improve overall evaluation reliability, rather than relying exclusively on either human or automatic evaluation.

\section{Conclusion}
We investigate checklist usefulness by focusing on three key questions: determining whether a checklist is necessary for LLM evaluators, designing useful checklists, and analyzing which items are effective.
Our experiments show that checklists do not always improve evaluations, and even \textit{negative} items often overlap with human-written ones, revealing limitations in current human evaluations.
This highlights the need to reconsider what makes an ideal checklist item that effectively combines human insight and automatic methods, targets relevant criteria, and adapts to different responses. Future work should focus on improving checklist creation and evaluation practices to ensure more reliable and meaningful evaluations.

\clearpage
\section*{Limitations}

Despite the comprehensiveness of our study, several limitations should be acknowledged.
First, while our datasets encompass a diverse range of input tasks, we utilize only a single English dataset for both the pairwise comparison and direct scoring tasks. This constraint may limit the generalizability of our findings across different generative tasks and languages.
Second, although we design our checklist generation policies to ensure broad coverage of possible checklist generation methods, there may exist alternative methods that we have not considered, such as those explicitly based on predefined evaluation criteria.
Finally, while the models used in our experiments cover multiple families of LLMs, they may still be insufficient to fully capture the necessary features of current LLMs, potentially limiting the scope of our conclusions.

\section*{Acknowledgments}
The authors would like to thank the anonymous reviewers for their helpful comments. 
This work was supported by JST FOREST Grant Number JPMJFR232R.
In this work, we used the ``mdx: a platform for building data-empowered society''.
We thank Satoru Katsumata, Hiroaki Sugiyama, Yugo Murawaki, and Sadao Kurohashi for their constructive comments and suggestions that helped improve this paper.

\bibliography{references}

\appendix

\section{Appendix}
\label{sec:appendix}
\subsection{Checklist Statistics}
\label{checklist stats}
Tables~\ref{the LLMBar_stats} and~\ref{the InFoBench_stats} show the statistical metrics of checklists for the LLMBar and the InFoBench, respectively.
Also, if a checklist cannot be obtained due to an API error or a formatting issue in the response, we regenerate it up to three times.
To ensure meaningful comparisons, we filter the datasets to include only questions with consistent checklist counts across all evaluation instances.

\begin{table}[h]
\footnotesize
\centering
\tabcolsep 3pt
\renewcommand{\arraystretch}{1.3}
\begin{tabular}{lccccc}
\toprule
Variations& Min & Max & Ave & S.D & Sum \\ 
\midrule
Baseline & 1 & 19 & 4.63 & 1.51 & 3,488 \\ 
\rowcolor{gray!20} Ticking & 2 & 10 & 4.92 & 1.02 & 3,710\\ 
Specify & 1 & 16 & 5.05 & 1.54 & 3,807 \\ 
\rowcolor{gray!20} Length * 0.5 & 1 & 10 & 2.27 & 0.74 & 1,714\\ 
Length * 1.5 & 2 & 28 & 6.98 & 2.38 & 5,266 \\ 
\rowcolor{gray!20} Self-refine & 1 & 19 & 4.62 & 1.55 & 3,490\\ 
\bottomrule
\end{tabular}
\caption{Statistical breakdown of generated checklists for each version of the LLMBar.
We generate checklists for 754 inputs.}
\label{the LLMBar_stats}
\end{table}

\begin{table}[h]
\footnotesize
\centering
\tabcolsep 3pt
\renewcommand{\arraystretch}{1.3} 
\begin{tabular}{lccccc}
\toprule
Variations& Min & Max & Ave & S.D & Sum \\ 
\midrule
Baseline & 2 & 9 & 4.9 & 1.64 & 245 \\ 
\rowcolor{gray!20} Ticking & 3 & 9 & 5.3 & 1.32 & 265\\ 
Specify & 2 & 9 & 5.32 & 1.69 & 266 \\ 
\rowcolor{gray!20} Length * 0.5 & 1 & 4 & 2.46 & 0.85 & 123\\ 
Length * 1.5 & 3 & 14 & 7.34 & 2.53 & 367 \\ 
\rowcolor{gray!20} Self-refine & 2 & 9 & 4.88 & 1.65 & 244\\ 
\bottomrule
\end{tabular}

\caption{Statistical breakdown of generated checklists for each version of the InFoBench.
We generate checklists for 250 tasks.}
\label{the InFoBench_stats}
\end{table}
\begin{figure*}[t]
    \centering
    \includegraphics[width=1.0
    \textwidth, height=0.4\textheight, keepaspectratio]
{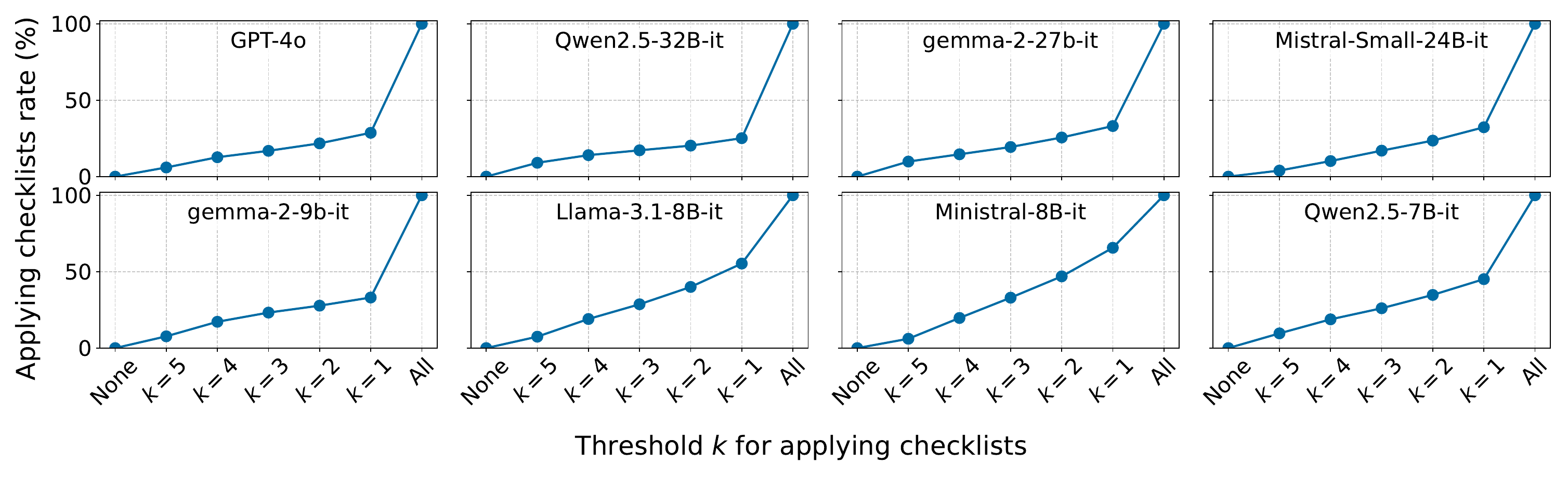} 
    \caption{
    Checklists application rate in the pairwise comparison (LLMBar).
    }
\label{fig:the LLMBar_checklist_rate}
\end{figure*}

\begin{figure*}[t]
    \centering
    \includegraphics[width=1.0
    \textwidth, height=0.4\textheight, keepaspectratio]
{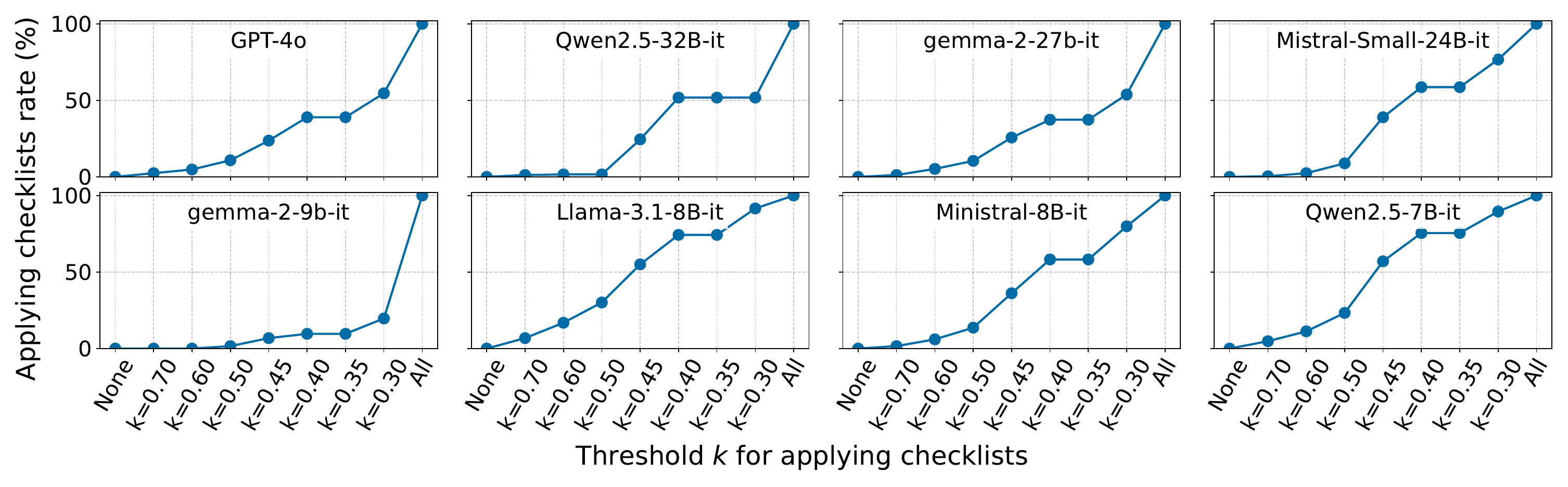} 
    \caption{
    Checklists application rate in the direct scoring (InFoBench).
    }
\label{fig:the InFoBench_checklist_rate}
\end{figure*}

\subsection{Checklists Application Rate}
\label{Checklists Application Rate}
Figures~\ref{fig:the LLMBar_checklist_rate} and \ref{fig:the InFoBench_checklist_rate} present the checklist application rates for different threshold values in pairwise comparison and direct scoring tasks, illustrating how the threshold $k$ influences the proportion of responses evaluated with checklists.

\subsection{Bootstrap Sampling}
\label{Bootstrap Sampling}
We conduct a bootstrap test to evaluate whether the selective application of checklists leads to improvements．
For the bootstrap procedure, we perform 1,000 resampling iterations, fix the random seed to 42, and determine statistical significance based on 95\% confidence intervals.

For pairwise comparison, we observe statistically significant differences in 20 out of 48 cases．
The detailed results for each model are presented below.
\begin{itemize}
\item GPT-4o: Shows statistically significant differences for all checklist-generation policies (6/6).
\item Qwen2.5-32B-it: Shows significant differences for Baseline, Length×0.5, and Self-refine (3/6).
\item Gemma-2-27B-it: Shows no significant difference for any checklist-generation policies (0/6).
\item Gemma-2-9B-it: Shows significant differences for all checklist-generation policies (6/6).
\item Qwen2.5-7B-it: Shows significant differences for all checklist-generation policies except Ticking (5/6).
\item The other three models (Mistral-Small-24B-It, Ministral-8B-It, and Llama-3.1-8B-It) do not show any significant differences (0/6 for each model).
\end{itemize}

\subsection{Checklist Retention Rates after Filtering}
\label{Question Retention Rates after Filtering}
\begin{figure*}[htbp]
    \centering
    % 1行目：2つの図
    \begin{minipage}{0.48\textwidth}
        \centering
        \includegraphics[width=\textwidth]{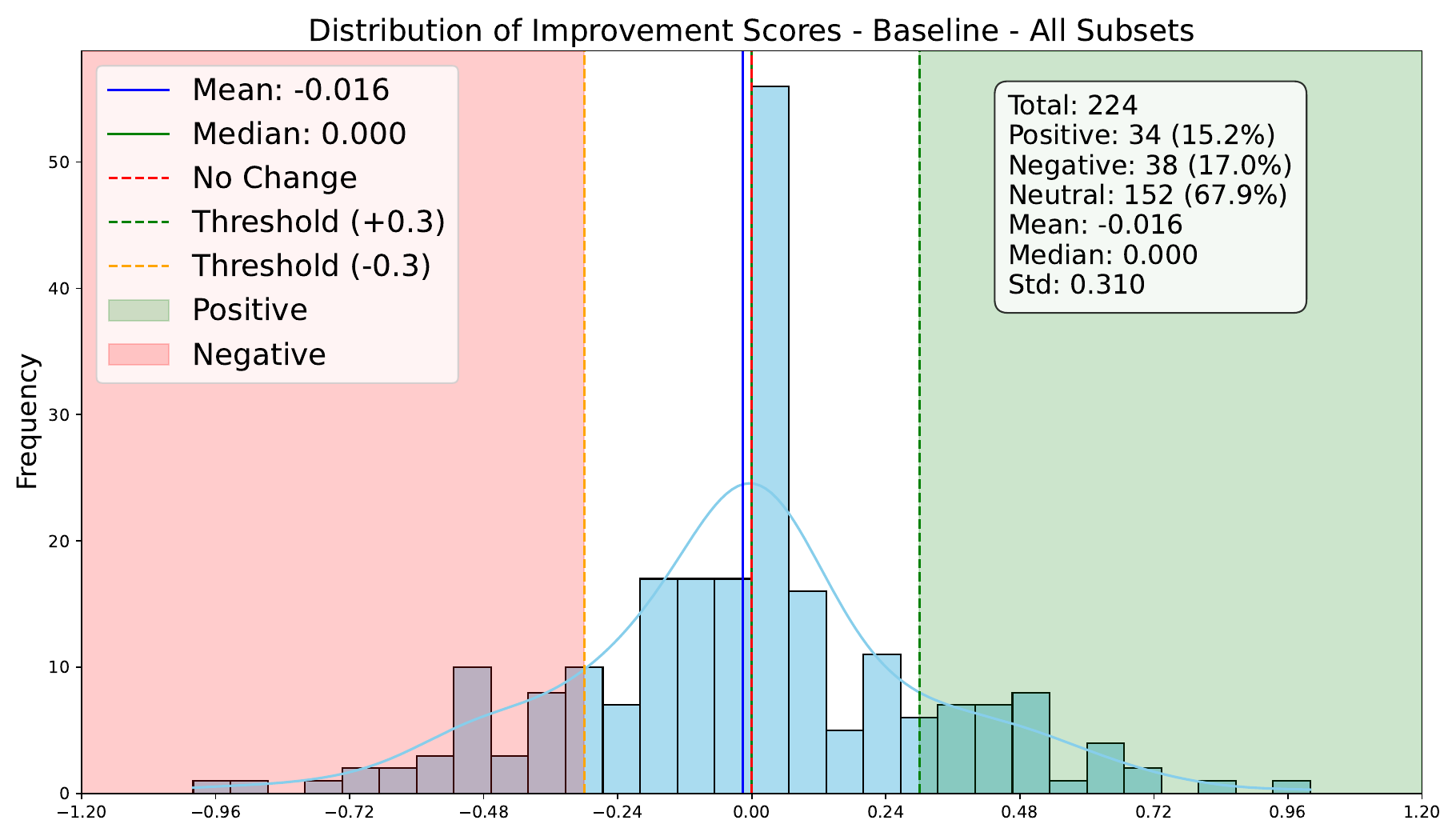}
    \end{minipage}\hfill
    \begin{minipage}{0.48\textwidth}
        \centering
        \includegraphics[width=\textwidth]{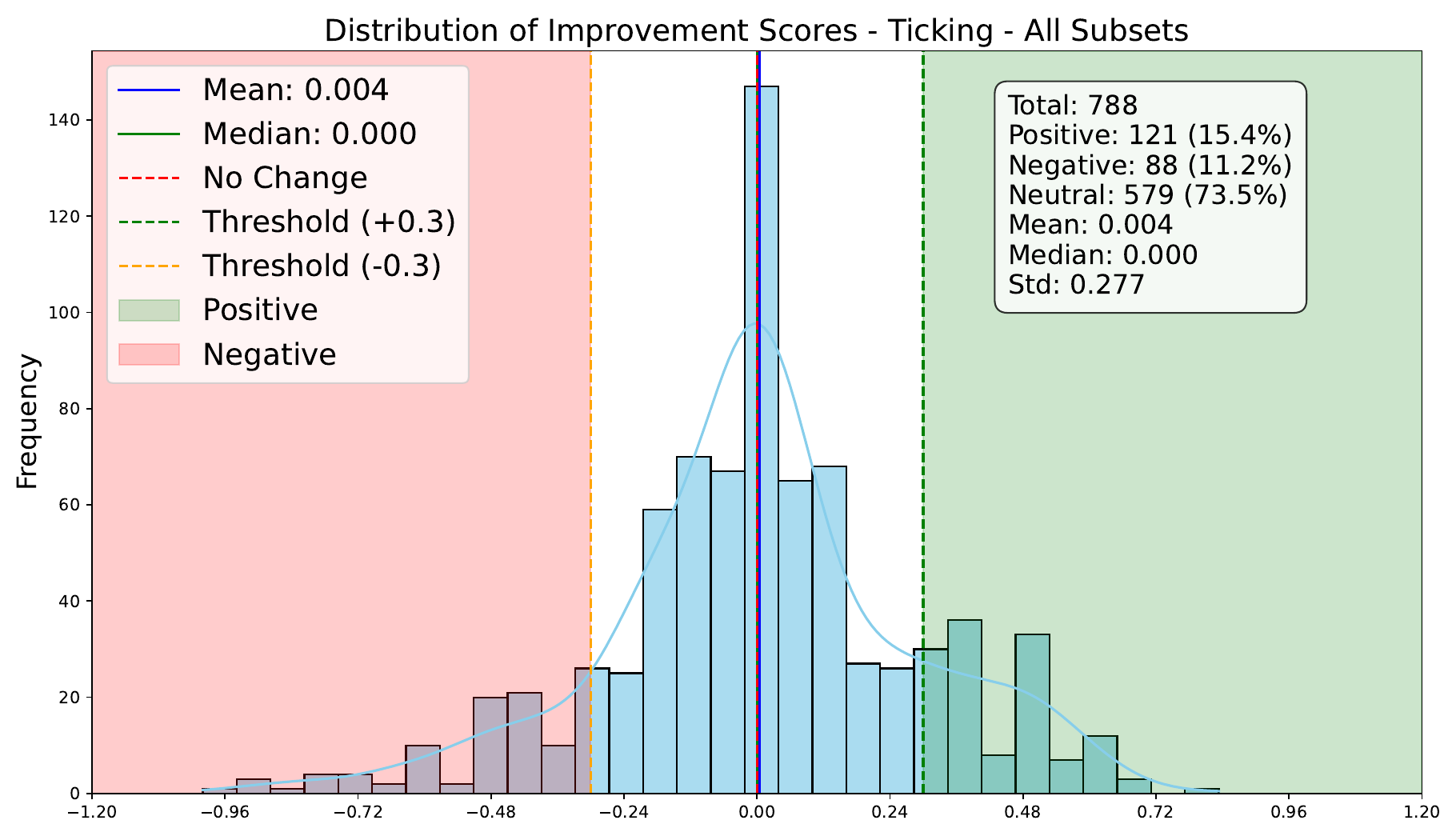}
    \end{minipage}

    \vskip 0.01em
\begin{minipage}{0.48\textwidth}
        \centering
        \includegraphics[width=\textwidth]{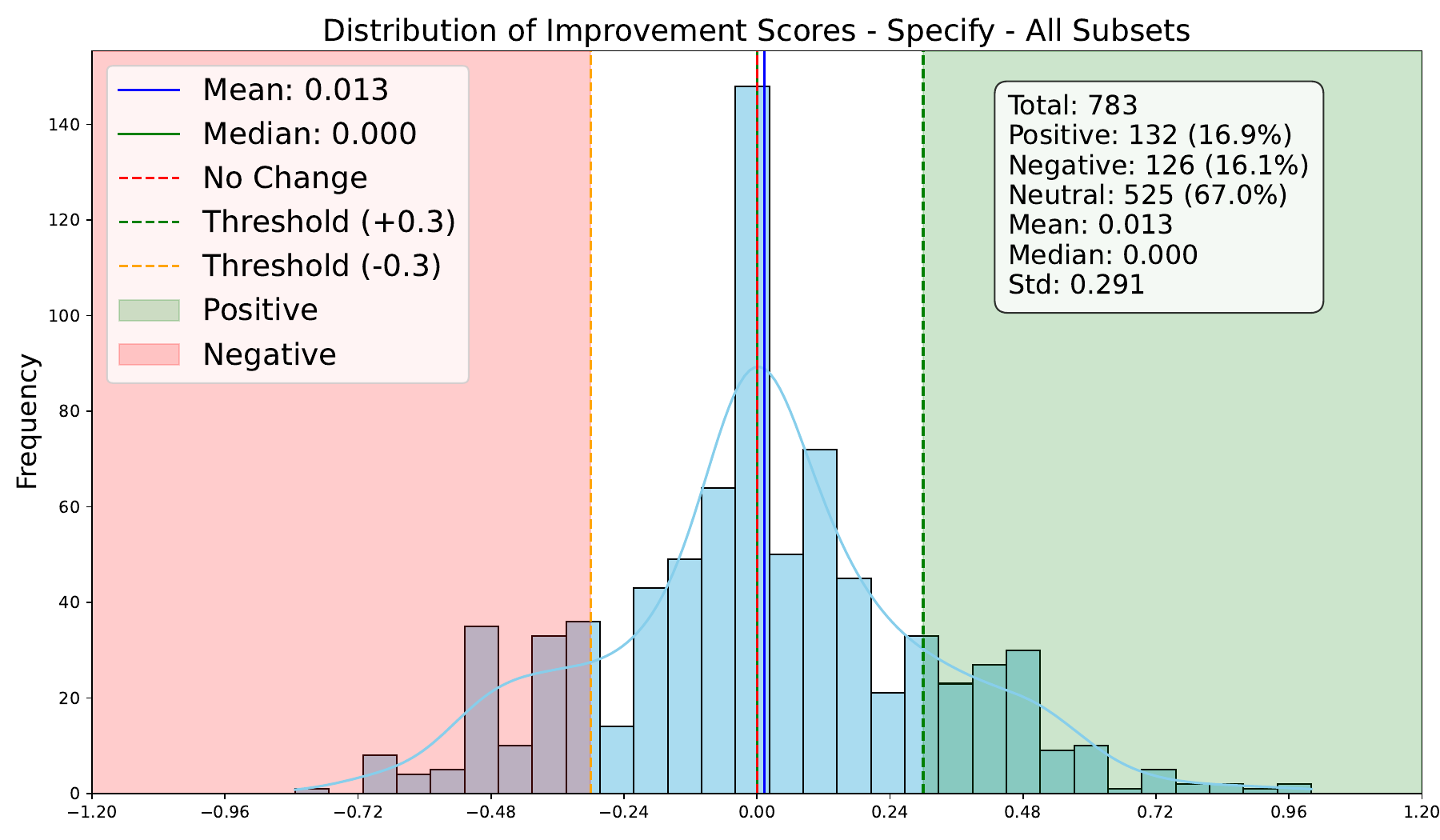}
    \end{minipage}\hfill
    \begin{minipage}{0.48\textwidth}
        \centering
        \includegraphics[width=\textwidth]{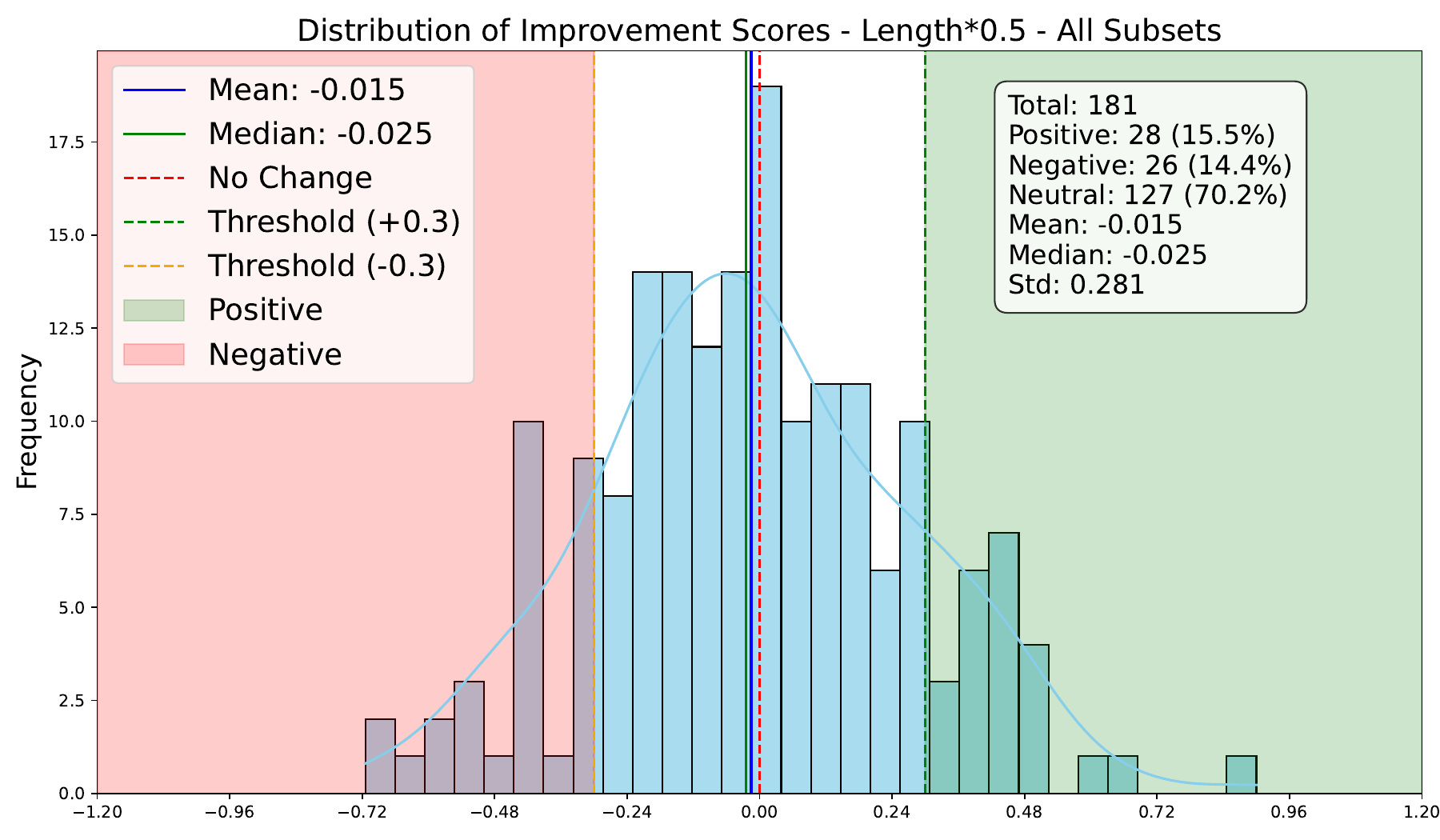}
    \end{minipage}
    
     \vskip 0.01em
    \begin{minipage}{0.48\textwidth}
        \centering
        \includegraphics[width=\textwidth]{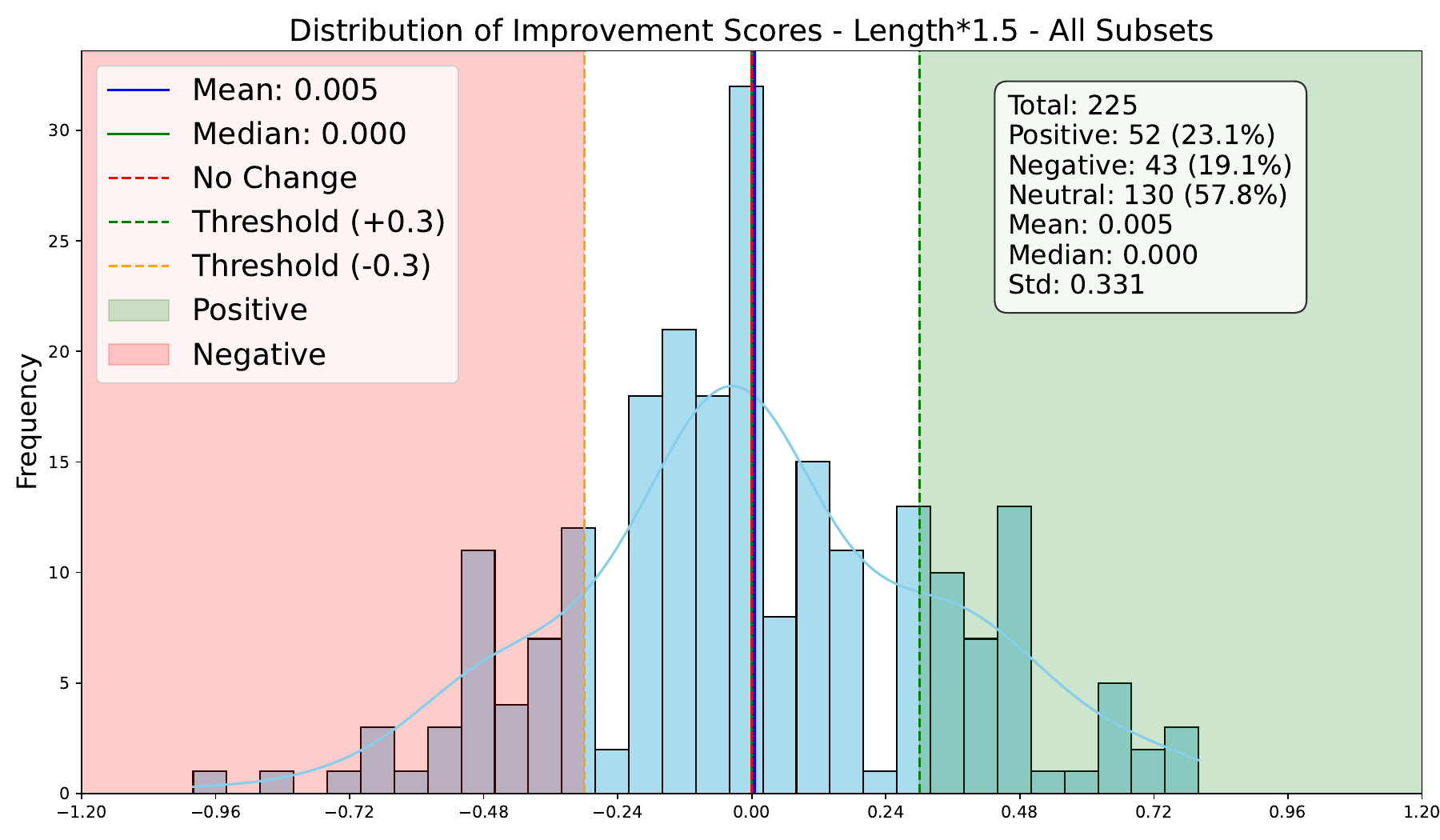}
    \end{minipage}\hfill
    \begin{minipage}{0.48\textwidth}
        \centering
        \includegraphics[width=\textwidth]{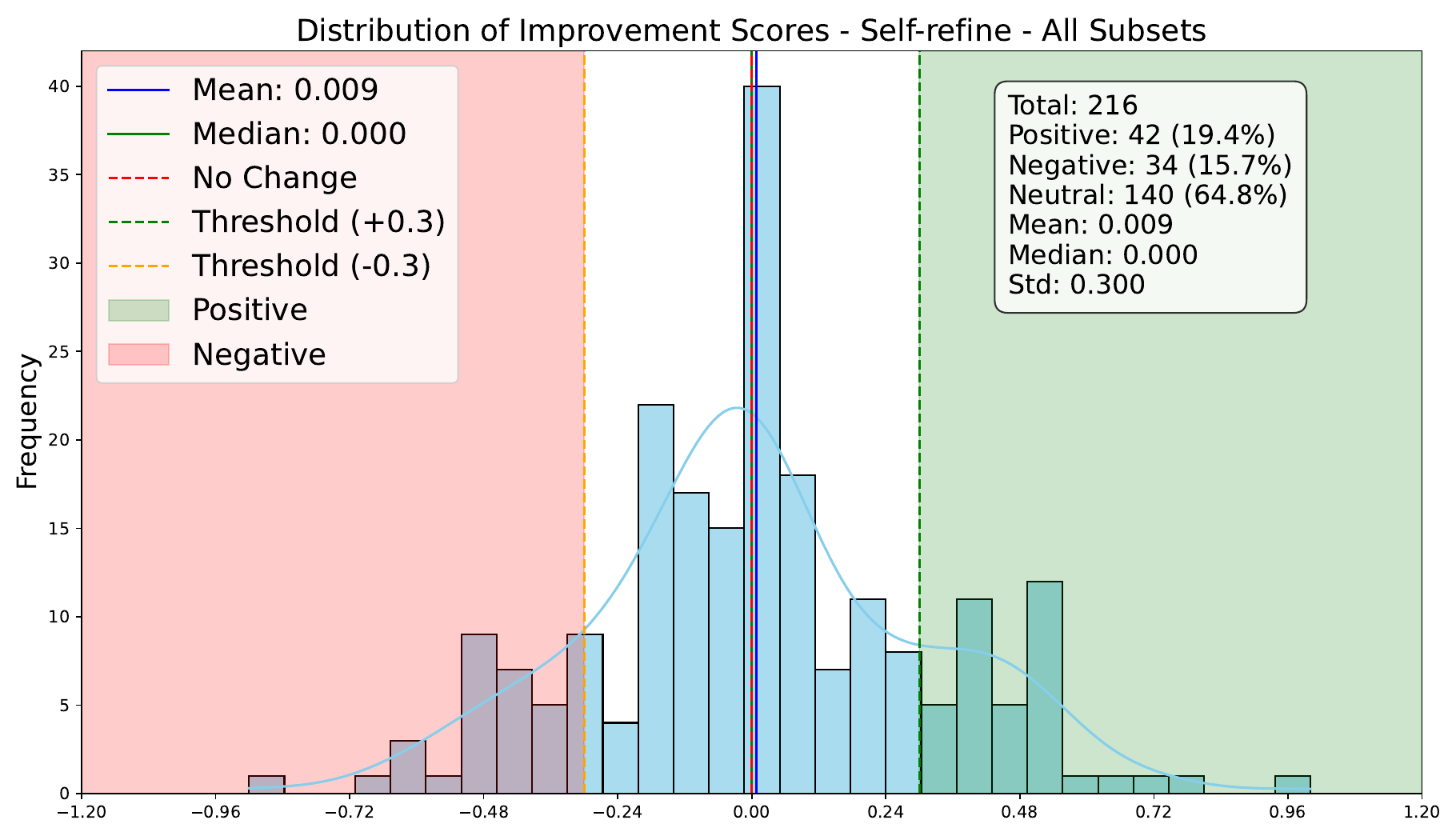}
    \end{minipage}
    \caption{
    This figure presents the improvement scores computed for LLMBar across different policies. For each policy, the positive and negative checklists respectively comprise approximately 20\% of the total items.\\
    The subfigures are arranged as follows: the top row shows the Baseline, Ticking, and Specify checklist policies; the bottom row shows the Length * 0.5, Length * 1.5, and Self-refine policies.
    }
    \label{LLMBar_Improvement_Score}
\end{figure*}

\begin{figure*}[htbp]
    \centering
    \begin{minipage}{0.48\textwidth}
        \centering
        \includegraphics[width=\textwidth]{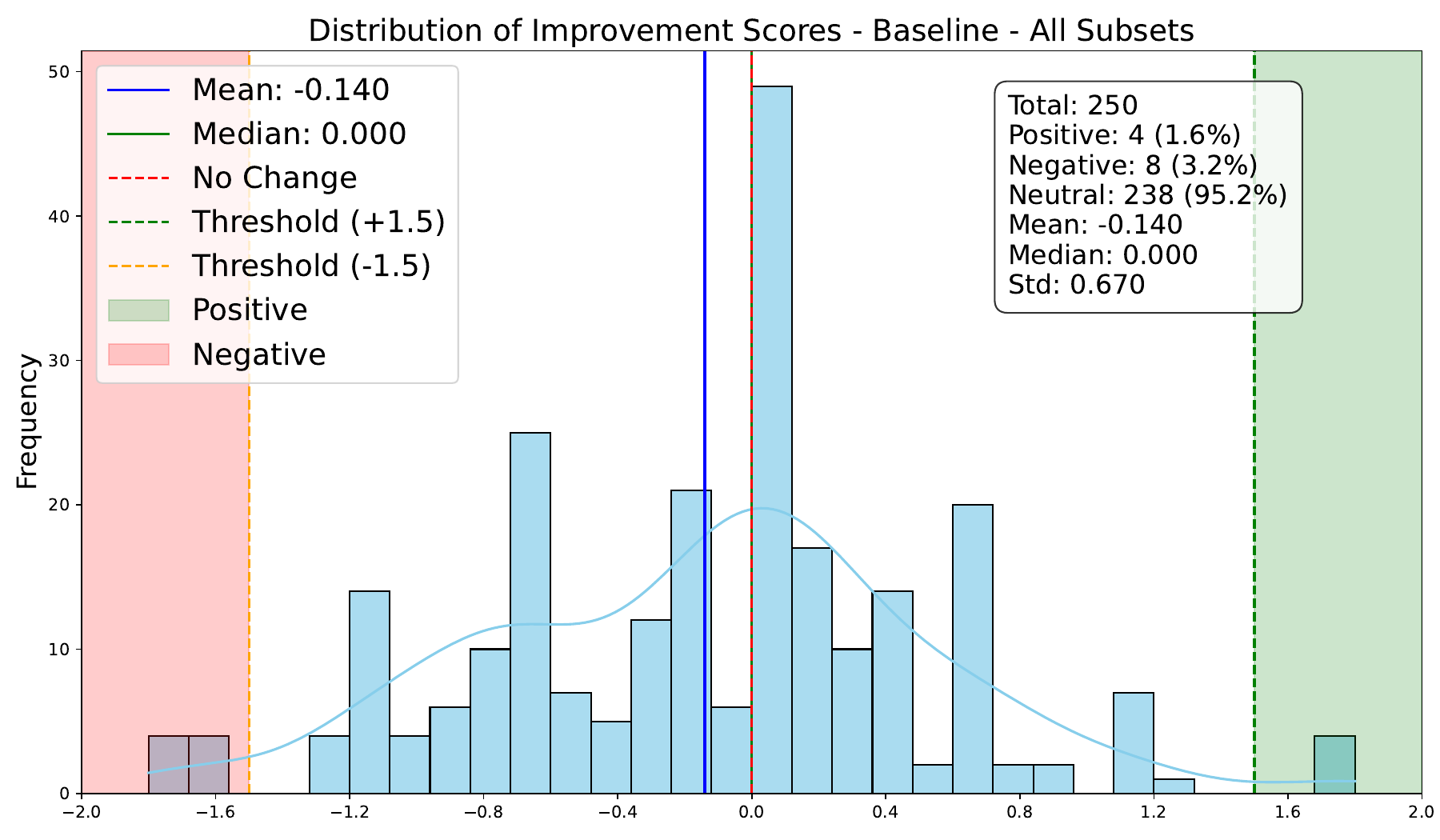}
    \end{minipage}\hfill
    \begin{minipage}{0.48\textwidth}
        \centering
        \includegraphics[width=\textwidth]{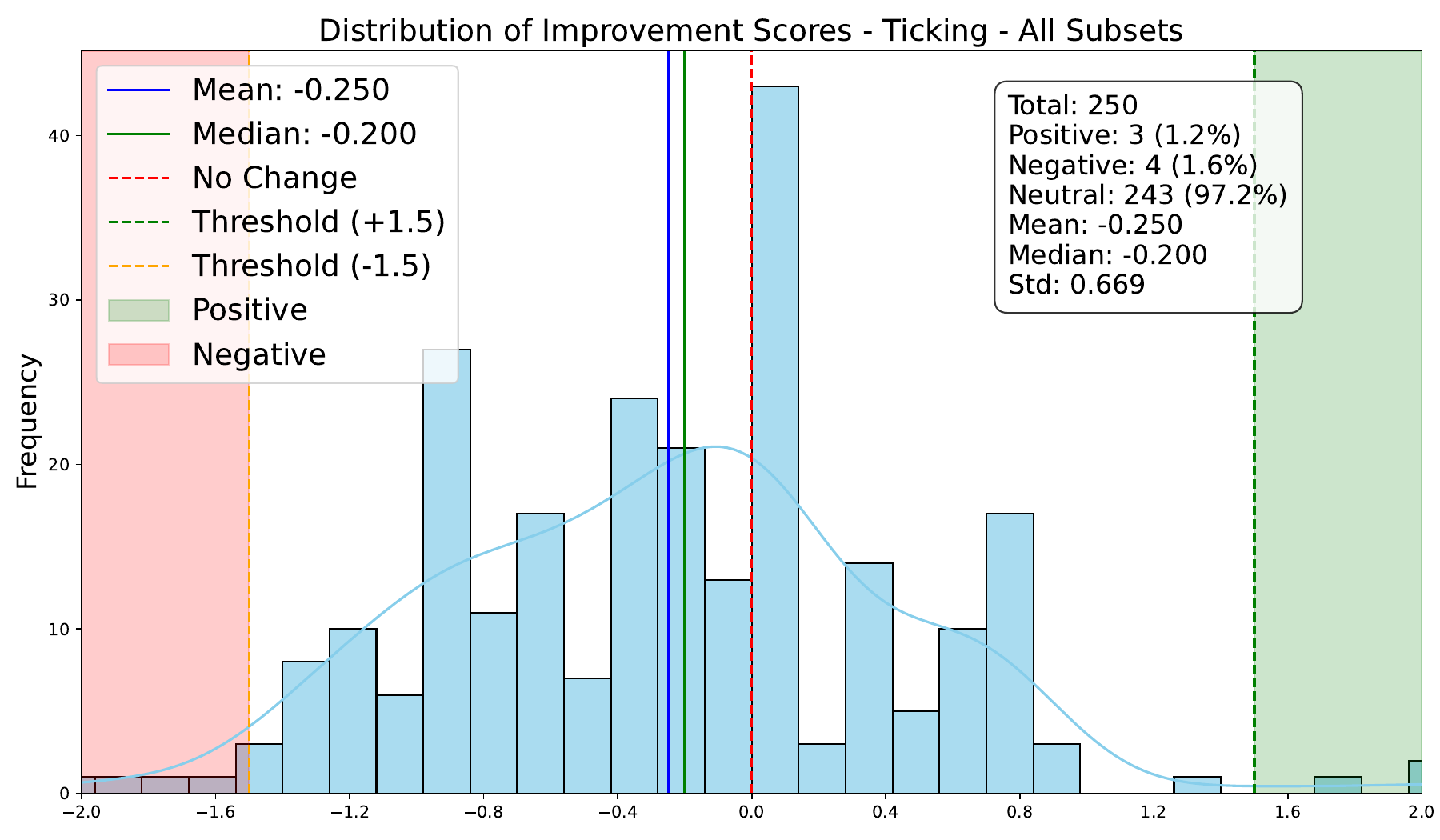}
    \end{minipage}
     \vskip 0.01em
    \begin{minipage}{0.48\textwidth}
        \centering
        \includegraphics[width=\textwidth]{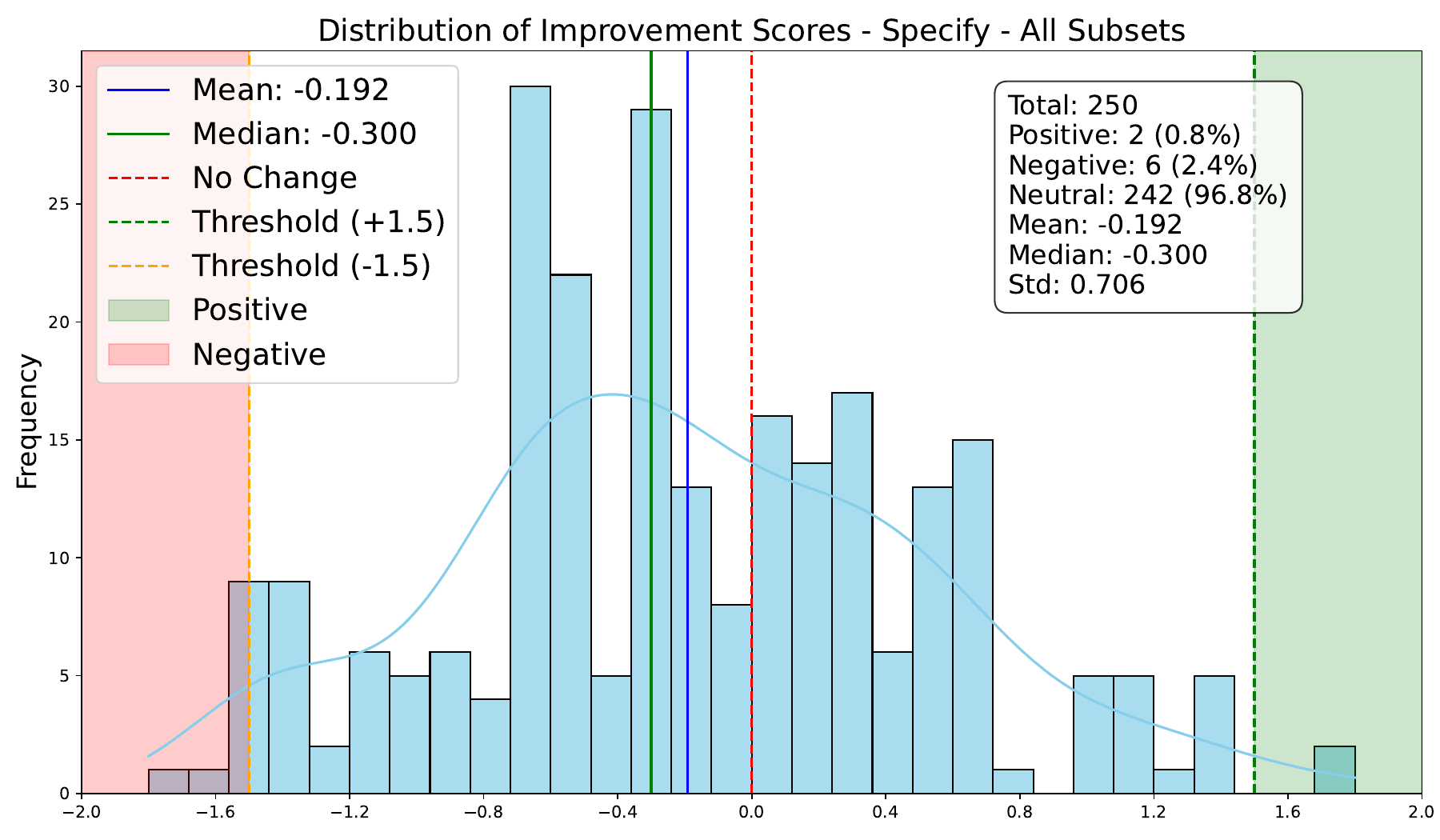}
    \end{minipage}\hfill
    \begin{minipage}{0.48\textwidth}
        \centering
        \includegraphics[width=\textwidth]{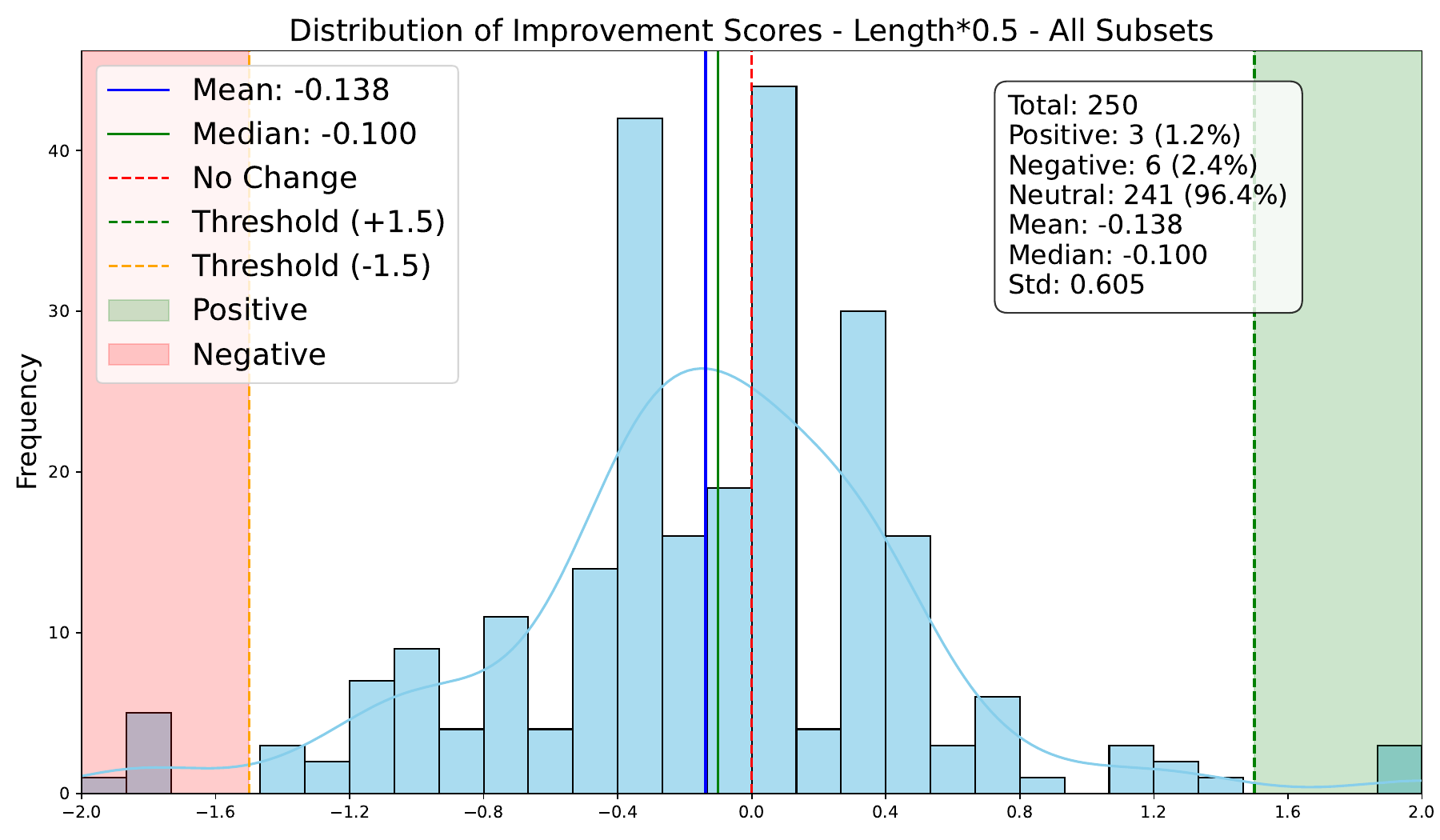}
    \end{minipage}

    \vskip 0.01em  
    \begin{minipage}{0.48\textwidth}
        \centering
        \includegraphics[width=\textwidth]{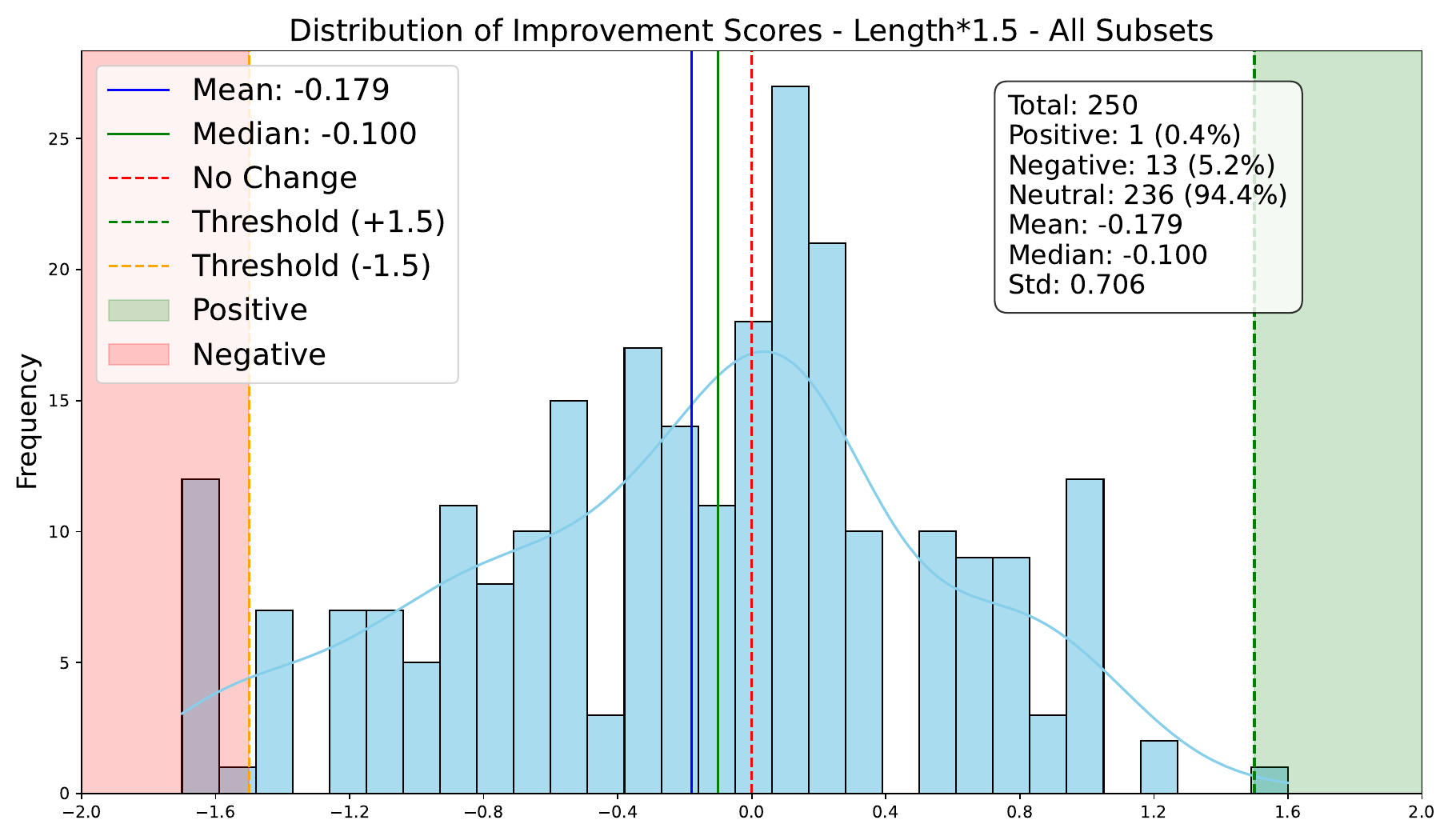}
    \end{minipage}\hfill
    \begin{minipage}{0.48\textwidth}
        \centering
        \includegraphics[width=\textwidth]{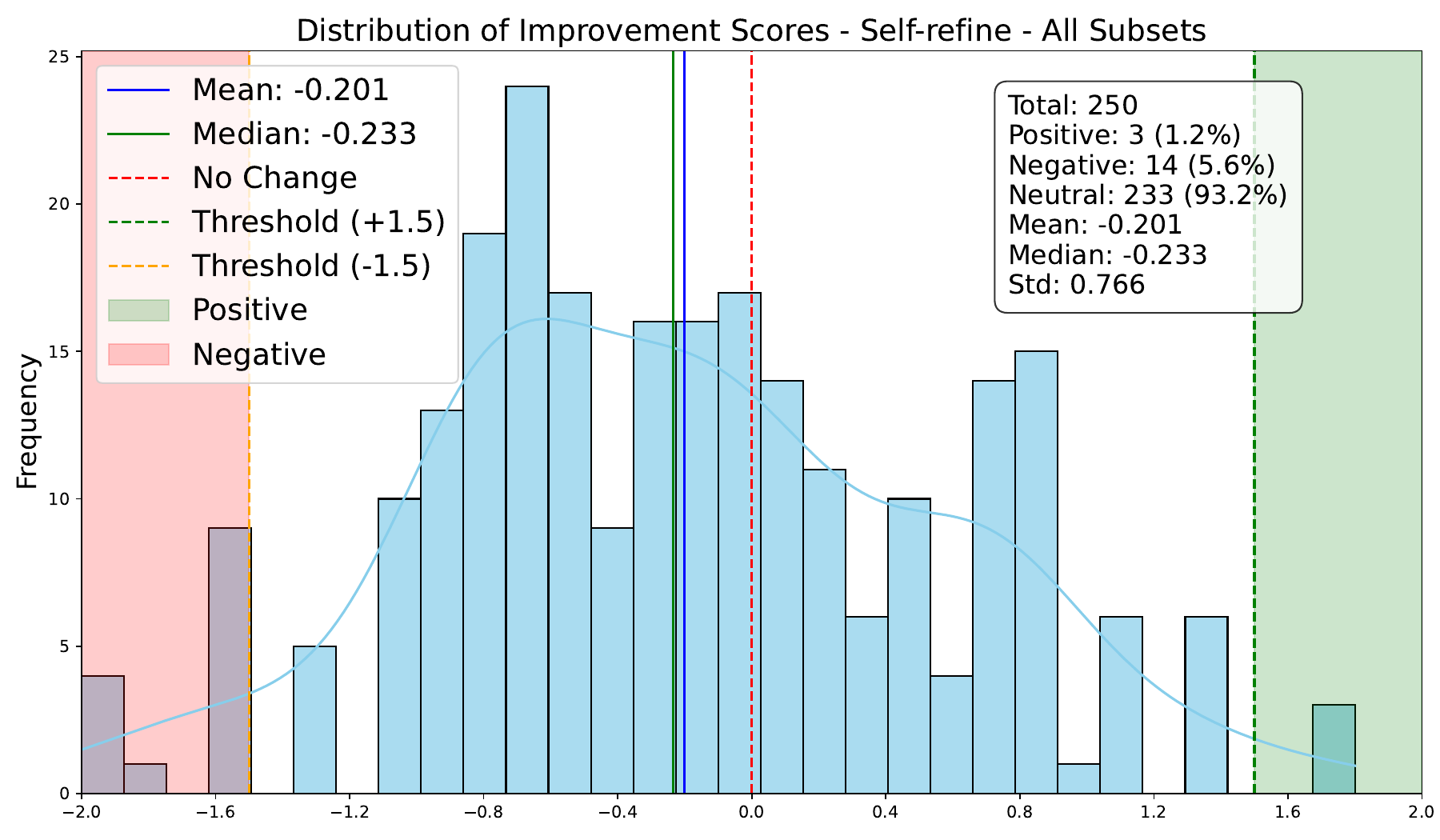}
    \end{minipage}

    \caption{
        This figure shows improvement scores calculated on InFoBench. For each evaluation policy, approximately 5\% of checklists are classified as positive or negative. \\
The subfigures are arranged as follows: the top row shows the Baseline, Ticking, and Specify checklist policies; the bottom row shows the Length * 0.5, Length * 1.5, and Self-refine policies.
}
    \label{InFoBench_Improvement_Score}
\end{figure*}

The filtering process results in different checklists retention rates across our datasets, as shown in Table~\ref{tab:checklist_reduction}.
In the LLMBar dataset, approximately 90\% of checklists are retained for the Ticking and Specify checklist policies, while other categories experience a significant reduction to around 25\% of the original checklist count.
In contrast, the InFoBench dataset maintains 100\% of checklists across all policies, indicating more consistent checklist application in this dataset.
A detailed breakdown of the checklist classification can be found in Figures~\ref{LLMBar_Improvement_Score} and~\ref{InFoBench_Improvement_Score}.
In the LLMBar dataset, both \textit{positive} and \textit{negative} checklists each constitute approximately 20\% of the total checklists, with the remainder falling into the \textit{neutral} category.
The InFoBench dataset shows a different distribution, with \textit{positive} and \textit{negative} checklists each representing only about 2\% across most checklist policies.
The Length * 1.5  and Self-refine policies stand out as exceptions, with negative checklists surpassing 5\% in these cases.

\subsection{Ablation Checklist}
\label{Ablation Checklist}
\subsubsection{Selecting Checklist for Ablation}
\label{Checklist Classification}

We determine which checklists to use for the ablation of checklist items.
To this end, we classify each checklist as:

\begin{itemize}
\item \textbf{Positive Checklists}: $\Delta \bar{s}_{\text{all}}$ $\geq$ threshold (significantly improves accuracy)
\item \textbf{Negative Checklists}: $\Delta \bar{s}_{\text{all}}$  $\leq -$threshold (significantly degrades accuracy)
\end{itemize}

We set different threshold values for each dataset: 0.3 for the pairwise comparison dataset and 1.5 for the direct scoring dataset. 
We then use the selected positive and negative checklists for checklist item ablation.

\subsubsection{Ablation Checklist Items}
\label{Ablation Checklist Items}
Based on $\Delta \bar{s}_{\text{abl}}$, we select the final positive and negative checklist items as follows:

\begin{itemize}
\item \textbf{Positive Checklist Item}: checklist item with $\Delta \bar{s}_{\text{abl}}$ < 0, indicating that removing this checklist item degrades performance compared to using all checklists.
\item \textbf{Negative Checklist Item}: checklist item with $\Delta \bar{s}_{\text{abl}}$ > 0, indicating that removing this checklist item improves performance compared to using all checklists.
\end{itemize}

\begin{table*}[h]
\centering
\small
\begin{tabular}{lcccccc}
\toprule
\textbf{Checklist Policy} & \textbf{Original} & \textbf{Verified} & \textbf{Reduced} & \textbf{Verified Ratio (\%)} & \textbf{Dataset} \\
\midrule
Baseline               & 3488 & 879  & 2609 & 25.2  & LLMBar \\
\rowcolor{gray!20} Ticking                & 3722 & 3365 & 357  & 90.4  & LLMBar \\
Specify                 & 3821 & 3467 & 354  & 90.7  & LLMBar \\
\rowcolor{gray!20} Length * 0.5   & 1714 & 391  & 1323 & 22.8  & LLMBar \\
Length * 1.5   & 5266 & 1349 & 3917 & 25.6  & LLMBar \\
\rowcolor{gray!20} Self-refine        & 3490 & 862  & 2628 & 24.7  & LLMBar \\
\midrule
Baseline               & 245  & 245  & 0    & 100   & InFoBench \\
\rowcolor{gray!20} Ticking                & 265  & 265  & 0    & 100   & InFoBench \\
Specify                 & 266  & 266  & 0    & 100   & InFoBench \\
\rowcolor{gray!20} Length * 0.5   & 123  & 123  & 0    & 100   & InFoBench \\
Length * 1.5   & 367  & 367  & 0    & 100   & InFoBench \\
\rowcolor{gray!20} Self-refine         & 244  & 244  & 0    & 100   & InFoBench \\
\bottomrule
\end{tabular}
\caption{Checklist verification results for each method on the LLMBar and the InFoBench, including the number of original, verified, and reduced checklists, and the reduced ratio for each checklist policy. The results show significant variation in checklist reduction, with Ticking and Specify methods achieving the highest reduction ratios on both datasets, while other methods like Length*0.5 show lower reductions.}
\label{tab:checklist_reduction}
\end{table*}

\subsubsection{Results of Checklists After Ablation}
\label{Results of Checklists After Ablation}
\begin{figure*}[htbp]
    \centering
    % 1行目：3つの図
    \begin{minipage}{0.48\textwidth}
        \centering
        \includegraphics[width=\textwidth]{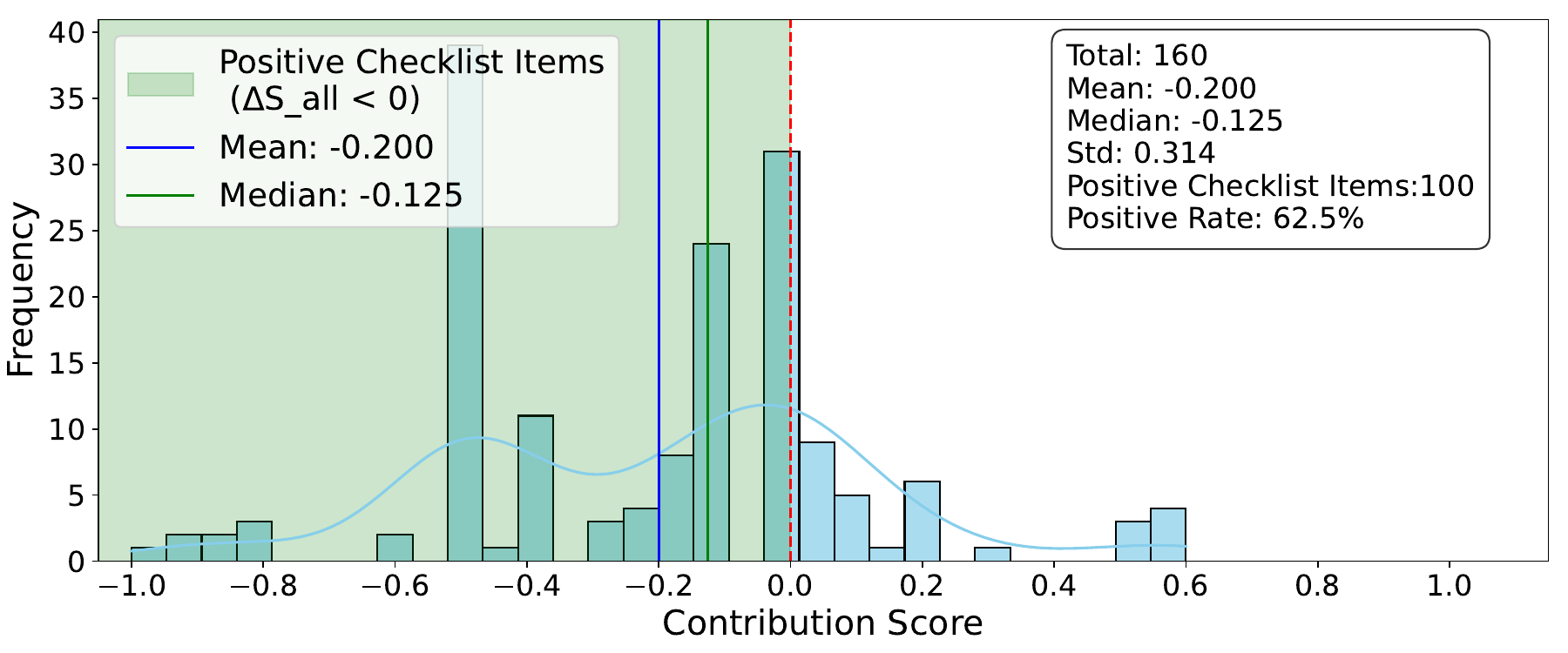}
    \end{minipage}\hfill
    \begin{minipage}{0.48\textwidth}
        \centering
        \includegraphics[width=\textwidth]{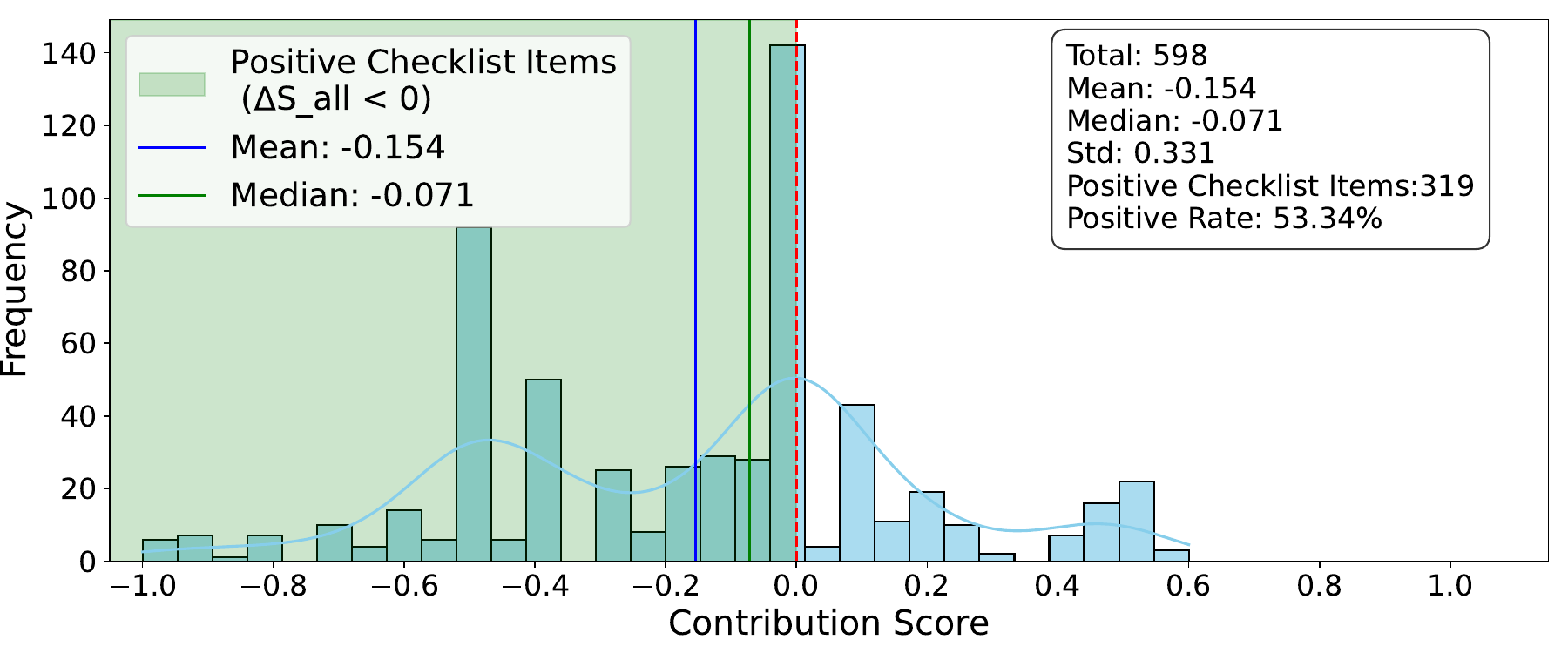}
    \end{minipage}
     \vskip 0.01em
    \begin{minipage}{0.48\textwidth}
        \centering
        \includegraphics[width=\textwidth]{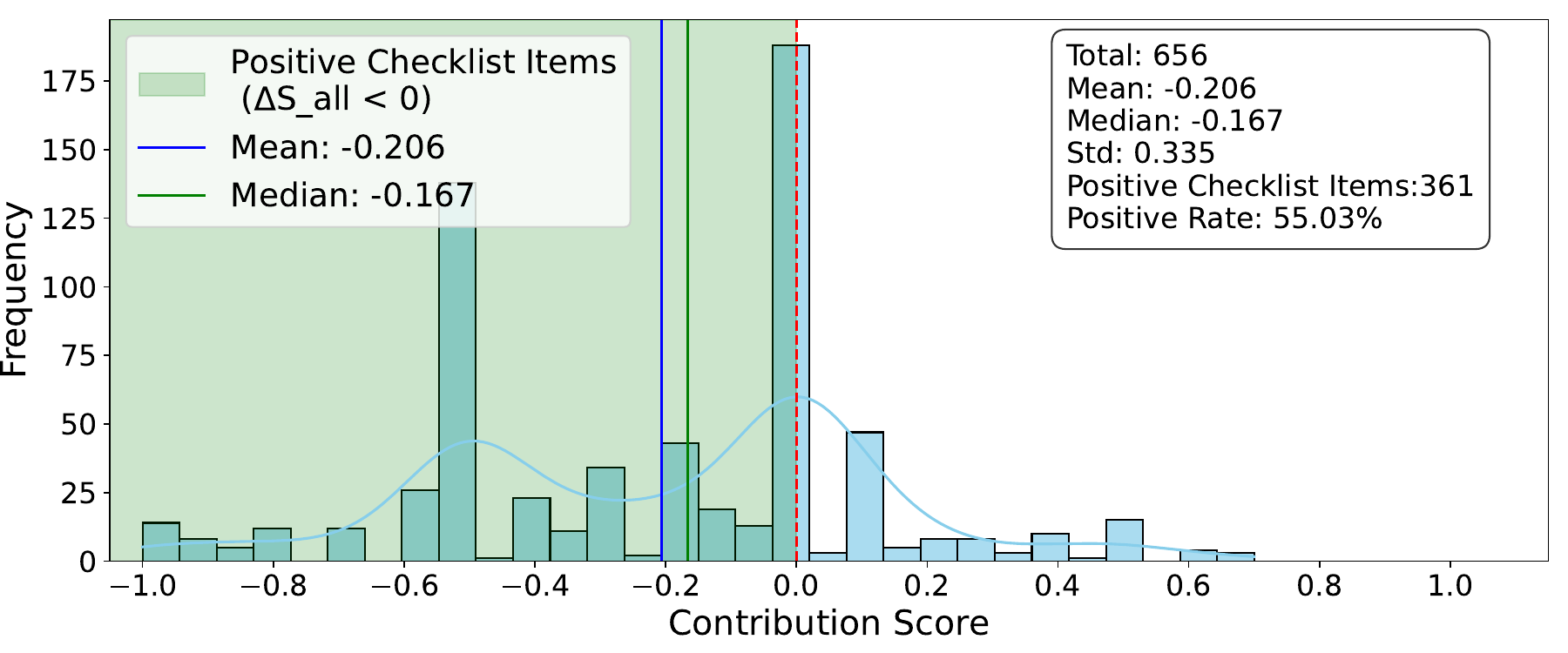}
    \end{minipage}\hfill
    \begin{minipage}{0.48\textwidth}
        \centering
        \includegraphics[width=\textwidth]{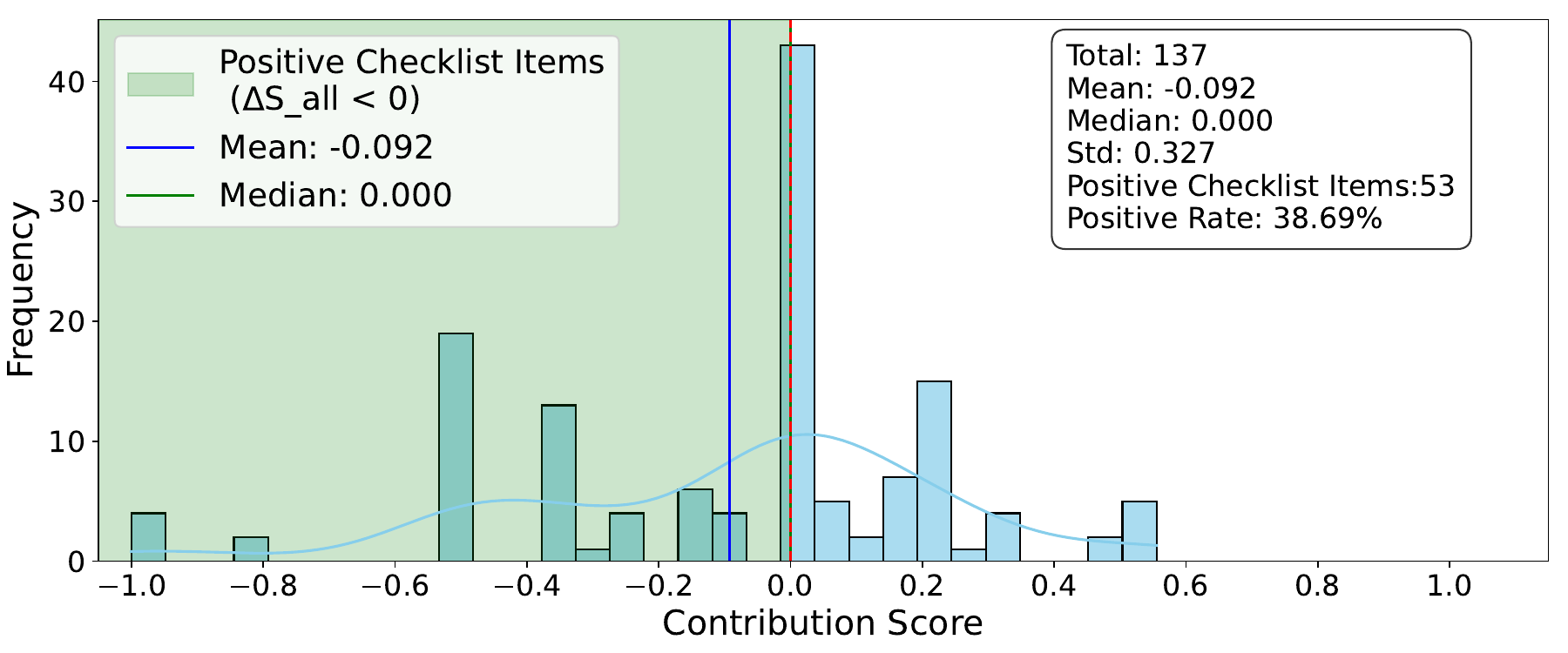}
    \end{minipage}
     \vskip 0.01em
    \begin{minipage}{0.48\textwidth}
        \centering
        \includegraphics[width=\textwidth]{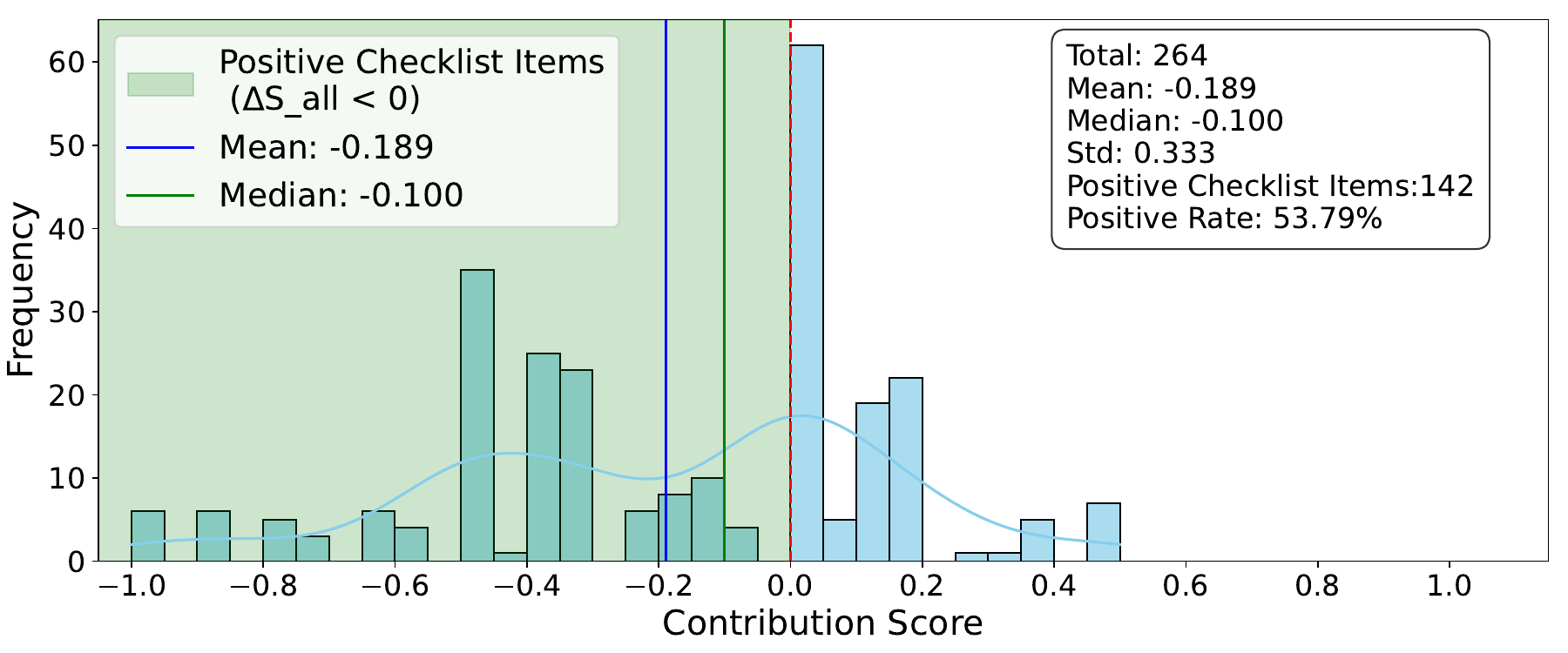}
    \end{minipage}\hfill
    \begin{minipage}{0.48\textwidth}
        \centering
        \includegraphics[width=\textwidth]{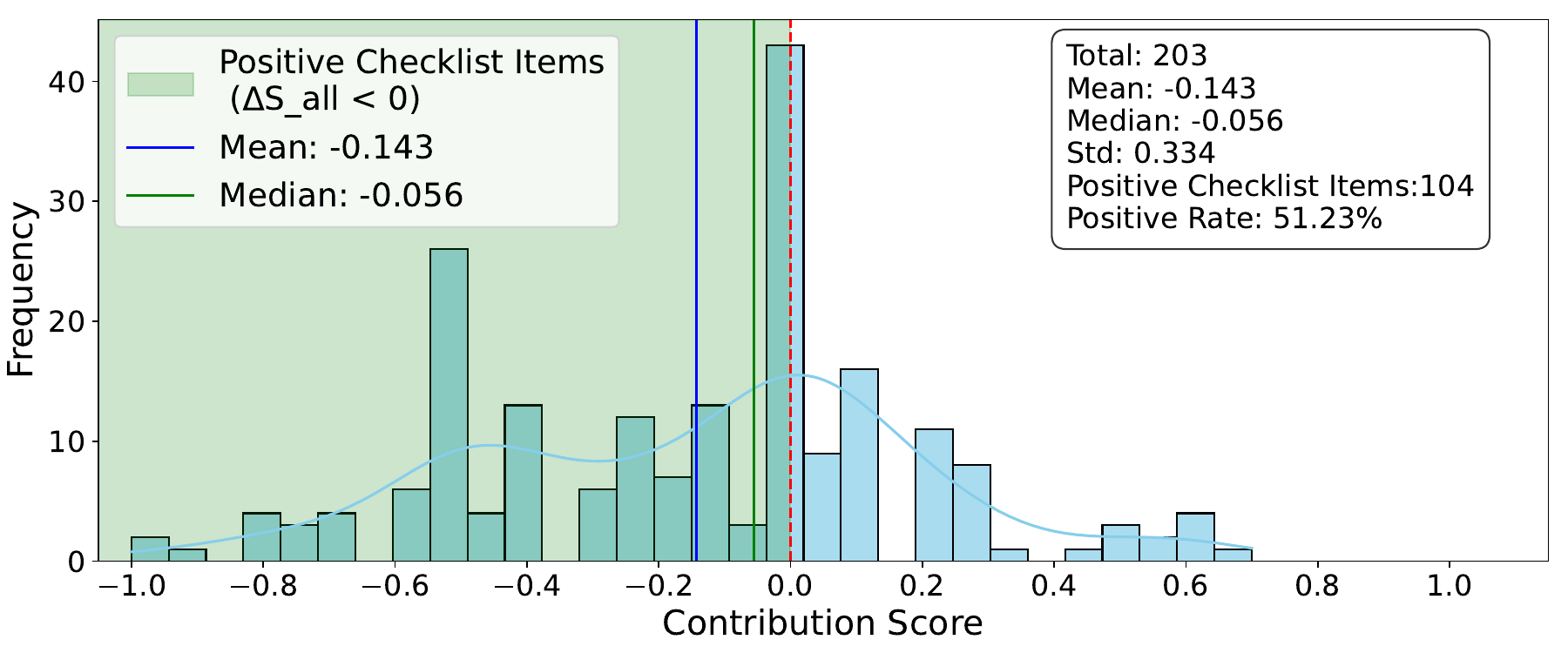}
    \end{minipage}

    \caption{Positive checklist ablation results on the LLMBar dataset. Each plot shows how removing checklist items impacts correlation with human evaluation. \textit{Positive} checklist items ($\Delta \bar{s}_{\text{abl}}$ < 0) are highlighted in green. Most scores lie between -0.1 and 0.1. The six generation policies—Baseline, Ticking, Specify, Length*0.5, Length*1.5, Self-Refine—are arranged top-left to bottom-right. All but Length0.5 have over 50\% \textit{positive} checklist items; Length0.5 falls below 40\%.
    }
    \label{LLMBar_ablation_positive_checklist}
\end{figure*}
\begin{figure*}[htbp]
    \centering
    \begin{minipage}{0.48\textwidth}
        \centering
        \includegraphics[width=\textwidth]{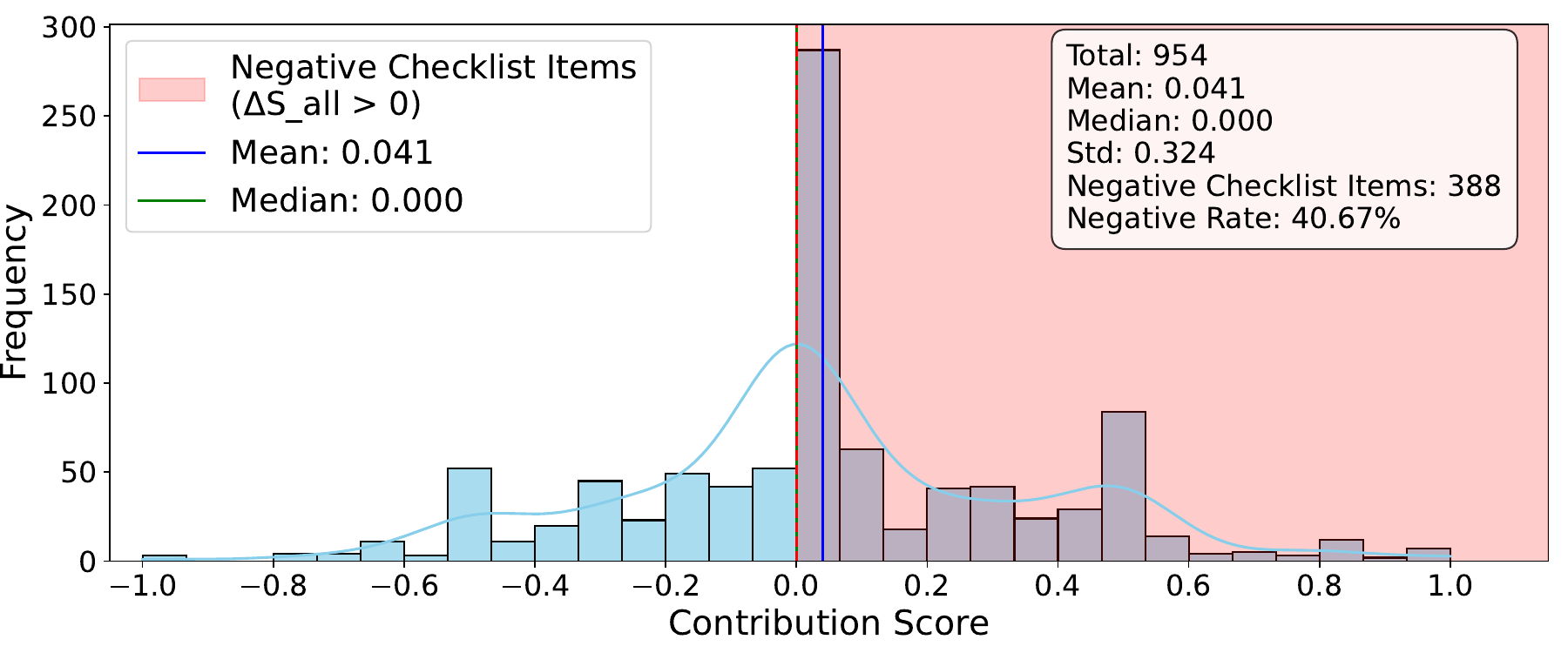}
    \end{minipage}\hfill
    \begin{minipage}{0.48\textwidth}
        \centering
        \includegraphics[width=\textwidth]{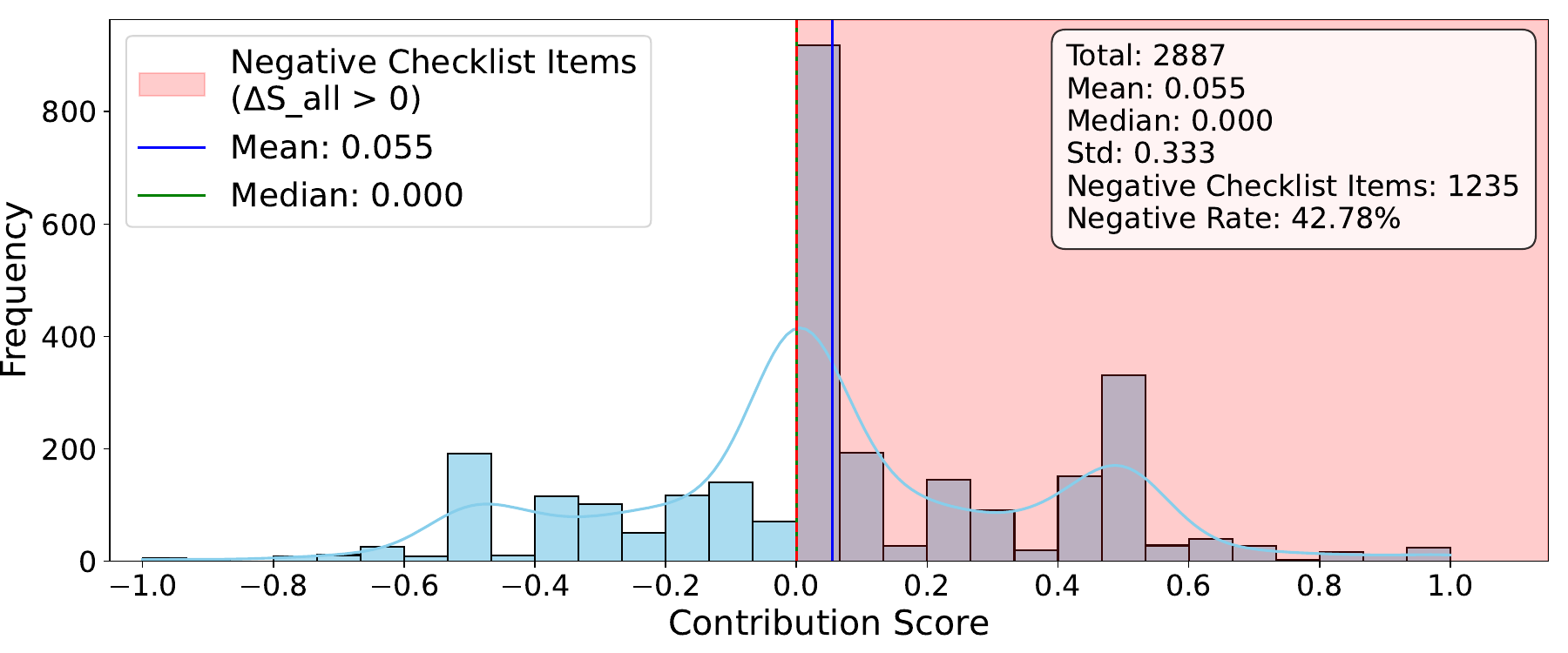}
    \end{minipage}
    \vskip 0.01em
    \begin{minipage}{0.48\textwidth}
        \centering
        \includegraphics[width=\textwidth]{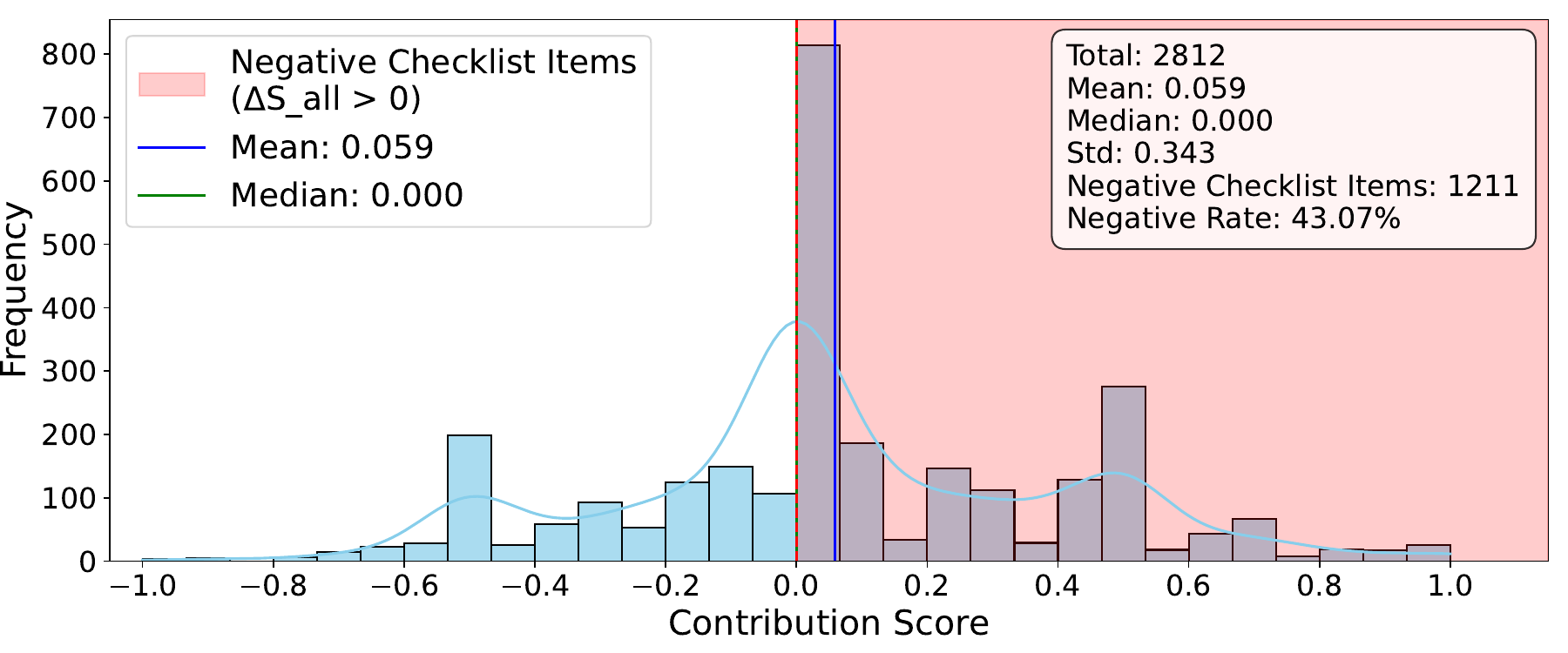}
    \end{minipage}\hfill
    \begin{minipage}{0.48\textwidth}
        \centering
        \includegraphics[width=\textwidth]{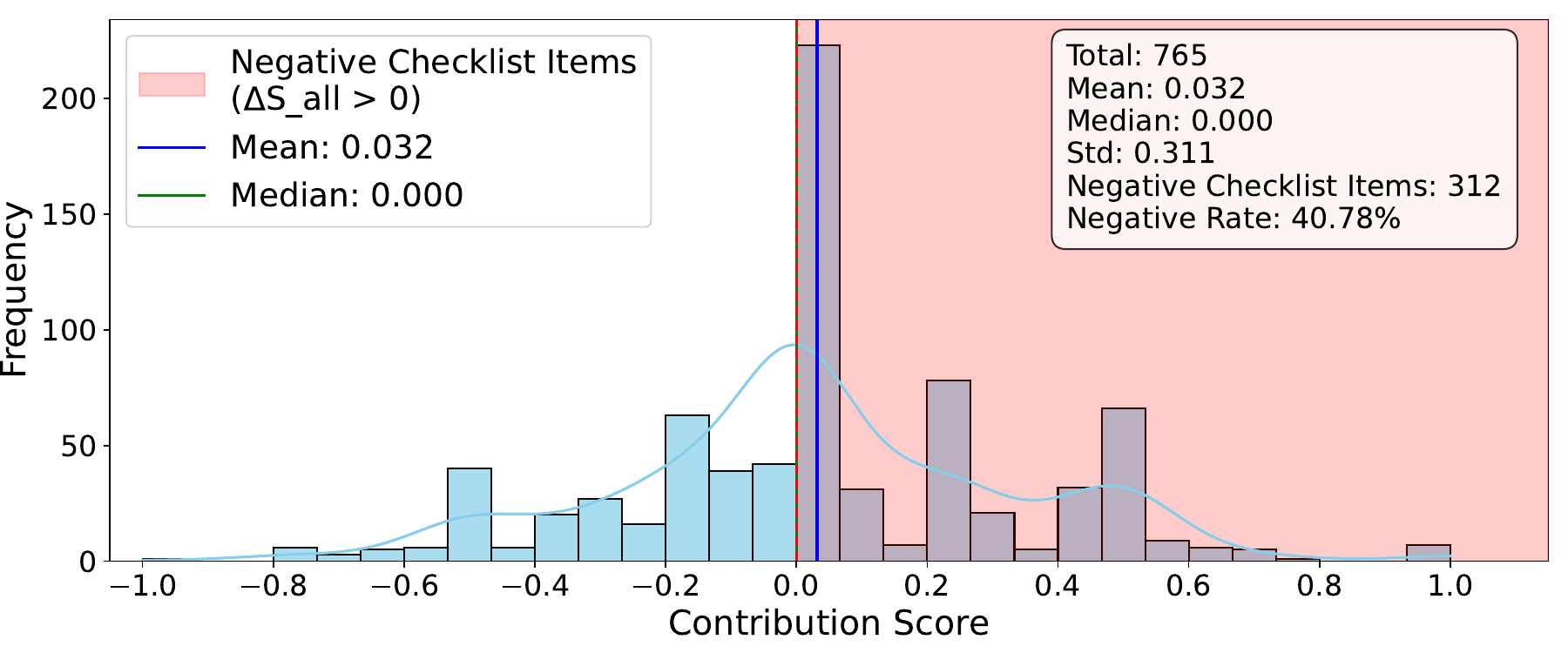}
    \end{minipage}
    
    \vskip 0.01em
    \begin{minipage}{0.48\textwidth}
        \centering
        \includegraphics[width=\textwidth]{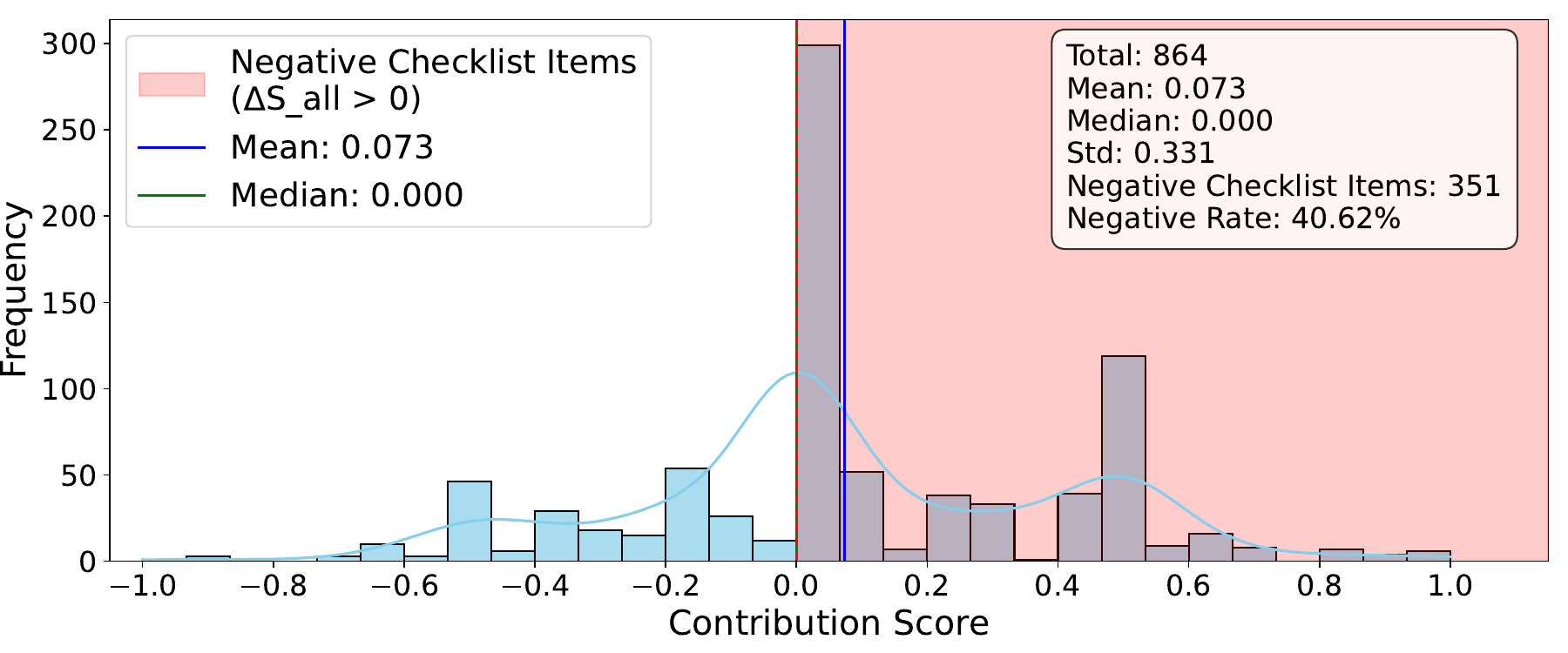}
    \end{minipage}\hfill
    \begin{minipage}{0.48\textwidth}
        \centering
        \includegraphics[width=\textwidth]{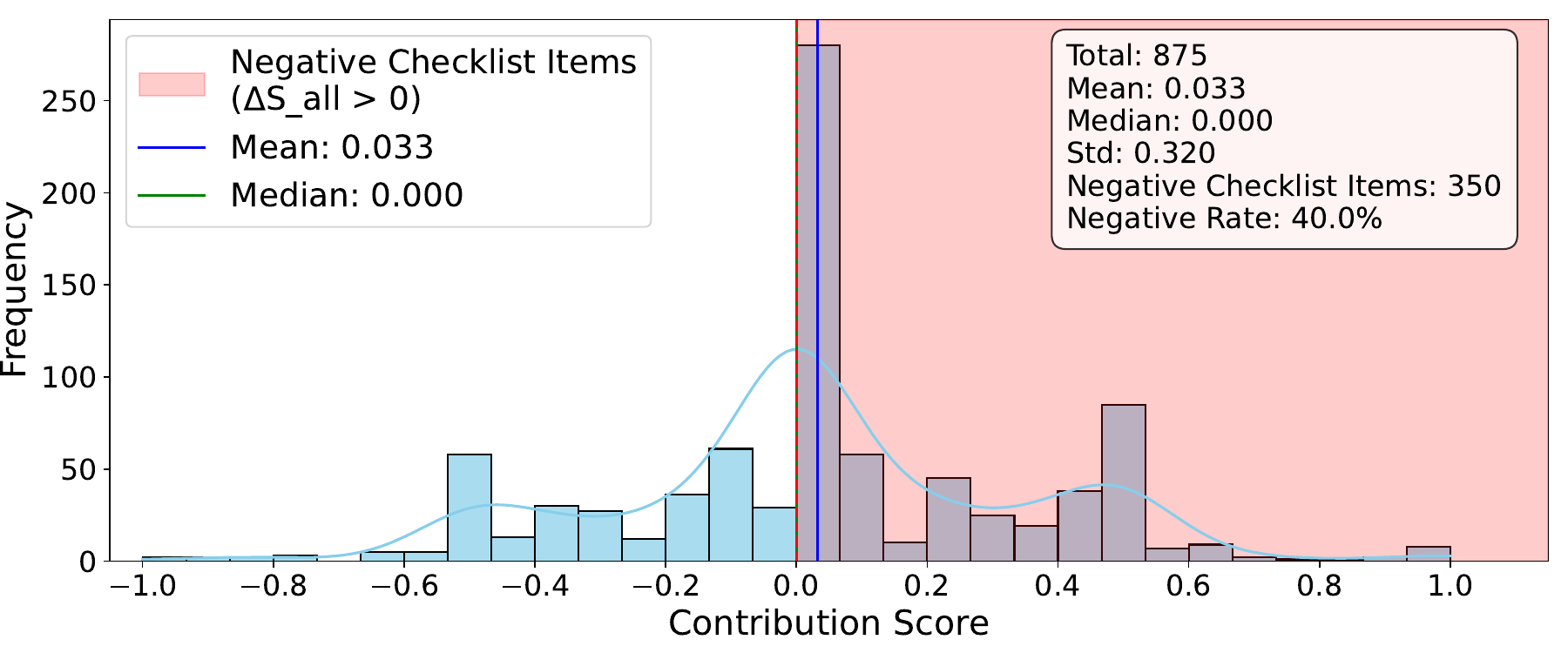}
    \end{minipage}

    \caption{Negative checklist ablation results on the LLMBar dataset. Each plot shows how removing checklist items impacts correlation with human evaluation. \textit{Negative} checklist items ($\Delta \bar{s}_{\text{abl}}$ > 0) are highlighted in red. Most scores lie between -0.1 and 0.1. The order of plots matches that of the \textit{positive} checklist items. For all checklist generation policies, the proportion of checklists in the final negative region is around 40\%.}
    \label{LLMBar_ablation_negative_checklist}
\end{figure*}
\begin{figure*}[htbp]
    \centering
    % 1行目：3つの図
    \begin{minipage}{0.48\textwidth}
        \centering
        \includegraphics[width=\textwidth]{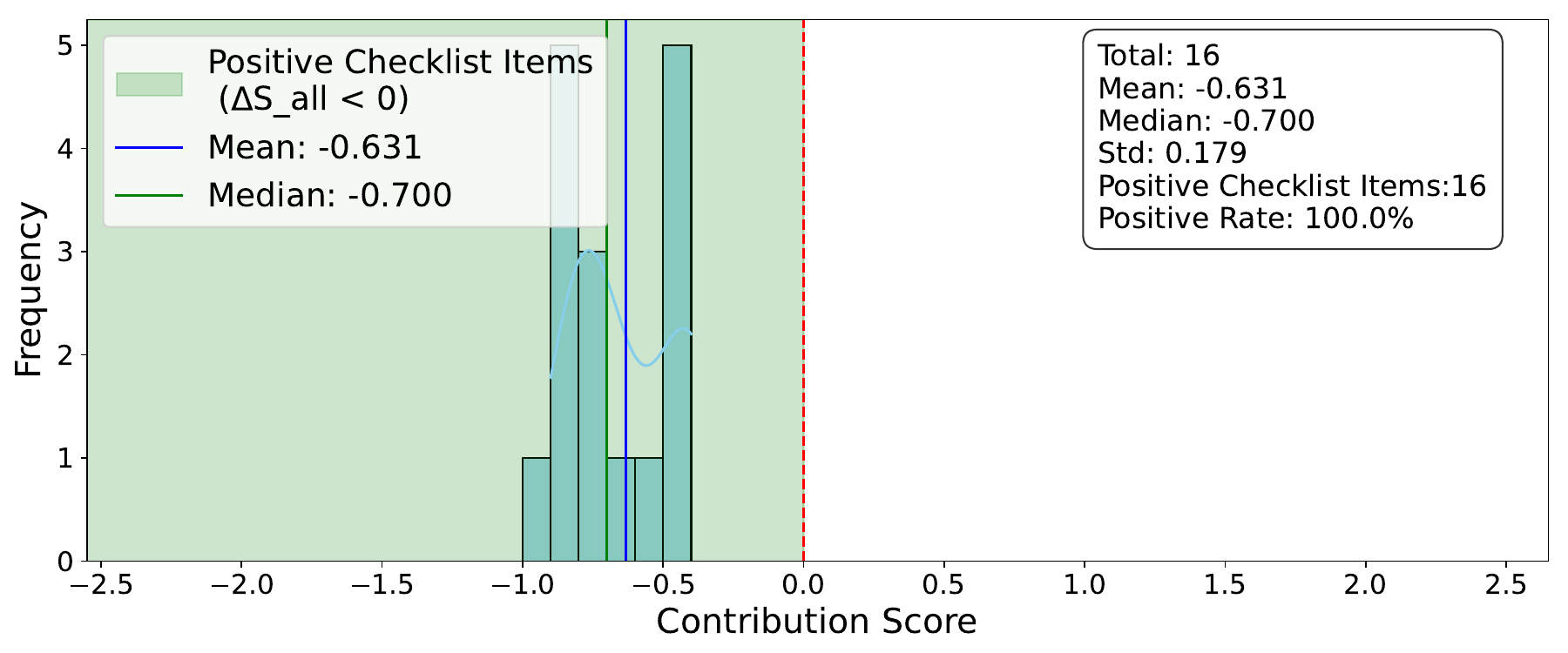}
    \end{minipage}\hfill
    \begin{minipage}{0.48\textwidth}
        \centering
        \includegraphics[width=\textwidth]{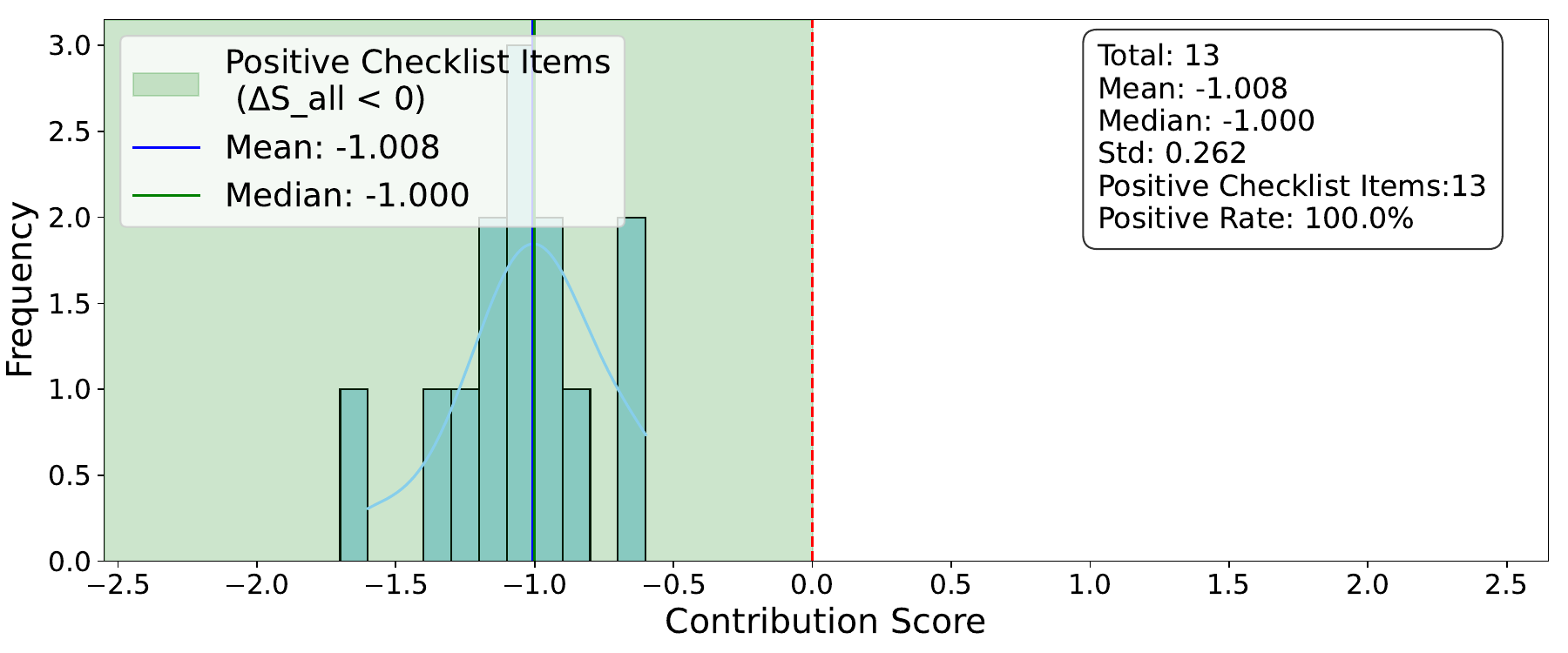}
    \end{minipage}
    
    \vskip 0.01em
    \begin{minipage}{0.48\textwidth}
        \centering
        \includegraphics[width=\textwidth]{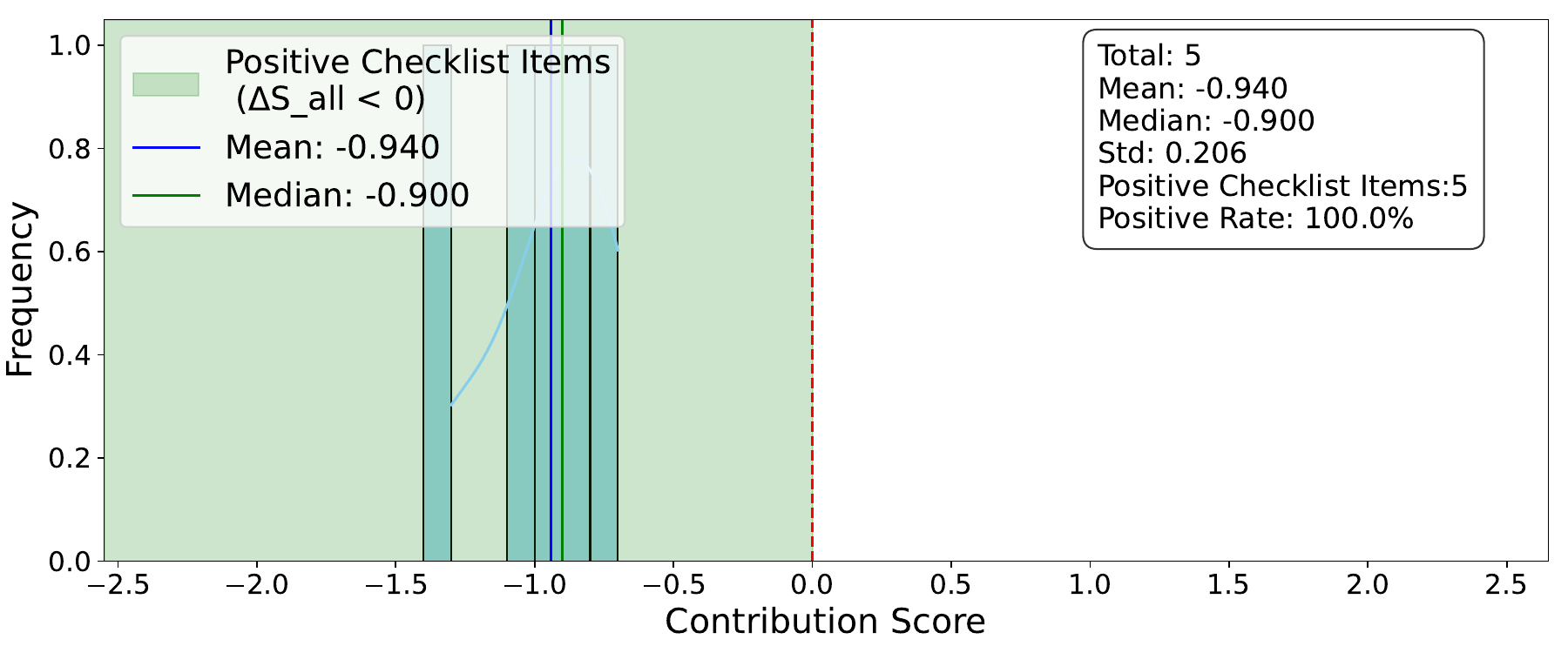}
    \end{minipage}\hfill
    \begin{minipage}{0.48\textwidth}
        \centering
        \includegraphics[width=\textwidth]{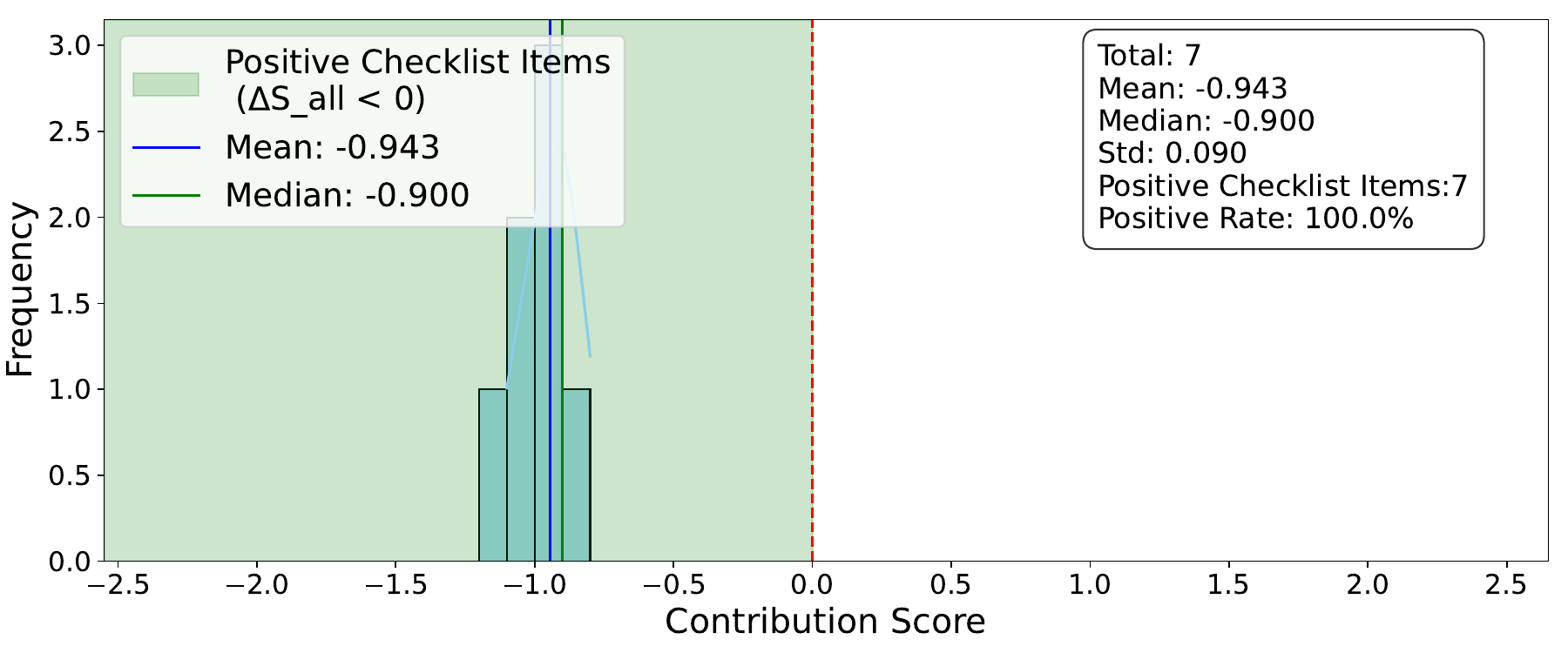}
    \end{minipage}

    \vskip 0.01em
    \begin{minipage}{0.48\textwidth}
        \centering
        \includegraphics[width=\textwidth]{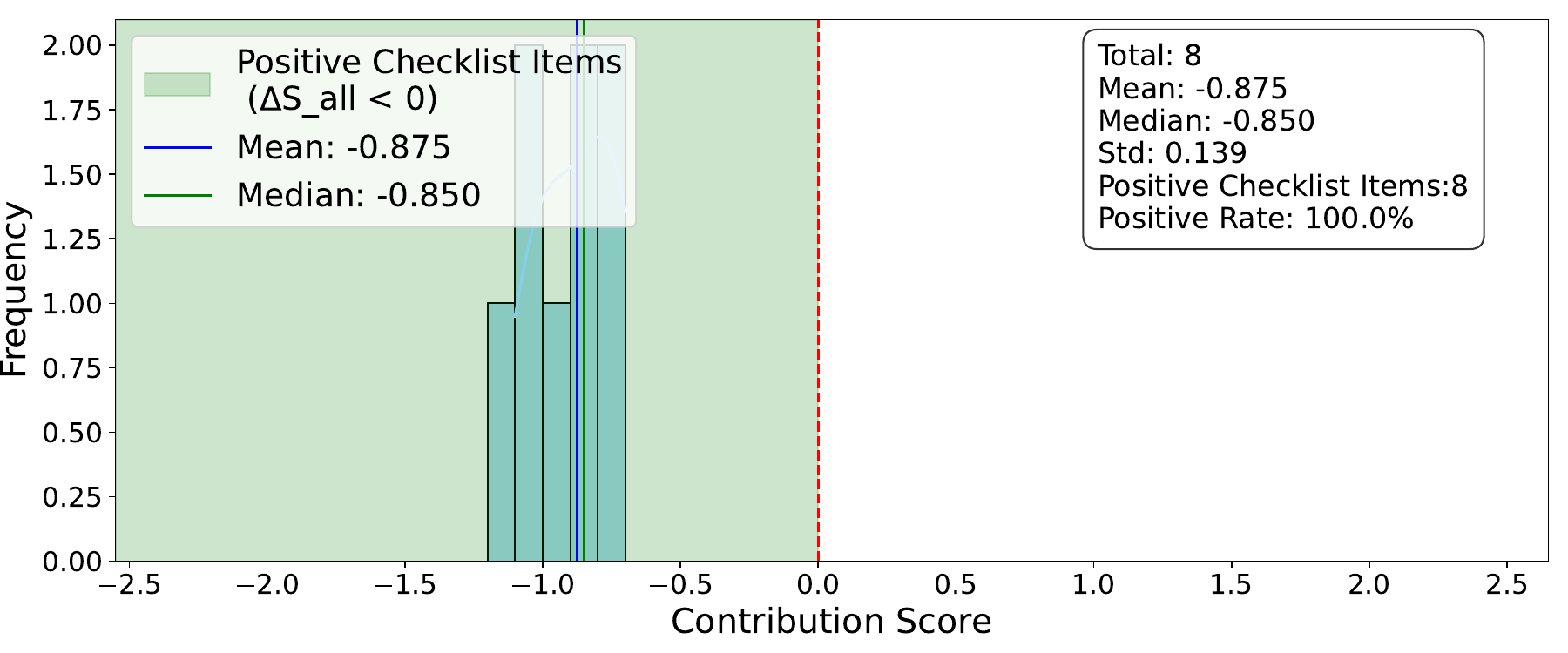}
    \end{minipage}\hfill
    \begin{minipage}{0.48\textwidth}
        \centering
        \includegraphics[width=\textwidth]{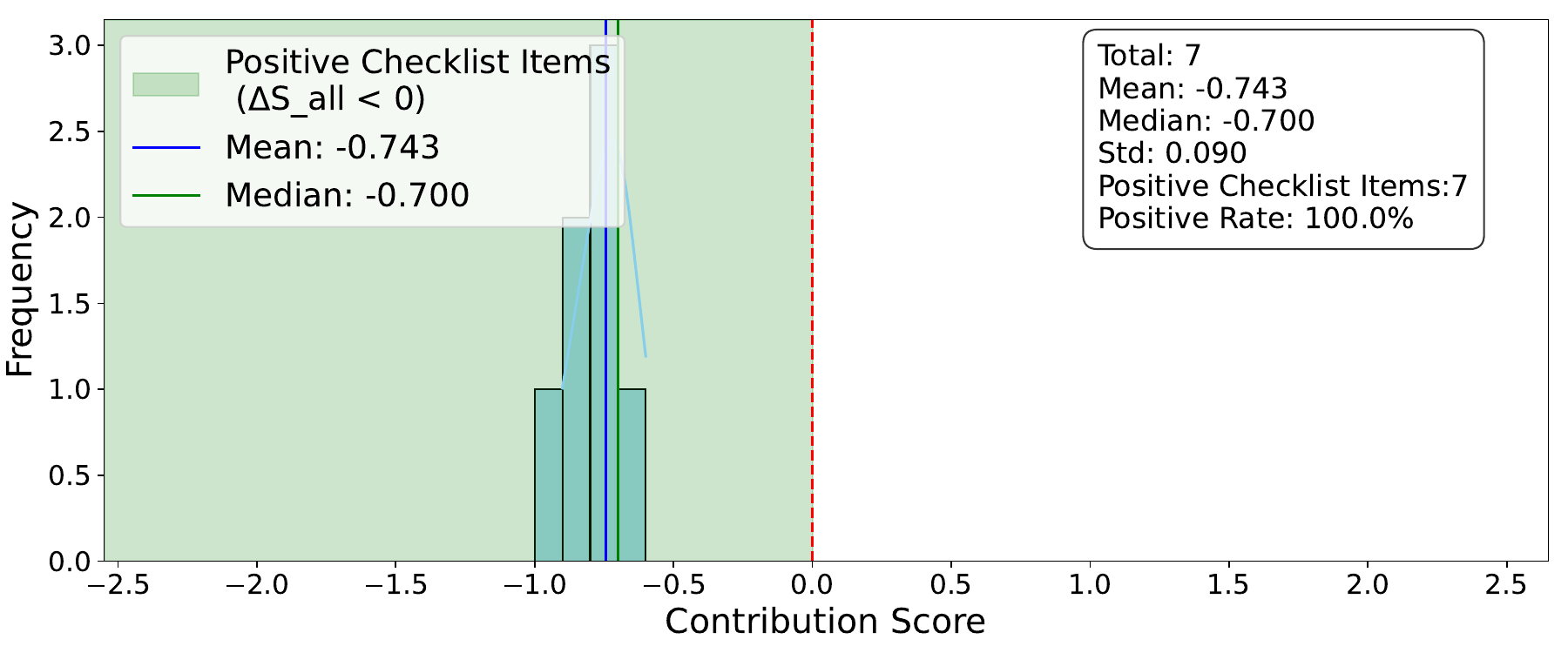}
    \end{minipage}

    \caption{Positive checklist ablation results on the InFoBench dataset. Each plot shows how removing checklist items impacts correlation with human evaluation. \textit{Positive} checklist items ($\Delta \bar{s}_{\text{abl}}$ < 0) are highlighted in green. All checklists are \textit{positive} checklist items.
    }
    \label{InFoBench_ablation_positive_checklist}
\end{figure*}
\begin{figure*}[htbp]
    \centering
    % 1行目：3つの図
    \begin{minipage}{0.48\textwidth}
        \centering
        \includegraphics[width=\textwidth]{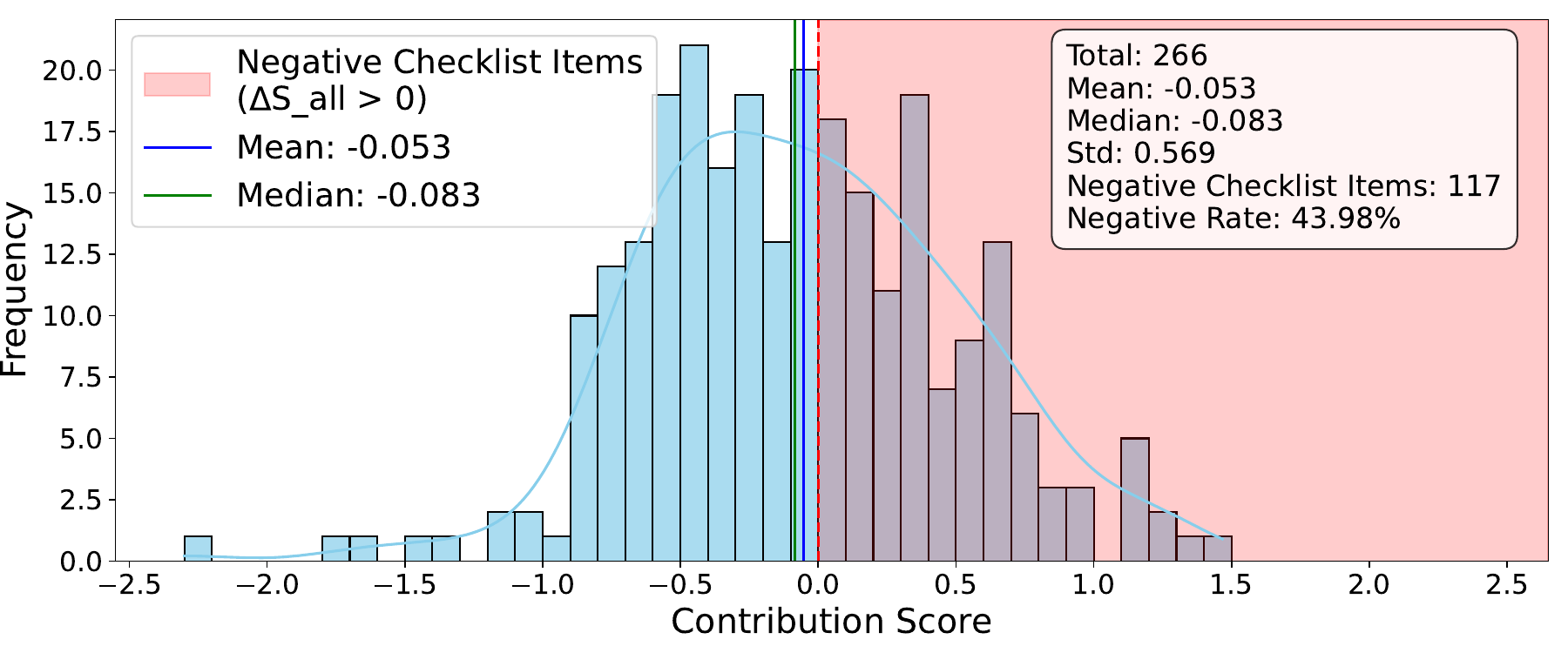}
    \end{minipage}\hfill
    \begin{minipage}{0.48\textwidth}
        \centering
        \includegraphics[width=\textwidth]{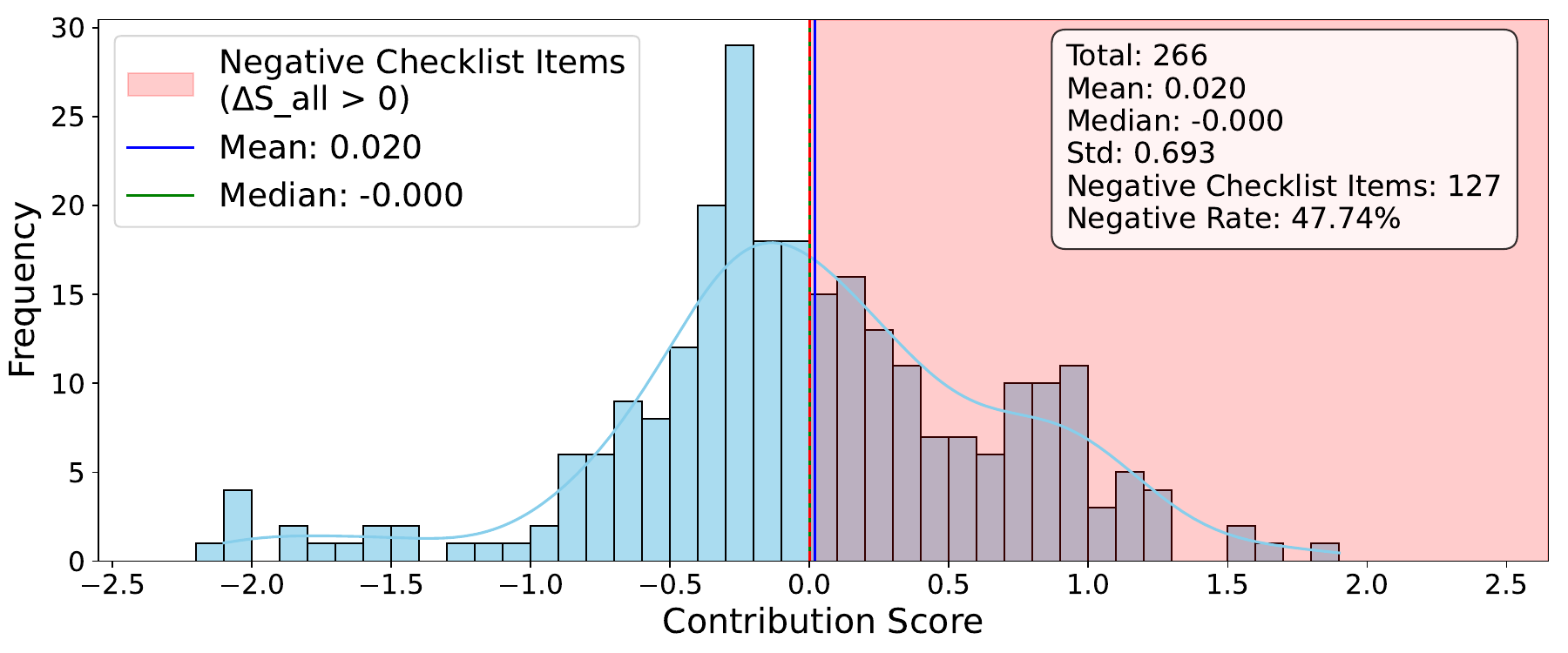}
    \end{minipage}

    \vskip 0.01em
    \begin{minipage}{0.48\textwidth}
        \centering
        \includegraphics[width=\textwidth]{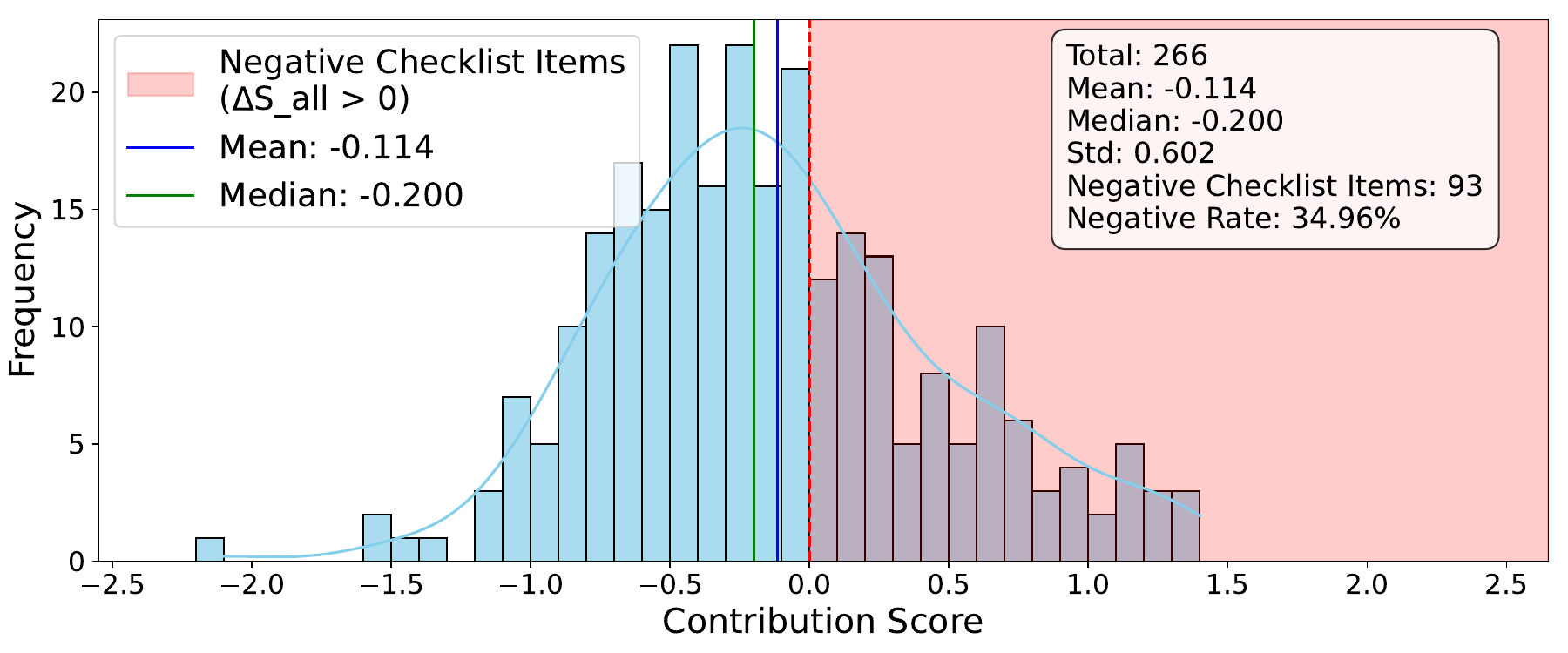}
    \end{minipage}\hfill
    \begin{minipage}{0.48\textwidth}
        \centering
        \includegraphics[width=\textwidth]{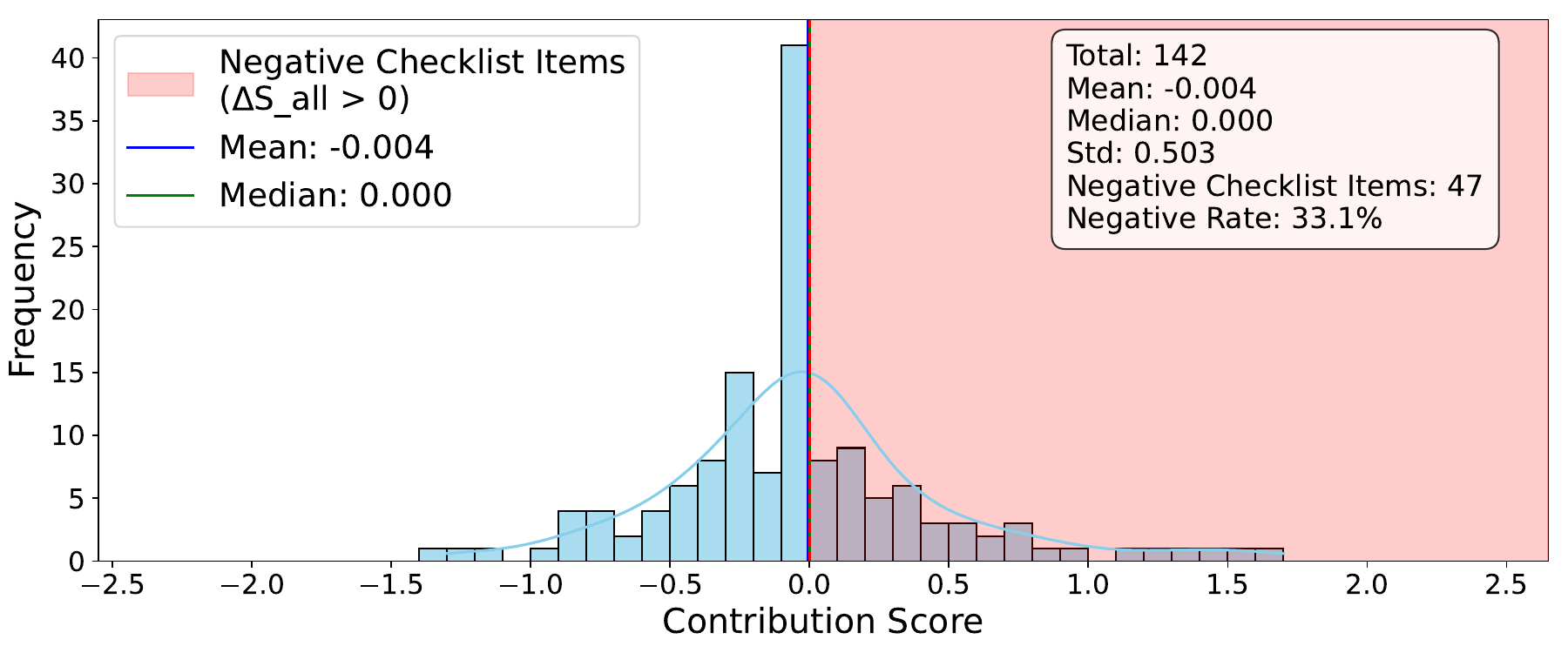}
    \end{minipage}

    \vskip 0.01em
    \begin{minipage}{0.48\textwidth}
        \centering
        \includegraphics[width=\textwidth]{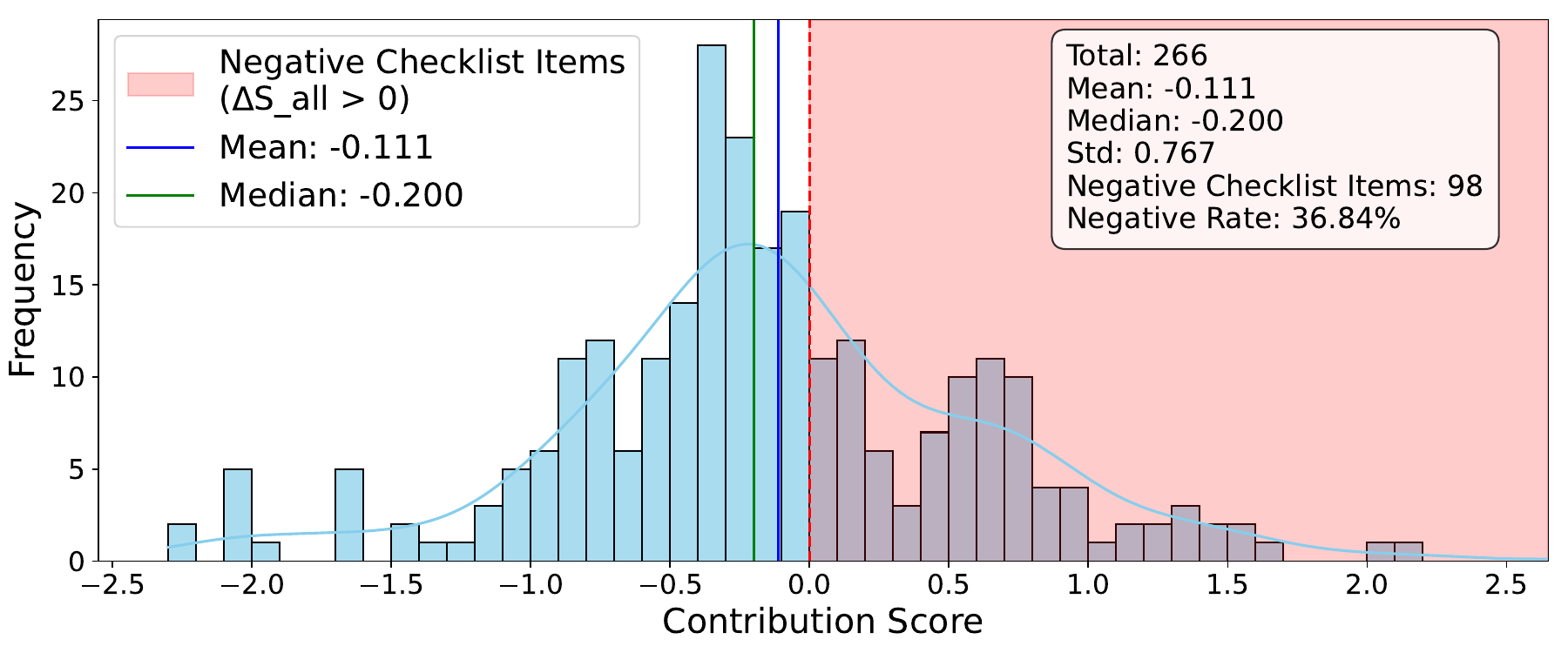}
    \end{minipage}\hfill
    \begin{minipage}{0.48\textwidth}
        \centering
        \includegraphics[width=\textwidth]{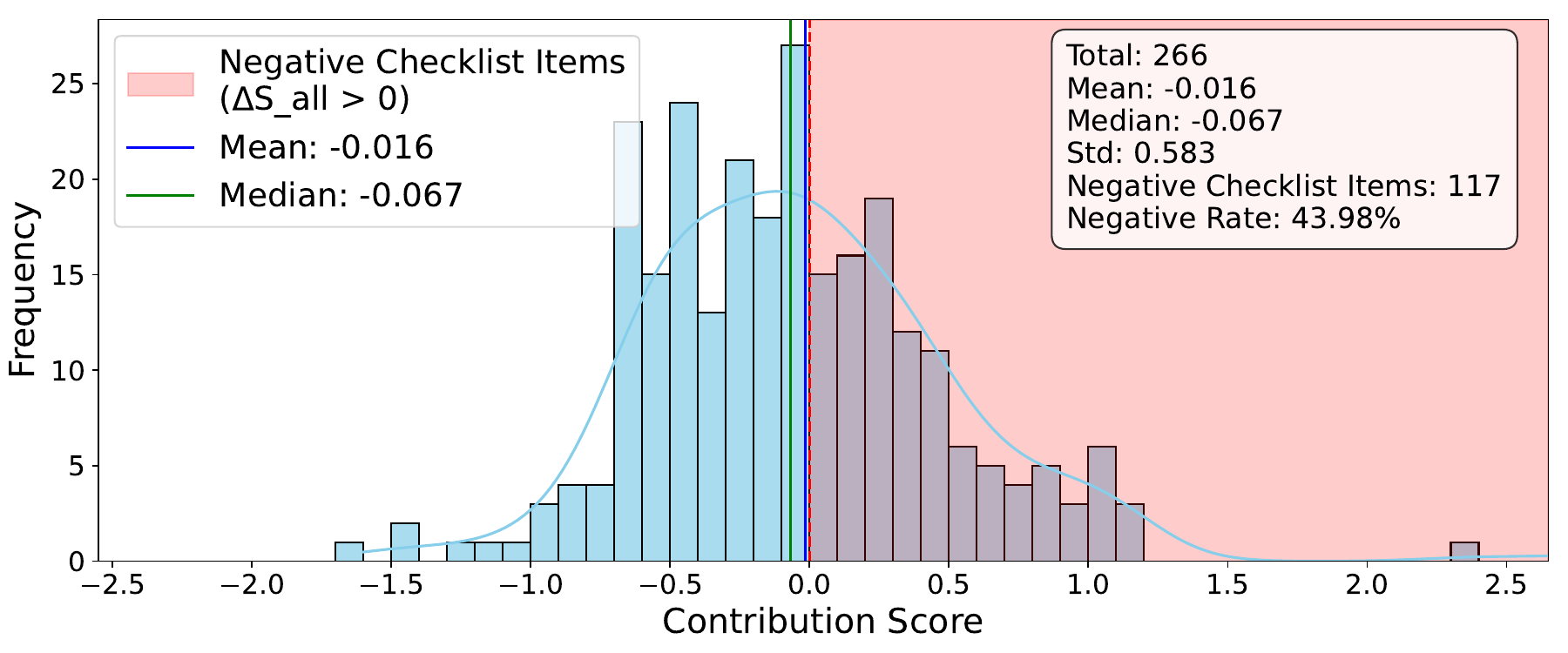}
    \end{minipage}

    \caption{Negative checklist ablation results on the InFoBench dataset. Each plot shows how removing checklist items impacts correlation with human evaluation. \textit{Negative} checklist items ($\Delta \bar{s}_{\text{abl}}$ > 0) are highlighted in red. Most scores lie between -0.5 and 0.5. For all checklist generation policies, the proportion of checklists in the final negative region is from 30\% to 50\%.}
    \label{InFoBench_ablation_negative_checklist}
\end{figure*}

Figure~\ref{LLMBar_ablation_positive_checklist} shows that positive checklist items predominantly cluster around the 0.0 score, indicating their limited impact on evaluation performance.
Similarly, Figure~\ref{LLMBar_ablation_negative_checklist} illustrates that nearly half of the negative checklist items have final scores between 0.0 and 0.1, suggesting that their negative impact is also minimal.
Across different policies, negative checklist items consistently represent approximately 40\% of all negative checklists.

However, the proportion of positive checklists that qualify as positive checklist items varies significantly across policies.
For example, in the Baseline, over 60\% of positive checklists (100 out of 160), whereas in the Length * 0.5, less than 40\% meet this criterion (53 out of 137).

In the InFoBench dataset, we identify 56 items as positive checklists and 1472 items as negative checklists in Figure~\ref{InFoBench_ablation_positive_checklist} and~\ref{InFoBench_ablation_negative_checklist}.
Notably, all positive checklists qualify as positive checklist items, while 599 items (40.7\%) qualify as negative checklist items.
The majority of these negative checklist items have final scores between -0.5 and 0.5.
Across different policies, negative checklist items represent between 30\% and 50\% of all negative checklists.
Almost all positive checklist items in the InFoBench dataset demonstrate final scores below -0.5, suggesting they consistently bring evaluation results closer to gold labels, albeit by a small margin.

\subsection{Qualitative Analysis}
\subsubsection{Threshold Determination for Qualitative Analysis of Checklist Effectiveness}
\label{Threshold Determination for Qualitative Analysis of Checklist Effectiveness}
We detail the specific criteria used to identify checklists with a substantial impact on evaluation performance in each dataset. These thresholds serve as the basis for our qualitative analysis.

For the LLMBar dataset, we analyze the following:
\begin{itemize}
\item \textit{Positive} checklist items with $\Delta \bar{s}_{\text{abl}}$ < $0.9$
\item \textit{Negative} checklist items with $\Delta \bar{s}_{\text{abl}}$ > $0.9$
\end{itemize}
Applying these stringent criteria, we identify 33 items (1.6\% of checklists) and 102 items (2.6\% of negative checklists) as having a substantial impact on evaluation.

For the InFoBench dataset, we examine the following:
\begin{itemize}
\item All positive checklist items
\textit{Negative} checklist items with $\Delta \bar{s}_{\text{abl}}$ > $0.9$
\end{itemize}
These thresholds yield 56 items (All positive checklist items) and 102 items (6.9\% of negative checklists) that show notable influence on evaluation quality.

\begin{figure*}[t]
\centering
\includegraphics[width=\linewidth]{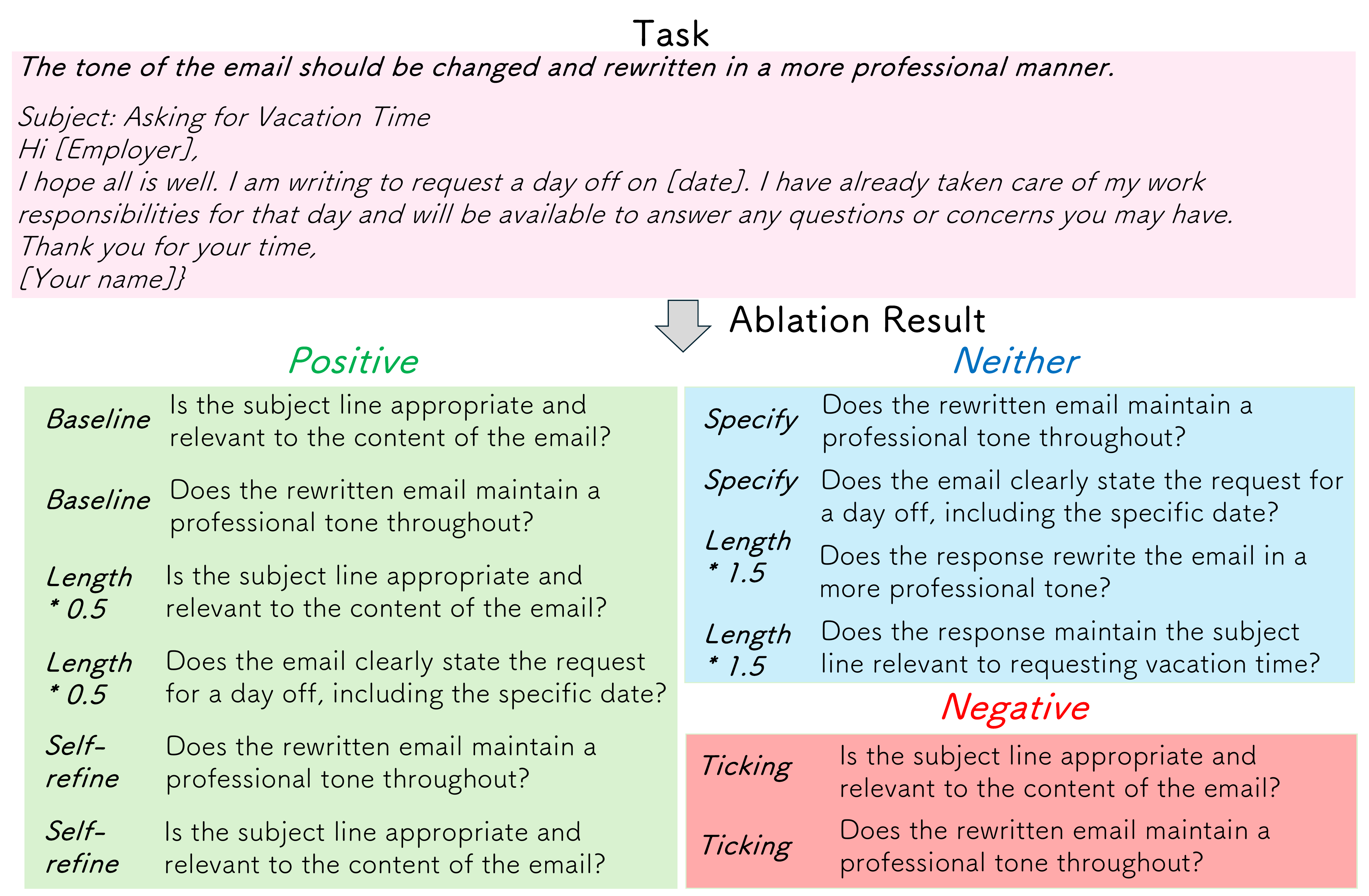}
\caption{
All checklist policies' checklist items.  We use Qwen2.5-7B-it as the evaluation model and select the following task from the InFoBench.
In the ablation result, we categorize outputs as \textit{Positive}, \textit{Negative}, or Neither, and we show one example.  
For this task, the Baseline, Length*0.5, and Self-refine variants produce outputs labeled as \textit{Positive}.  
In contrast, only the Ticking variant produces \textit{Negative} checklist items.
}
\label{tb:checklist_example}
\end{figure*}

\subsubsection{In-Depth Qualitative Analysis of Generated Checklist Items}
\label{In-Depth Qualitative Analysis of Generated Checklist Items}

We conduct a qualitative analysis focusing on checklist items that significantly impact evaluation performance, aiming to identify which characteristics contribute to alignment with human evaluation.  
Based on this analysis, we identify six labels for \textit{positive} checklist items and four labels for \textit{negative} ones.

The following are the \textit{positive} labels:

\begin{itemize}
\item \textbf{Explicit Focus (Explicit)}: The checklist item clearly states which aspect of the question the response should focus on.
\item \textbf{Implicit Focus (Implicit)}: The focus is not explicitly stated in the item, but it reflects an important aspect necessary for answering or evaluating the question.
\item \textbf{Proposal Answer (Ans)}: The item encourages or expects a concrete answer, such as a proposal or opinion, in the response.
\item \textbf{Clarity}: The item evaluates how easy the response is to understand, such as ``Is the response informative and provides a clear explanation?''.
\item \textbf{Additional Content (Add)}: The item includes aspects not strictly required to answer the question, but still useful for checklist-based evaluation.
\item \textbf{Tone}: The item assesses the appropriateness of the response's tone or style, such as ``Is the language clear and formal, appropriate for a legal notice?''.
\end{itemize}

\begin{table*}[t]
\centering
\small
\setlength{\tabcolsep}{1.8pt}
\renewcommand{\arraystretch}{1.3}
\begin{tabular}{p{0.08\textwidth}p{0.38\textwidth}cccp{0.44\textwidth}}
\toprule
\multicolumn{1}{c}{Evaluator} & \multicolumn{1}{c}{Task} & H & N & C & \multicolumn{1}{c}{Checklist} \\
\midrule
Qwen2.5-7B-it & Identify the programming language used to write the given code. \newline \newline if (20 $>$ 18) {printf(``20 is greater than 18'');} & 5 & 5 & 3 & $\square$ Does the response correctly identify the use of the `printf' function as characteristic of the C programming language?\newline $\square$  Does the response recognize the syntax of the conditional statement as typical of C-style languages? \\
\rowcolor{gray!20}
gemma-2-27b-it & A confirmation email should be written appropriately for the situation. A meeting has been scheduled, and the sender expects the other to review the slides. & 5 & 3 & 5 & $\square$ Does the response confirm the details of the scheduled meeting?\newline $\square$ Does the response mention that the recipient is expected to review the slides before the meeting? \\
Ministral-8B-it & Think of alternatives and paraphrases for the underlined word. \newline \newline what we have $\_$expected & 5 & 3 & 5 & $\square$ Does the response provide alternatives to the word ``expected''? \newline $\square$  Does the response offer paraphrases that fit in the context of ``what we have expected''? \\
\bottomrule
\end{tabular}
\caption{Analysis of evaluator and checklists in InFoBench: examples of models with significant changes in correlation when using checklists (H: human labels, N: without checklists, C: checklists applied to a response).
}
\label{tb:checklist_analysis}
\end{table*}

\begin{table}[t]
\footnotesize
\centering
\tabcolsep 3pt
\renewcommand{\arraystretch}{1.3} 
\begin{tabular}{lcccccc}
\toprule
Pos& Explicit&	Implicit&Ans&	Clarity	&Add&Tone \\ 
\midrule
LLMBar&	18&	5	&3	&2&	5&	0\\
\rowcolor{gray!20}InFoBench&	33	&20&0&	0	&2&	1\\
Sum&	51&	25&	3	&2	&7&	1\\
\rowcolor{gray!20}Rare(\%)&	57.3	&28.1	&3.4&	2.2&	7.9&	1.1\\
\bottomrule
\end{tabular}
\caption{Label distribution of checklist items classified as \textit{positive}.
The majority (≈60\%) are \textbf{Explicit Focus (Explicit)} items, clearly aligning with elements explicitly stated in the question. About 30\% are \textbf{Implicit Focus (Implicit)}, reflecting important but implicit evaluation criteria.}
\label{positive_checklist_count}
\end{table}

\begin{table}[h]
\footnotesize
\centering
\tabcolsep 3pt
\renewcommand{\arraystretch}{1.3}
\begin{tabular}{lcccc}
\toprule
Neg& Non-negative&	Limited&Clarity	&Add\\ 
\midrule
LLMBar&	85&	11	&3	&3\\
\rowcolor{gray!20}InFoBench&92&10&0&0\\
Sum&177&21&	3	&3\\
\rowcolor{gray!20}Rare(\%)&	86.8	&10.3	&1.5&	1.5\\
\bottomrule
\label{negative_checklist_count}
\end{tabular}
\caption{Distribution of textit{negative} checklist item labels.
While more than 85\% of the checklist items labeled as \textbf{Non-negative} are still aligned with the question and valid upon manual inspection, about 10\% are found to be \textbf{Limited Content (Limited)} in evaluating the response, suggesting room for improvement in checklist quality.}
\label{negative_checklist_count}
\end{table}

The following are the \textit{negative} labels:

\begin{itemize}
\item \textbf{Non-negative}: Although classified as a \textit{negative} item, it is still reasonably usable as a checklist item for the given question.
\item \textbf{Limited Content (Limited)}: The item reflects only a narrow or insufficient aspect of the response, failing to adequately capture its quality.
\item \textbf{Clarity}: The item evaluates how easy the response is to understand, such as ``Is the response logically consistent with the analogy format presented in the question?''.
\item \textbf{Additional Content (Add)}: The item includes aspects not strictly required to answer the question, but still useful for checklist-based evaluation.
\end{itemize}

Tables~\ref{positive_checklist_count} and~\ref{negative_checklist_count} present the distribution of \textit{positive} and \textit{negative} checklist item labels for the LLMBar and the InFoBench, respectively.
These tables illustrate how frequently each label type appears in the generated checklists.

\subsection{Generate Checklist Example}
\label{sec:generate_checklist_example}
Table~\ref{tb:checklist_analysis} shows an analysis of evaluators and checklists in the InFoBench.
This example illustrates how the evaluation results change for multiple evaluators when comparing the cases without a checklist (N) and with a checklist (C), in relation to human labels (H). 
It also presents the corresponding checklists used in each case.
Qwen2.5-7B-it shows lower evaluation performance when using a checklist, reducing its alignment with human judgments.
In contrast, Gemma-2-27B-it and Mistral-8B-it improve their alignment with human evaluations when using checklists.

Figure~\ref{tb:checklist_example}
 shows all the checklist policies' checklist items. 
 We use Qwen2.5-7B-it as an evaluation model and one of the \textit{open} questions as a task.
 The column labeled ``Ablation Result'' reports the outcome of ablation studies conducted on individual checklist items.
Items marked as \textit{Positive} contribute to alignment with human evaluations, while those marked as \textit{Negative} do not. Items labeled \textit{Neither} show no clear effect in either direction.
For each policy, we provide two representative examples to illustrate the effects.
 
\subsection{Prompts for Checklist Generation and Response Evaluation}
\label{checklist_generation_prompt}
\label{sec:appendix_generate_checklist}
\begin{figure*}[h]
    \centering
    \includegraphics[width=\textwidth, keepaspectratio]
    {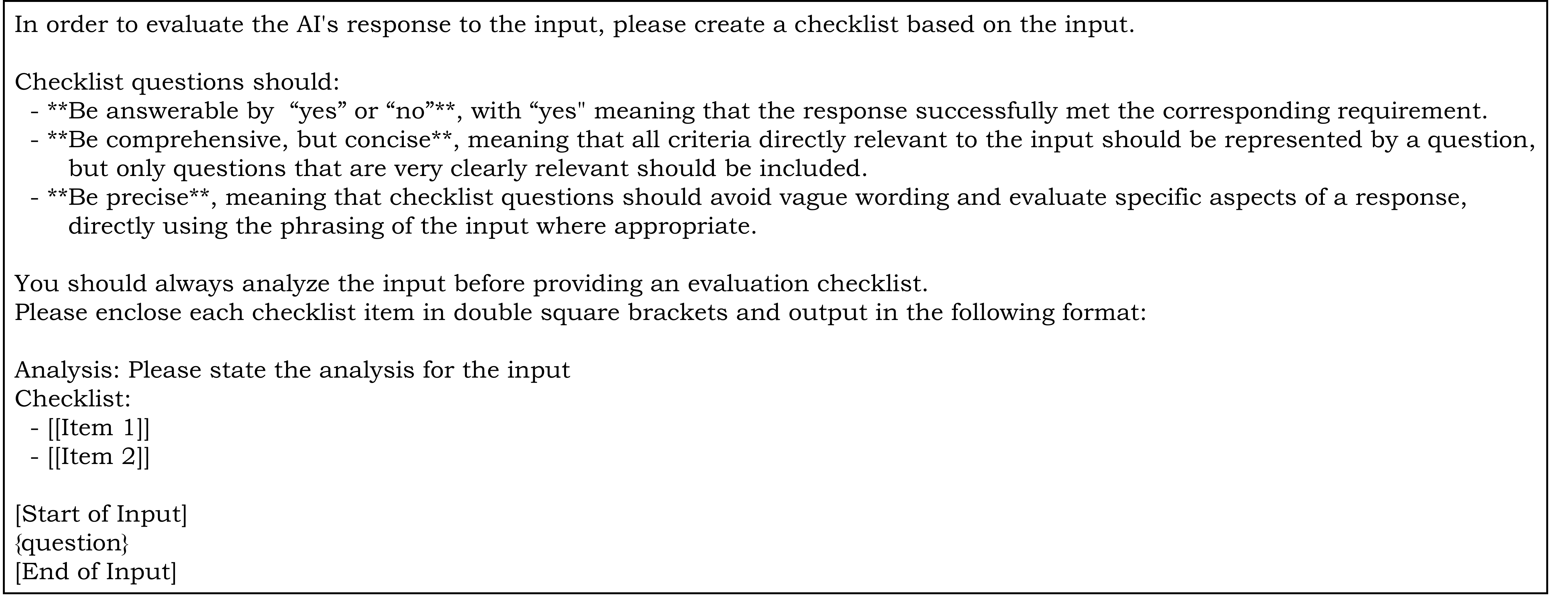} 
    \caption{Prompt used in the Baseline.}
    \label{baseline_checklist}
\end{figure*}

\begin{figure*}[h]
    \centering
    \includegraphics[width=\textwidth, keepaspectratio]
    {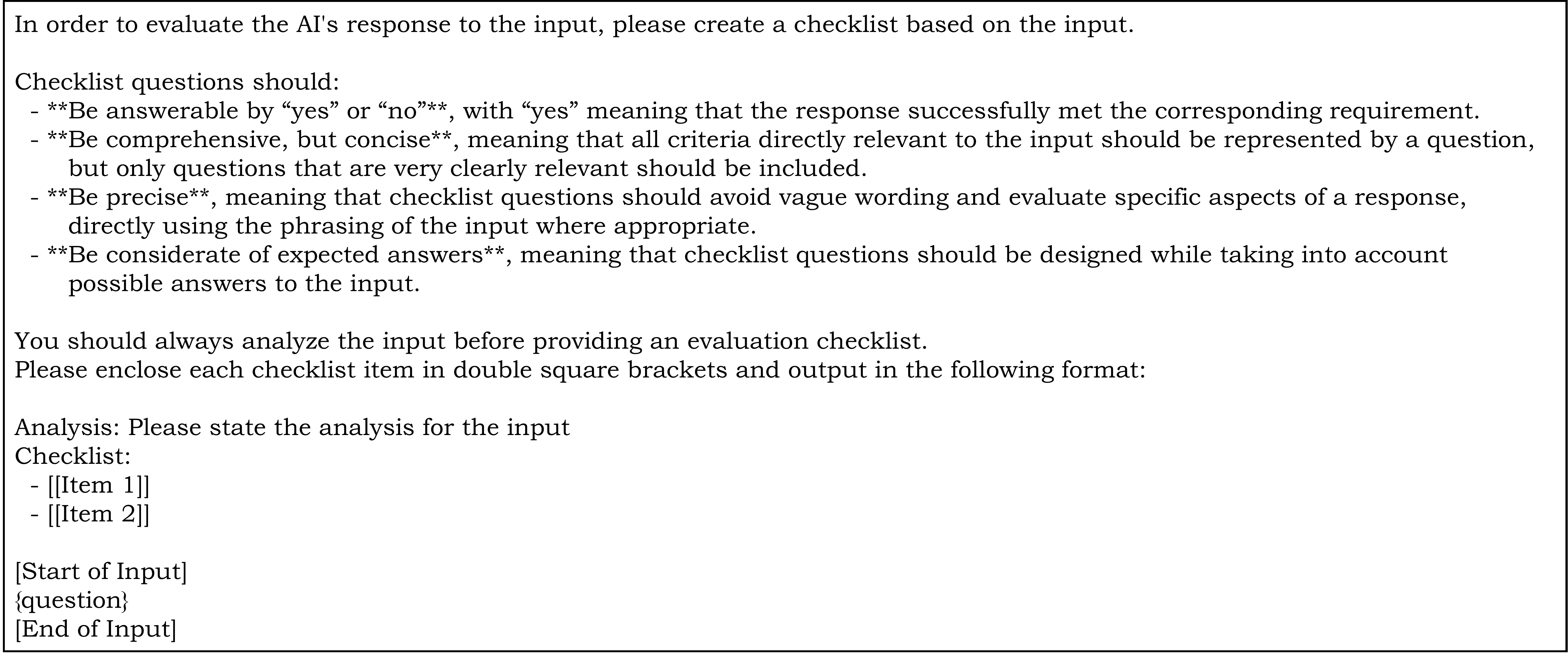} 
    \caption{Prompt used in the Specify.}
    \label{specify_checklist}
\end{figure*}

\begin{figure*}[h]
    \centering
    \includegraphics[width=\textwidth, keepaspectratio]
    {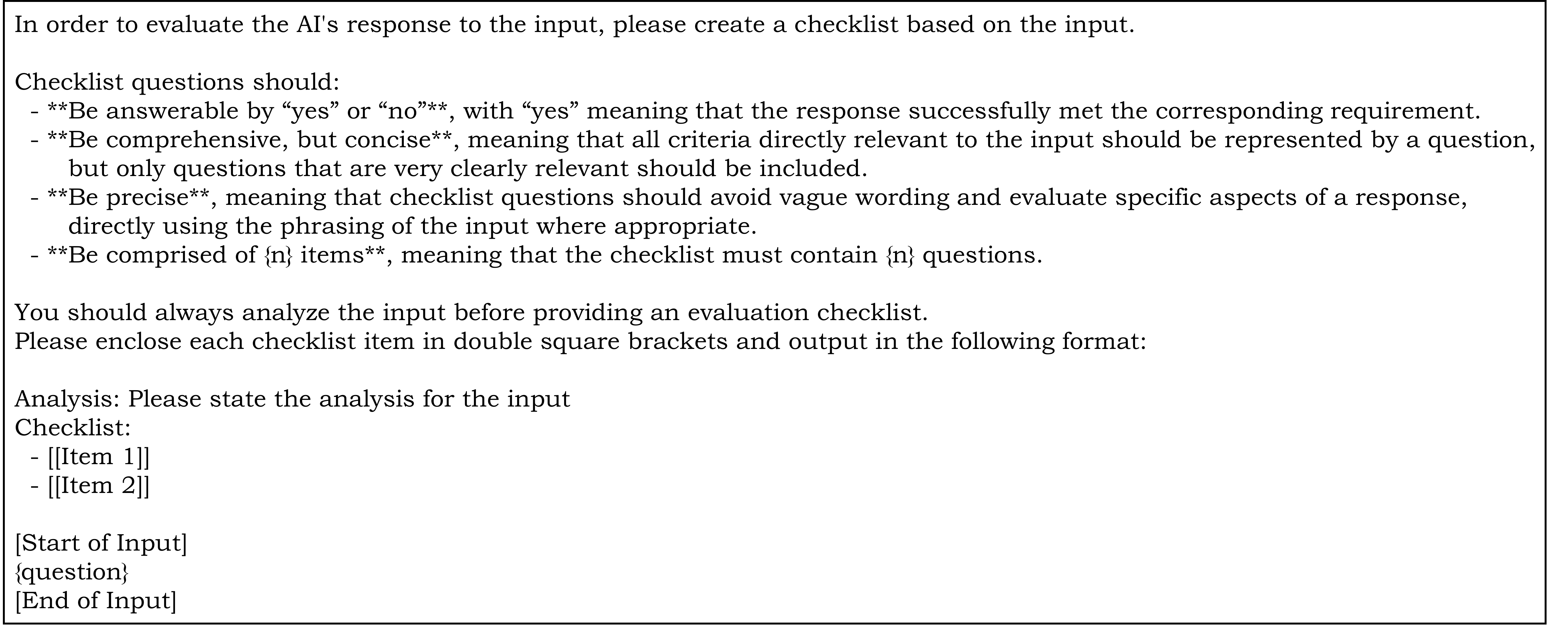} 
    \caption{Prompt used in the Checklist Length.
    First, the number of checklist items for the Baseline method is calculated. Then, the prompt instructs to generate a checklist with the number of items multiplied by 0.5 or 1.5.}
    \label{length_checklist}
\end{figure*}

\begin{figure*}[h]
    \centering
    \includegraphics[width=\textwidth, keepaspectratio]
    {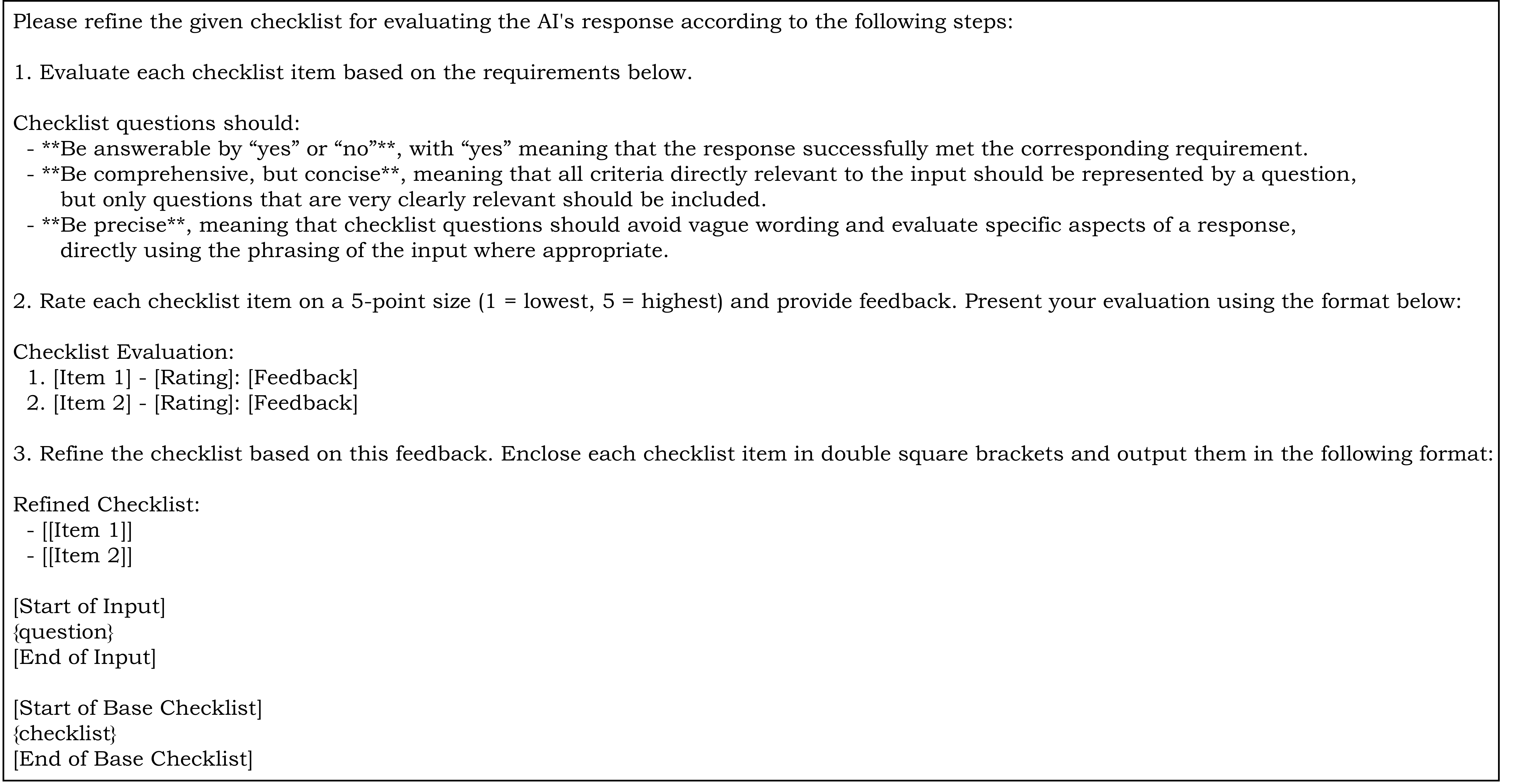} 
    \caption{Prompt used in the Self-refine.
    First, LLM generates checklists using the Baseline method.
Next, LLM outputs a Likert scale~(1-to-5) evaluation and its rationale for the generated checklists.
Finally, based on this rationale, LLM regenerates checklists.}
\label{self-refine_checklist}
\end{figure*}

\begin{figure*}[h]
    \centering
    \includegraphics[width=\textwidth, keepaspectratio]
    {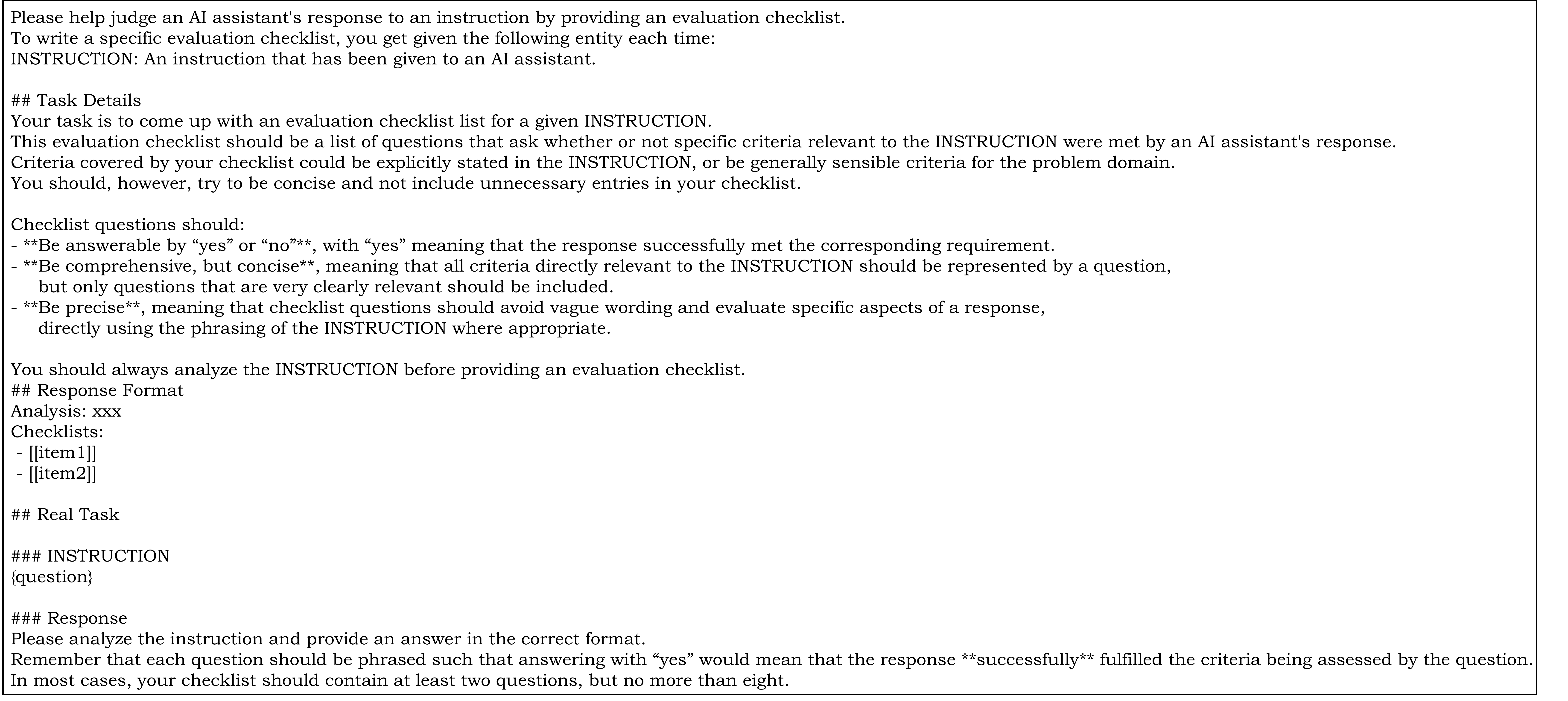} 
    \caption{Prompt uses in the Ticking.
   We use TICKing's~\cite{Cook2024TICKingAT} prompt.
Since their paper does not contain examples that they use, we remove the specific in-context examples to use this prompt.}
\label{ticking_checklist}
\end{figure*}

Figures~\ref{baseline_checklist}, ~\ref{specify_checklist}, ~\ref{length_checklist}, ~\ref{self-refine_checklist}, and ~\ref{ticking_checklist} present the prompt used for generating checklists(Baseline, Specify, Length,Self-refine, and Ticking). We use GPT-4o as the generation model.

\begin{figure*}[h]
    \centering
    \includegraphics[width=\textwidth, keepaspectratio]
    {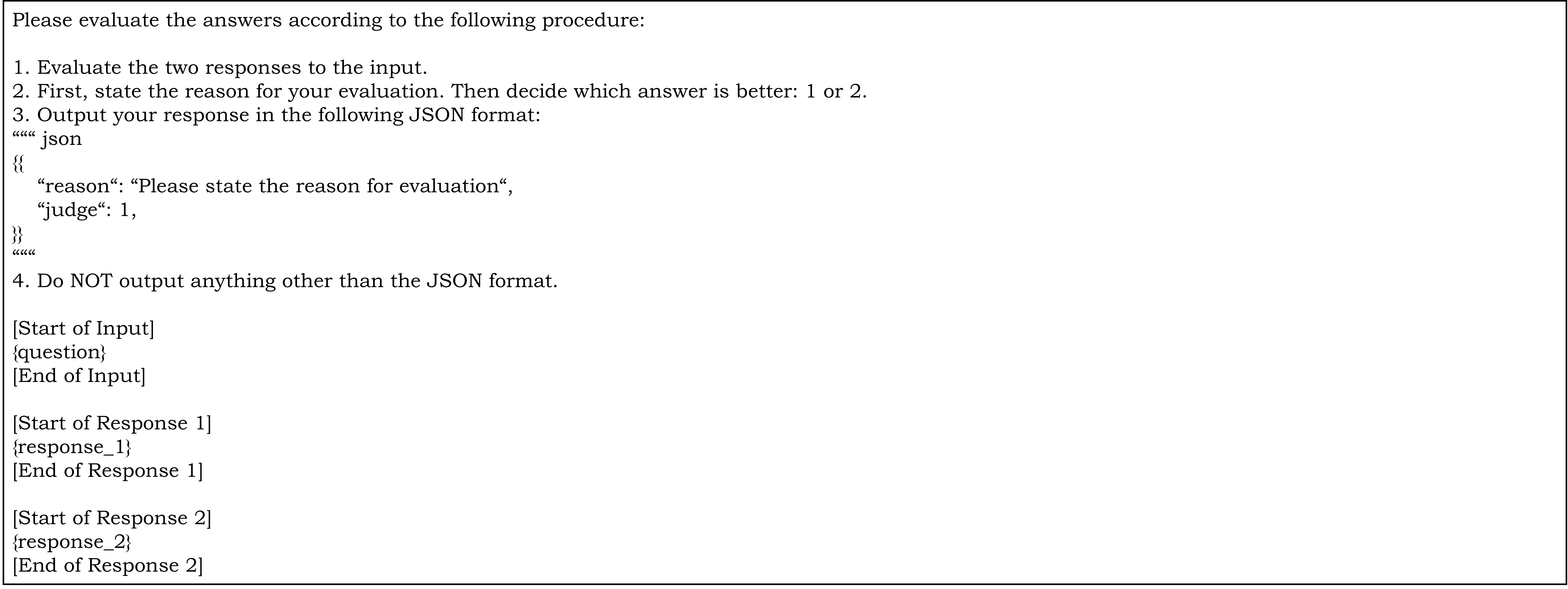} 
    \caption{Prompt used for pairwise response comparison, conducted \textbf{without} any checklist-based evaluation criteria.}
    \label{without_checklist_pairwise}
\end{figure*}

\begin{figure*}[h]
    \centering
    \includegraphics[width=\textwidth, keepaspectratio]
    {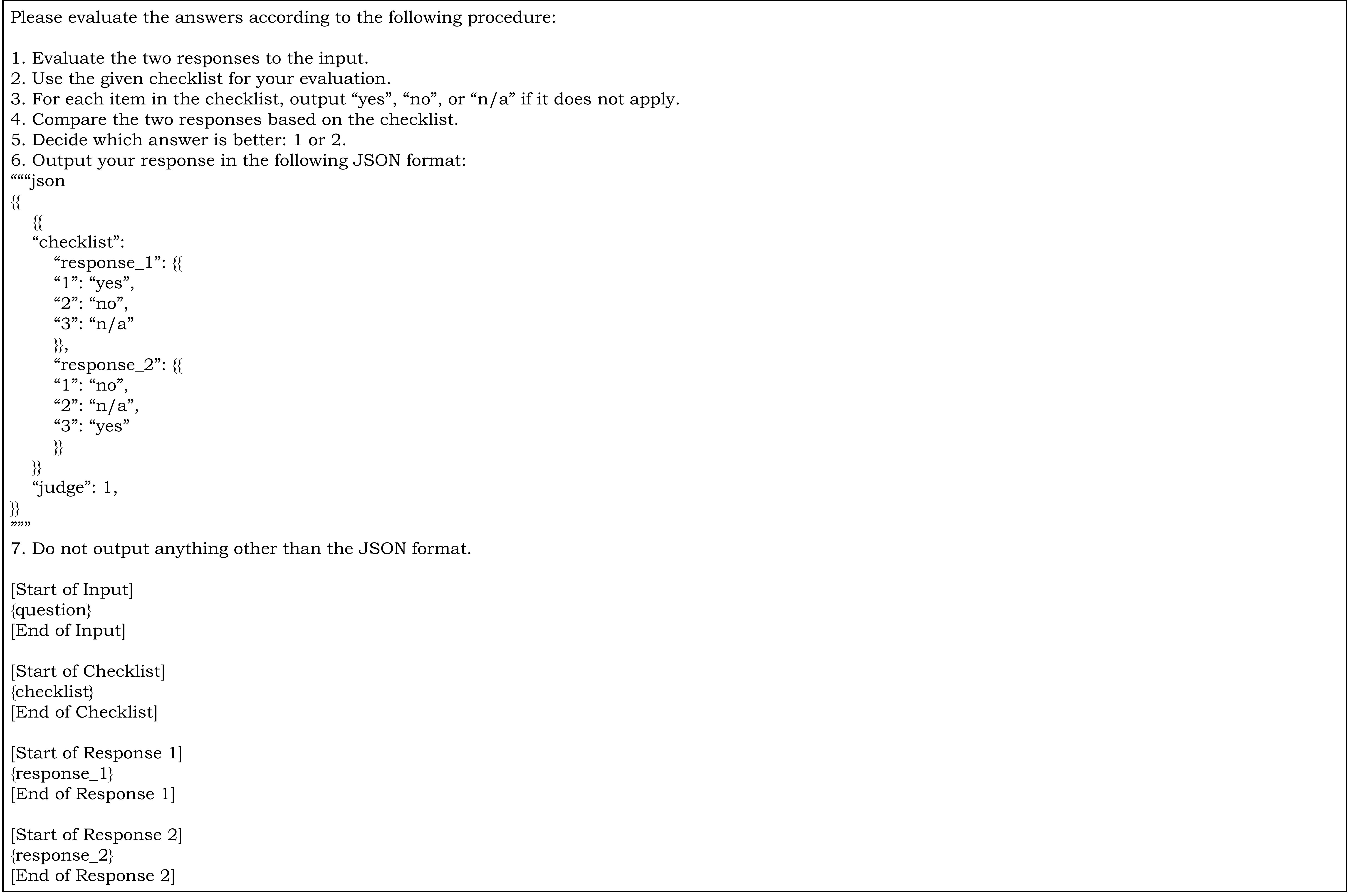} 
    \caption{Prompt used to evaluate response pairs using checklist-based criteria.}
   \label{with_checklist_pairwise}
\end{figure*}

\begin{figure*}[h]
    \centering
    \includegraphics[width=\textwidth, keepaspectratio]
    {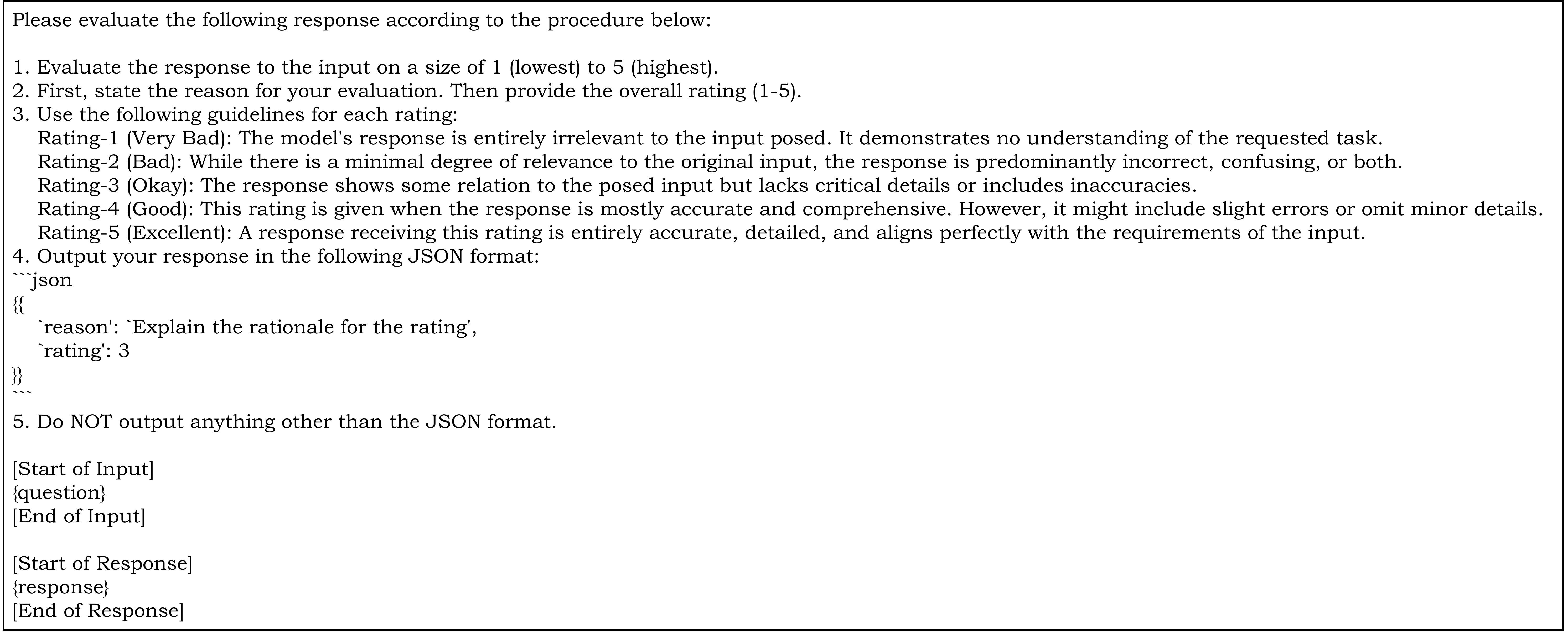}
     \caption{Prompt used for direct scoring evaluation without checklists.}
     \label{without_checklist_scoring}
\end{figure*}

\begin{figure*}[h]
    \centering
    \includegraphics[width=\textwidth, keepaspectratio]
{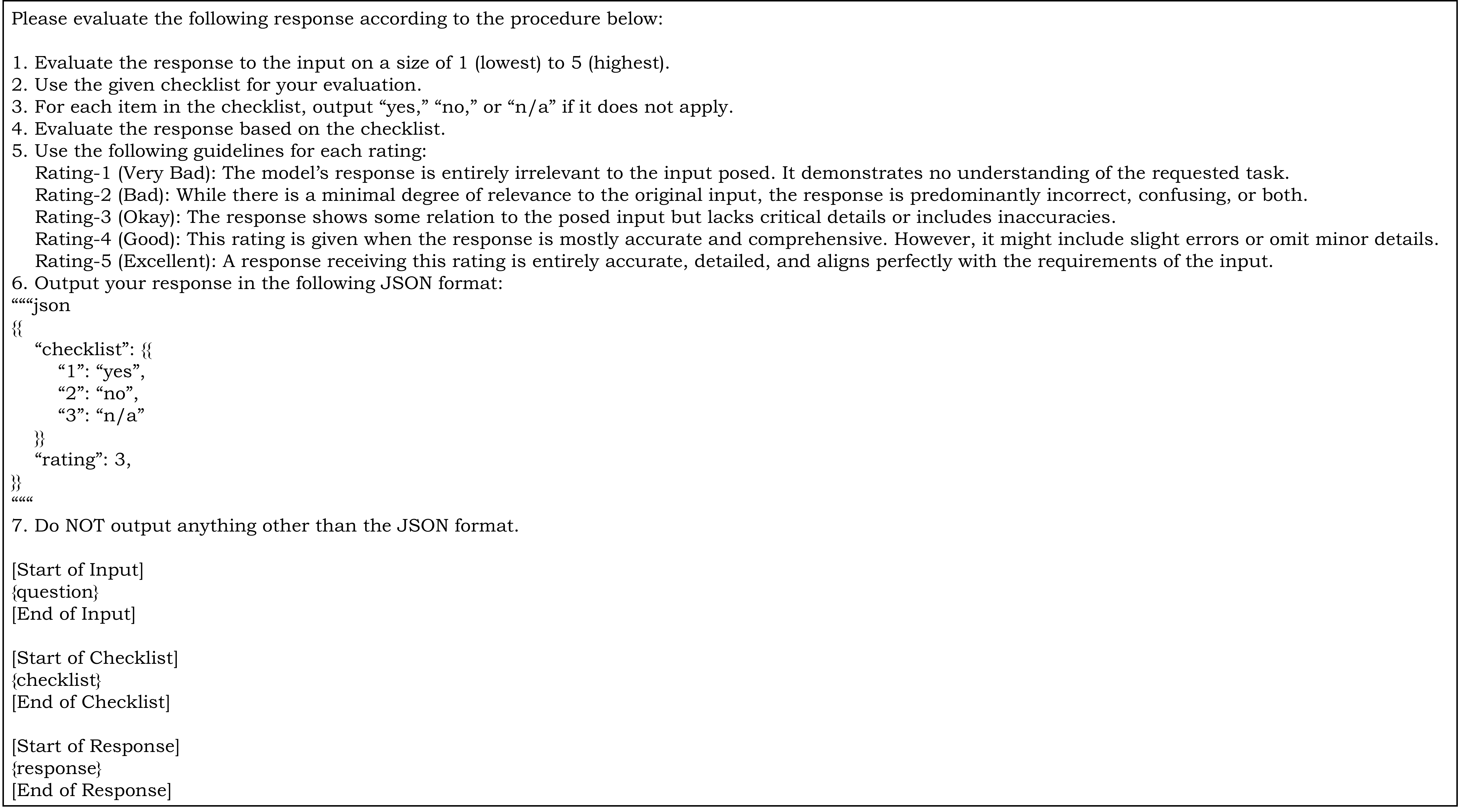} 
 \caption{Prompt used for direct scoring evaluation with checklists.}
 \label{with_checklist_scoring}
\end{figure*}

Figures~\ref{without_checklist_pairwise}, \ref{with_checklist_pairwise}, \ref{without_checklist_scoring}, and \ref{with_checklist_scoring} present the prompts used for evaluating responses.
Figures~\ref{without_checklist_pairwise} and \ref{with_checklist_pairwise} show the evaluation prompts for the pairwise comparison datasets, while Figures~\ref{without_checklist_scoring} and \ref{with_checklist_scoring} show the prompts for the direct scoring datasets.

\end{document}